\documentclass[journal]{IEEEtran}

\usepackage{cite}
\usepackage{amsmath}
\usepackage{algorithm}
\usepackage{algpseudocode}
\usepackage{array}
\usepackage[caption=false,font=footnotesize]{subfig}
\usepackage{soul}
\usepackage{amsmath}
\usepackage{graphicx}
\usepackage{color}
\usepackage{multirow}
\usepackage{epstopdf}
\usepackage{ulem}
\usepackage{amsfonts}
\usepackage{booktabs}
\usepackage{multirow}

\begin{document}
\bstctlcite{bstctl:nodash}

\title{PNEN: Pyramid Non-Local Enhanced Networks}

\author{Feida~Zhu,~\IEEEmembership{}
        Chaowei Fang,~\IEEEmembership{}
        and~Kai-Kuang~Ma,~\IEEEmembership{Fellow,~IEEE}
\thanks{F. Zhu and K.-K. Ma are with the School of Electrical \& Electronic Engineering, Nanyang Technological University, Singapore (e-mail: {feida.zhu, ekkma}@ntu.edu.sg).}
\thanks{C. Fang is with the School
of Artificial Intelligence, Xidian University, No. 2 South Taibai Road, Xi’an, Shaanxi, 710071, P. R. China (e-mail: cwfang@xidian.edu.cn).}
\thanks{Corresponding author: Chaowei Fang.}
}

\markboth{IEEE TRANSACTIONS ON IMAGE PROCESSING,~Vol.~xx, No.~x, xx~xxxx}
{F. Zhu \MakeLowercase{\textit{et al.}}: Pyramid Non-Local Enhanced Networks} 

\maketitle

\begin{abstract}
Existing neural networks proposed for low-level image processing tasks are usually implemented by stacking convolution layers with limited kernel size. Every convolution layer merely involves in context information from a small local neighborhood. More contextual features can be explored as more convolution layers are adopted. However it is difficult and costly to take full advantage of long-range dependencies. We propose a novel non-local module, \textit{Pyramid Non-local Block}, to build up connection between every pixel and all remain pixels. The proposed module is capable of efficiently exploiting pairwise dependencies between different scales of low-level structures. The target is fulfilled through first learning a query feature map with full resolution and a pyramid of reference feature maps with downscaled resolutions. Then correlations with multi-scale reference features are exploited for enhancing pixel-level feature representation. The calculation procedure is economical considering memory consumption and computational cost.
Based on the proposed module, we devise a \textit{Pyramid Non-local Enhanced Networks} for edge-preserving image smoothing which achieves state-of-the-art performance in imitating three classical image smoothing algorithms. Additionally, the pyramid non-local block can be directly incorporated into convolution neural networks for other image restoration tasks. We integrate it into two existing methods for image denoising and single image super-resolution, achieving consistently improved performance. 
\end{abstract}

\section{Introduction}
\IEEEPARstart{I}{mpressive} progress has been achieved in low-level computer vision tasks as the development of convolution neural networks (CNN), e.g. edge-preserving image smoothing \cite{li2016deep,fan2017generic,zhu2019benchmark}, image denoising \cite{mao2016image,zhang2017beyond,zhang2020residual} and image super-resolution \cite{Tai-DRRN-2017,Tai-MemNet-2017,zhang2020residual}. In this paper, we propose a novel pyramid non-local block oriented for effectively and efficiently mining long-range dependencies in low-level image processing tasks.

Inherently there exist many texels with high similarity in natural images. In~\cite{efros1999texture}, such self-similarities is exploited to synthesize textures with realistic appearances. Several non-local algorithms~\cite{buades2005non,dabov2007bm3d,mairal2009non} are devised for exploring dependencies between similar texels in image restoration tasks. They mainly focus on estimating similarities between local patches. Each query patch is reconstructed with similar reference patches which are probably distant to the query patch. Dependencies on similar texels should be beneficial to other low-level image processing tasks as well. For example, in edge-aware image smoothing, similar textures are very likely to spread on the surface of the same object. Non-local correlations are beneficial for erasing such textures and identifying the real object edges. In image super-resolution, we can make use of the replication of structures to recover degraded content which is caused by small spatial size. 

Recently convolution neural networks have been extensively applied in pixel-level image processing tasks. Typical convolution layers operate on a small local neighborhood without considering non-local contextual information. One common practice for capturing long-range dependencies is to enlarge the receptive field by stacking large number of convolution layers \cite{kim2016accurate} or dilated convolution layers \cite{yu2017dilated}. However, it is difficult to deliver information between distant positions in such a manner \cite{hochreiter1997long}. To make full usage of long-range dependencies, a few literatures \cite{lefkimmiatis2017non,wang2018non,liu2018non,li2018non} propose non-local algorithms which can be integrated into deep models, through enhancing feature representation with self-similarities. Reference~\cite{wang2018non} presents non-local neural networks which enhances the feature representation of each position with its correlations to all remain positions, for video classification. In~\cite{li2018non}, non-local neural networks are further applied into image de-raining with the help of the encoder-decoder network. The computational cost and memory consumption of the non-local operation arise quadratically as the spatial size of input feature map increases. Considering the limitation of memory resource, non-local blocks are usually placed after downscaled high-level feature maps. This hinders their adaptation to low-level computer vision tasks, where high-resolution feature maps are demanded to produce appealing pixel-level outputs. Reference~\cite{liu2018non} proposes a non-local recurrent network (NLRN), which confines the neighborhood for calculating pairwise correlations and achieves excellent performance in image restoration. One drawback of using limiting neighborhood is that only correlation information within tens of pixels are explored while valuable dependencies from distant pixels are ignored. Last but not the least, similar texels usually appear to possess various spatial scales due to their intrinsic physical properties or scene depth. A robust method to estimate similarities between different scales of texels is paramount for sufficiently excavating the contextual correlation information.

To efficiently explore non-local correlation information in pixel-level image processing tasks, we propose a pyramid structure, named \textit{Pyramid Non-local Block} (PNB). Alike to \cite{wang2018non}, query and reference feature maps are used to set up correlations between different positions, which is subsequently employed to enhance the pixel-level feature representation. The novelty of our method is inspired from two aspects. First, for sake of relieving the computation burden of non-local operation, we adopt reference feature map with downscaled resolution while preserving the resolution of query feature map. Secondly a pyramid of reference feature maps are constructed to robustly estimate correlations between texels with different scales. Through intervening pyramid non-local blocks and dilated residual blocks \cite{yu2017dilated}, we set up a novel deep model, \textit{Pyramid Non-Local Enhanced Networks}, for edge-preserving image smoothing. It achieves state-of-the-art performance in imitating various classical image smoothing filters. In addition, the pyramid non-local blocks can be easily incorporated into deep CNN-based methods for other pixel-level image processing tasks. We demonstrate the effectiveness of PNB on two classical tasks, image denoising and single image super-resolution (SISR). Two existing models, RDN \cite{zhang2020residual} and MemNet \cite{Tai-MemNet-2017}, are adopted as baseline models for both image denoising and SISR. The PNB-s are plugged into them to exploit long-range dependencies. Performance improvements over baseline models have been consistently achieved thanks to the adoption of PNB-s.

Main contributions of this manuscript can be summarized as follows.
\begin{itemize}
    \item[1)] A pyramid non-local block is proposed for exploiting long-range dependencies in low-level image processing.
    \item[2)] Based on pyramid non-local blocks and dilated residual blocks, we set up a novel model, \textit{Pyramid Non-local Enhanced Network}, which achieves state-of-the-art performance in edge-preserving image smoothing.
    \item[3)] We integrate the pyramid non-local block into existing methods, RDN \cite{zhang2020residual} and MemNet \cite{Tai-MemNet-2017}, achieving improved performance in image denoising and single image super-resolution.
\end{itemize}

\begin{figure*}[!t]
\centering
\includegraphics[width=1\textwidth]{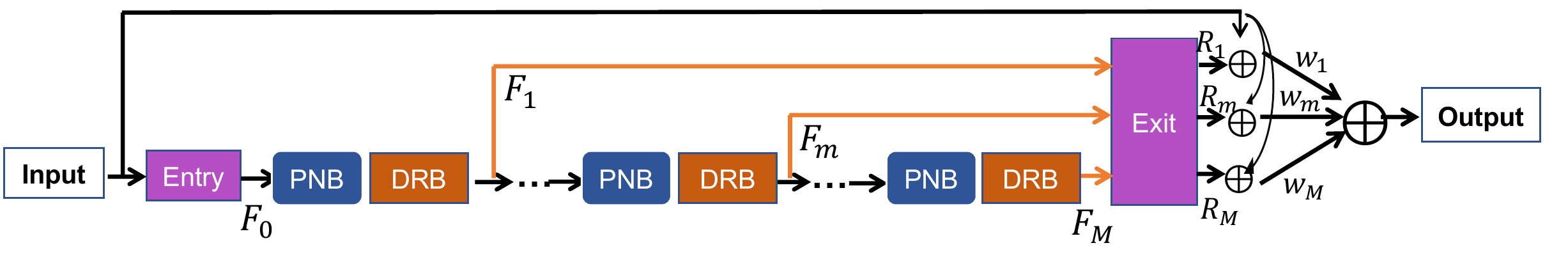}
\caption{The overall architecture of our proposed pyramid non-local enhanced network (PNEN). }
\label{fig:architecture}
\end{figure*}

 \begin{figure*}[!t]
\captionsetup[subfigure]
{subrefformat=simple, listofformat=subsimple,farskip=1pt,justification=centering}
\centering
\subfloat[Non-local Block]{\includegraphics[width=0.32\textwidth]{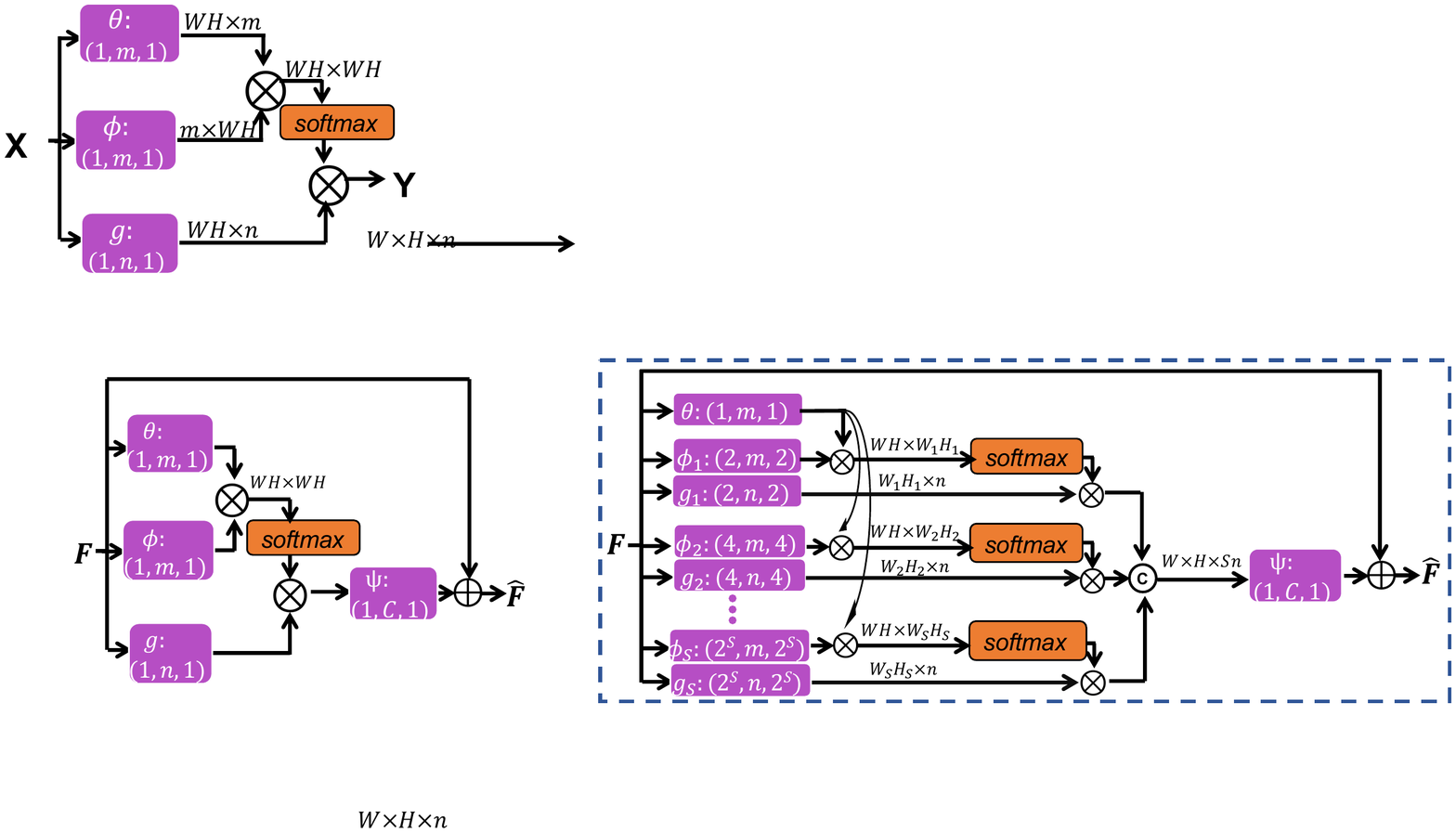}}
\subfloat[Pyramid Non-local Block]{\includegraphics[width=0.68\textwidth]{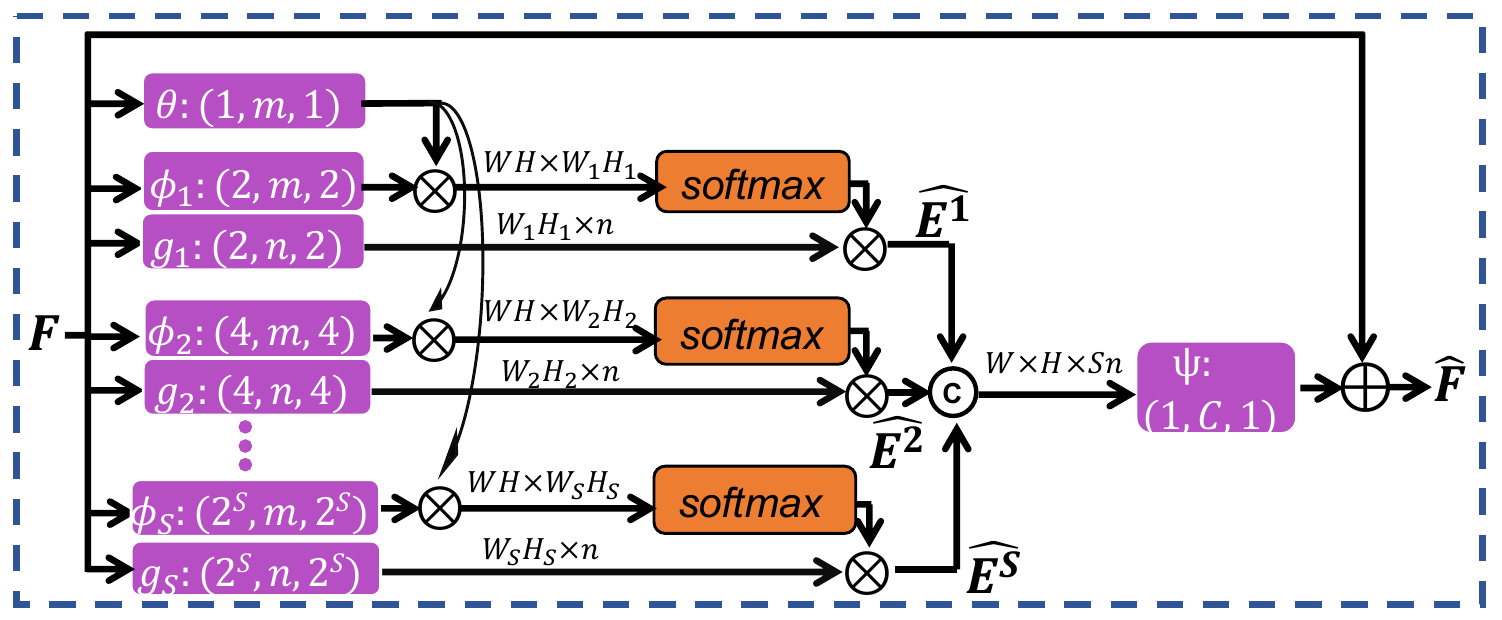}}
\caption{Architecture of prior non-local block \cite{wang2018non} and our pyramid non-local block (PNB). In PNB, $\phi_k$ and $g_k$ are implemented with convolutional layers with different strides, while $\theta$ is shared in all scales. The kernel size \textit{k}, number of filters \textit{f} and stride \textit{s} are indicated as (\textit{k}, \textit{f}, \textit{s}) for each convolution layer.}
\label{fig:PNB}
\end{figure*}

\section{Related Work}

\subsection{Deep Learning based Low-level Image Processing}

Because of the strong feature learning ability, deep neural networks attract a lot of attention in low-level image processing. In this section, we will give an overview to relevant literatures in edge-preserving image smoothing, image denoising and super-resolution. Detailed survey on image processing is not elaborated here.

Edge-preserving image smoothing aims to preserve significant image structures while filtering out trivial textures. There exist piles of traditional algorithms solving this problem such as filtering based methods~\cite{tomasi1998bilateral,ma2013constant,zhang2014100+,zhang2014rolling} and variation based algorithms~\cite{subr2009edge,xu2011image,ham2017robust}.  The pioneering deep learning based method \cite{xu2015deep} employs a three-layer CNN to predict a gradient map which is subsequently used to guide the smoothing procedure. In \cite{liu2016learning}, a recurrent network is adopted to efficiently propagate spatial contextual information across pixels. Reference \cite{fan2017generic} demonstrates a cascaded framework which implements image reflection removal and image smoothing under guidance of a predicted edge map. Reference \cite{fan2018image} presents an unsupervised deep model, which is optimized with a handcrafted objective function. Reference \cite{lu2018deep} attempts to identify out structures and textures in input images, facilitating structure and texture aware filtering.  A CNN based pipeline is proposed for joint image filtering (e.g., RGB and depth images) in \cite{li2016deep}.

Deep neural networks have been extensively applied in image super-resolution and image denoising. In \cite{dong2015image}, the authors put forward a 3-layer CNN for image super-resolution. Since then, there emerged lots of literatures which use deep learning to settle the image super-resolution task. Reference \cite{kim2016accurate} provides effective strategies (e.g. using residual learning and large learning rate) to train very deep networks. Based on residual learning, a pyramid network is further proposed to progressively enlarge input images in \cite{lai2017deep}. Residual blocks \cite{he2016deep} and dense connections \cite{huang2017densely} are widely adopted in recent super-resolution models such as \cite{ledig2017photo,lim2017enhanced,li2018multi,fang2019self,zhang2020residual}. Similar to image super-resolution, denoising is also a pixelwise image restoration problem. Reference \cite{NIPS2012_4686} employs stacked denoising auto-encoders to train a deep CNN model for image restoration. Reference \cite{mao2016image} presents a encoder-decoder architecture with skip connections. Residual learning is introduced to solve blind Gaussian denoising with CNNs in \cite{zhang2017beyond}. A recursive unit for memorizing context information learned with various receptive fields is proposed in \cite{Tai-MemNet-2017}. In \cite{zhang2018image}, the authors exploit residual in residual architectures to build up very deep networks and adopts global channel attention to enhance the learning of high-frequency information. 

The generative adversarial networks~\cite{goodfellow2014generative} is famous for generating samples through the minimax optimization. It has been applied in many image restoration tasks, such as image super-resolution~\cite{ledig2017photo,wang2018esrgan} and denoising~\cite{chen2018image}, to pursue visually realistic outputs. Even though the models trained with adversarial loss may achieve lower PSNR compared to those trained with pixel-level reconstruction loss, they can acquire significant gains in perceptual quality.

This paper concentrates on exploring non-local dependencies with convolutional neural networks for pixelwise image processing. We propose a pyramid non-local block which can efficiently make use of global correlation information to enhance pixelwise feature representation. Furthermore, pyramid non-local enhanced networks are built up, achieving state-of-the-art performance in imitating three classical edge-preserving image smoothing methods \cite{zhang2014100+,xu2011image,ham2017robust}. Based on existing methods \cite{zhang2020residual,Tai-MemNet-2017}, our method produces appealing results in image super-resolution and denoising.

\subsection{Non-local Context Information}
Image non-local self-similarity has been widely exploited in many non-local methods for image restoration, such as \cite{dabov2007bm3d} and \cite{gu2014weighted}. Recently a few studies attempt to incorporate non-local operations into deep neural networks for capturing long-range dependencies. References \cite{lefkimmiatis2017non,cruz2018nonlocality,wang2018non} present trainable non-local neural networks based on non-local filtering, for image denoising and video classification respectively. However, the computation complexity of their non-local operations grows dramatically as size of the input feature map increases. A non-local module is incorporated  into the RNN architecture for image restoration in \cite{liu2018non}. Nevertheless, the measurement of self-similarity is restricted within a small neighborhood. Our method differs from the above non-local models in two points. Firstly, our method can robustly measure similarities between different scales of texels as we adopt a pyramid structure for the non-local operation. Secondly, computation burden and memory consumption is greatly relieved because the spatial resolutions of reference features in the non-local operation are downscaled. An asymmetrical pyramid non-local block (APNB) is proposed in \cite{zhu2019asymmetric}, using multi-scale features which are sampled via pyramid pooling as the key and value representations. It is implemented through computing a uniform correlation matrix for all key representations. Our proposed pyramid nonlocal block deals with the correlation with different scales of representations independently which can fully take advantage of multi-scale correlation information. Besides, APNB uses the max pooling operation to generate multi-scale feature maps with very low resolutions, while our method employs parallel convolution layers to produce multi-scale feature maps. Very recently, non-local blocks are employed to help computing local and non-local attention for enhancing the feature representation in image restoration \cite{zhang2019residual}.

\begin{figure*}[!t]
\centering
\includegraphics[width=0.8\textwidth]{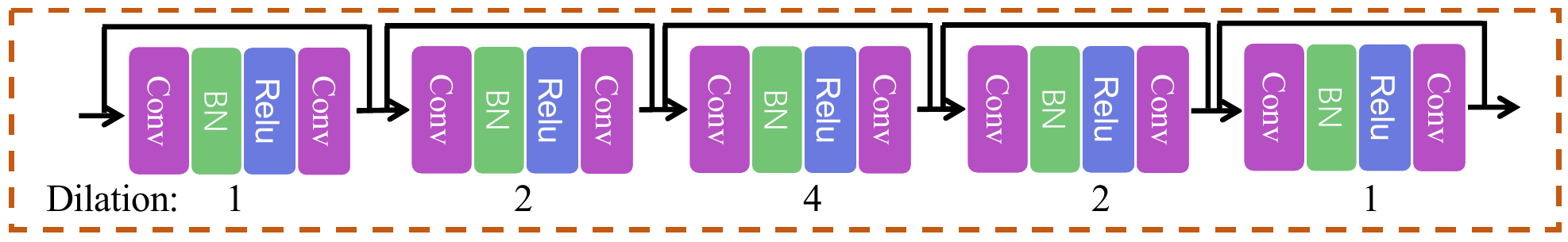}
\caption{Architecture of dilated residual block (DRB). Our proposed DRB contains five Conv-BN-ReLU-Conv groups, which have dilation factor 1, 2, 4, 2, 1, respectively.  }
\label{fig:DRB}
\end{figure*}

\section{Method}

In pixel-level image processing tasks, non-local correlations between texels can help providing contextual information and exploring dependencies on other similar texels.
In this paper, we propose a deep pyramid non-local enhanced network (PNEN) for edge-preserving image smoothing. It adopts pyramid non-local blocks (PNB) to mine long-range correlation information and the overall architecture is illustrated in Fig. \ref{fig:architecture}. The pyramid non-local block is carefully designed to involve in correlations with multi-scale texels. At the same time, high computation efficiency is guaranteed when estimating non-local correlations. Dilated residual blocks (DRB) \cite{yu2017dilated} are employed to extract full structural and textural information from the input image. In the following sections, the proposed architecture will be elaborated in detail.

\subsection{Entry and Exit Network}\label{sec:network}
Define the input color image as $\mathbf X$ with size of $h\times w \times c$. $h$, $w$ and $c$ represents the height, width and channel of the input image respectively. Our proposed PNEN uses one convolution layer as the entry net to extract a pixel-level feature map $\mathbf F_0$  (with size of $h\times w\times d$). Formally, we have,
\begin{align}
\label{eq:entry}
     \mathbf F_{0} = \mathcal{F}_{entry}(\mathbf X,\mathbf W_{entry}),
\end{align} 
where the $\mathcal{F}_{entry}(\cdot,\cdot)$ denotes the convolution operation in the entry net and $\mathbf W_{entry}$ represents related convolution parameters. Subsequently, $M$ blocks, each of which is consisting of a pyramid non-local block and a dilated residual block, are stacked to induce deep features. We define the feature produced by the $m$-th block as $\mathbf F_m$. We have,
\begin{align}
\label{eq:pnb-drb}
     \mathbf F_{m} = \mathcal{F}_{PNB}(\mathcal F_{DRB}(\mathbf F_{m-1}, \mathbf W_{DRB}^m),\mathbf W_{PNB}^m),
\end{align} 
where $\mathcal{F}_{PNB}(\cdot,\cdot)$ and $\mathcal{F}_{DRB}(\cdot,\cdot)$ indicates the calculation procedure inside the pyramid non-local block and dilated residual block which will be elaborated in Section \ref{sec:pnb} and \ref{sec:drb} respectively. $\mathbf W_{DRB}^m$ and $\mathbf W_{PNB}^m$ represent their parameters correspondingly.
Inspired by MemNet \cite{Tai-MemNet-2017}, features generated by all blocks $\{\mathbf F_m|m=1,\cdots,M\}$ are accumulated to generate residual images using the exit network. The residual image produced with $\mathbf F_m$ is defined as,
\begin{align}
\label{eq:exit}
     \mathbf R_m = \mathcal{F}_{exit}(\mathbf F_m,\mathbf W_{exit}^m),
\end{align} 
where $\mathcal F_{exit}(\cdot,\cdot)$ denotes the convolution operations in the exit network and $\mathbf W_{exit}^m$ represents the related parameters. The exit network contains three convolutional layers. Suppose $\mathbf Y_m=\mathbf X+\mathbf R_m$. The final reconstructed images is computed by,
\begin{align}
\label{eq:final}
     \mathbf Y = \sum_{1}^{M} w_m \cdot \mathbf Y_m,
\end{align} 
where $\{w_m|m=1,\cdots,M\}$ are trainable weights. During the training stage, supervisions are imposed to intermediate predictions $\mathbf Y_m$-s and the final output $\mathbf Y$. The mean squared error is used as the loss function:
\begin{align}
\label{eq:loss}
     \mathcal L= \frac{1}{hwc}(\|\mathbf{G}-\mathbf Y\|^2 + \sum_{m=1}^M\|\mathbf{G}-\mathbf Y_m\|^2),
\end{align} 
where $\mathbf G$ represents the ground truth image.

\subsection{Pyramid Non-local Block (PNB)} \label{sec:pnb}

Let $\mathbf F \in \mathbb{R}^{h\times w \times d}$ denotes the input feature activation map.
Here $h$, $w$ and $d$ represents the height, width and channel, respectively. A general formulation of non-local operation~\cite{wang2018non} can be defined as,
\begin{align}
\label{eq:matrix_form}
     \mathbf{\hat F}=\mathcal T(\frac{1}{\mathcal D(\mathbf F)} \mathcal{M}(\mathbf{F})\mathcal{G}(\mathbf{F}))+\mathbf F,
\end{align} 
where $\mathbf{\hat F}$ is the enhanced feature representation. $\mathcal{M}(\mathbf F) \in \mathbb{R}^{hw \times hw}$ is the self-similarity matrix, where each element $\mathcal{M}(\mathbf F)_{i,j}$ indicates the similarity between pixel $i$ and $j$. $\mathcal{G}(\mathbf F) \in \mathbb{R}^{hw \times n}$ gives rise to a $n$-dimensional pixel-wise embedding. $\mathcal{D}(\mathbf F)$ produces a diagonal matrix for normalization purpose. $\mathcal T(\cdot)$ is a transformation function which converts the embedded $n$-dimensional features back into the original $d$-dimensional space of input feature. In this way, the feature representation is non-locally enhanced through involving in correlations with all positions in the feature map.
One instantiation \cite{wang2018non} can be constructed by taking the linear embedded Gaussian kernel as the distance metric to compute correlation matrix $\mathcal{M}$, and linear function to compute $\mathcal{G}$:
\begin{align}
\label{eq:embedded Gaussian}
  \mathcal M(\mathbf F) & = \textrm{exp}(\mathcal F_{emb}(\mathbf F, \mathbf W_\theta) \mathcal F_{emb}(\mathbf F, \mathbf W_\phi)^{\textrm T}), \\
  \mathcal G(\mathbf F) & = \mathcal F_{emb}(\mathbf F, \mathbf W_{g}). 
\end{align} 
The embedding function $\mathcal F_{emb}(\mathbf F, \mathbf W)$ is implemented with a convolutional operation of parameter $\mathbf W$. The result is flattened into a 2-dimensional tensor in which each column represents one embedding channel. When calculating $\mathcal M(\mathbf F)$, a query and a reference feature with same size of $hw\times m$ are generated using convolution kernel $\mathbf W_\theta$ and $\mathbf W_\phi$ respectively.  
The diagonal elements of $\mathcal D(\mathbf F)$ are obtained through calculating column summation of $\mathcal M(\mathbf F)$. $\mathcal T(\cdot)$ is also implemented with a convolution operation of parameter $\mathbf W_{\psi}$. 
All convolutions use kernel size of $1\times1$. An example non-local block is illustrated in Fig. \ref{fig:PNB}(a). The computation complexity and memory occupation of the correlation matrix increases quadratically as the number of pixels grows. For sake of reducing computation burden, previous work \cite{liu2018non} utilize a small neighborhood to restrict the range of non-local operation. In comparison, we propose a novel pyramid non-local block to effectively mitigate the computation demand.

At first, we produce one query feature, $\mathbf E_{\theta}=\mathcal F_{emb}(\mathbf F, \mathbf W_\theta)$, using one convolution layer. The spatial kernel size and stride of $\mathbf W_\theta$ is $1\times 1$ and $1$ respectively. Then, multi-scale reference features and embedding features can be generated with 
parallel convolutions using different kernel sizes and strides. 
Suppose there are totally $S$ scales.
$S$ independent branches are devised to compute multi-scale key and embedding features. To extract feature representations of texels with larger scales, larger convolution kernels are used to calculate high-level feature maps. Strides of convolution operations are increased as kernel sizes to reduce the resolution of feature maps.
\begin{align}
  \label{eq:ppnb-embr}\mathbf E_\Phi^s& = \mathcal F_{emb}(\mathbf F, \mathbf W_\Phi^s),\\
  \label{eq:ppnb-embg}\mathbf E_g^s &= \mathcal F_{emb}(\mathbf F, \mathbf W_g^s),
\end{align}
The kernel size and stride of the convolution layer in the $s$-th scale are both set to $2^s$. This implies that the number of rows in $\mathbf E_{\Phi}^s$ and $\mathbf E_{g}^s$ is reduced to $hw/4^s$. Thus the amount of computation when calculating the self-similarity matrix can be greatly reduced.

The non-local operation in the $s$-th scale is executed as in the following formulation, which gives rise to an enhanced embedding feature $\mathbf{\hat E}^s$.  
\begin{equation}
\label{eq:pnb-nonlocal}
\mathbf{\hat E}^s =  \frac{1}{\mathbf D^s}\text{exp}\{ \mathbf E_\theta (\mathbf E_\Phi^s)^{\textrm T}\} \mathbf E_g^s.
\end{equation}

Finally, the enhanced embedding features $\{\mathbf{\hat E}^s|s=1,\cdots,S\}$ are concatenated together, followed by one $1 \times 1$  convolution layer to produce residual values to $\mathbf F$. Formally, the final output of the pyramid non-local block can be achieved by,
\begin{align}
\label{eq:pnb-trans}
   \mathbf{\hat F} =  \mathcal F_{\Psi} (\{\mathbf {\hat E}^1,\cdots,\mathbf {\hat E}^S\},\mathbf W_{\psi})  + \mathbf F.
\end{align}

Except for relieving the computation burden of the non-local operation, the specific design of our pyramid non-local module can enhance the feature representation capability with multi-scale self-similarities. We summarize the whole computation procedure with the function $\mathcal{F}_{PNB}(\cdot,\cdot)$, as mentioned in Section \ref{sec:network}. One characteristic of pyramid non-local block is the flexibility of balancing accuracy and computation resources through adjusting kernel sizes and strides in different scales. An illustration of the pyramid non-local block is shown in Fig. \ref{fig:PNB}(b). We set $m=64, n=32$ and $S=3$ in practice.

\subsection{Dilated Residual Block (DRB)}
\label{sec:drb}
In edge-preserving image smoothing, high-resolution feature maps are favorable for reconstructing complicate textural details while large receptive field benefits the capability of grabbing high-level contextual information. Considering the above issues, we employ dilated convolutions \cite{yu2017dilated} to rapidly increase the receptive field without sacrificing spatial resolutions of intermediate feature maps.  As shown in Fig. \ref{fig:DRB}, 5 residual modules equipped with dilated convolutions are cascaded, forming an independent architecture named dilated residual block. The calculation procedure of DRB is indicated with the function $\mathcal{F}_{DRB}(\cdot,\cdot)$, as mentioned in Section \ref{sec:network}.

As illustrated in Fig. \ref{fig:architecture}, the PNEN is built up through intervening PNB-s and DRB-s. Every group of consecutive PNB and DRB contributes a feature map for the final prediction as described in (\ref{eq:pnb-drb}), (\ref{eq:exit}) and (\ref{eq:final}).

\begin{figure*}[!t]
\captionsetup[subfigure]
{subrefformat=simple, listofformat=subsimple,farskip=1pt,justification=centering}
\centering
\subfloat[Original Image]{\includegraphics[width=0.32\textwidth]{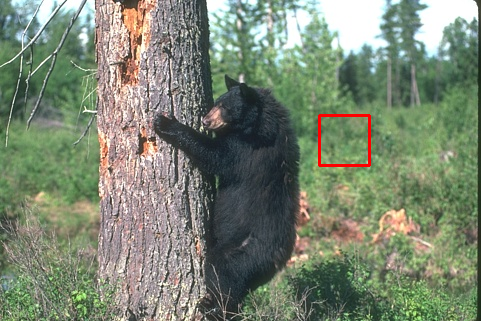}}
\hspace{-0.1\textwidth}\subfloat{\includegraphics[width=0.1\textwidth]{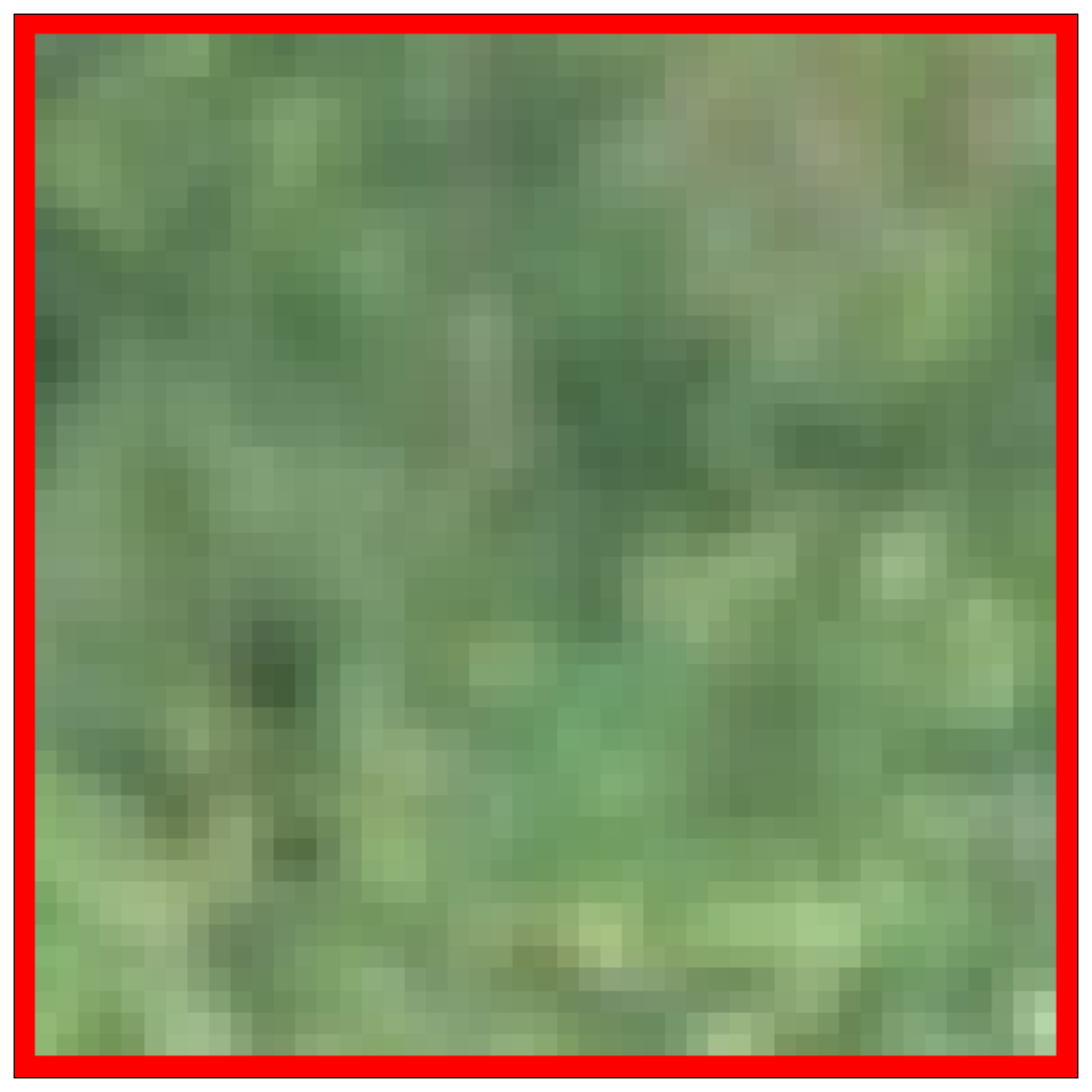}}
\addtocounter{subfigure}{-1}
\subfloat[Ground Truth]{\includegraphics[width=0.32\textwidth]{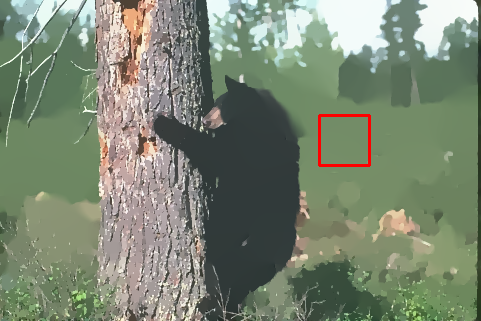}}
\hspace{-0.1\textwidth}\subfloat{\includegraphics[width=0.1\textwidth]{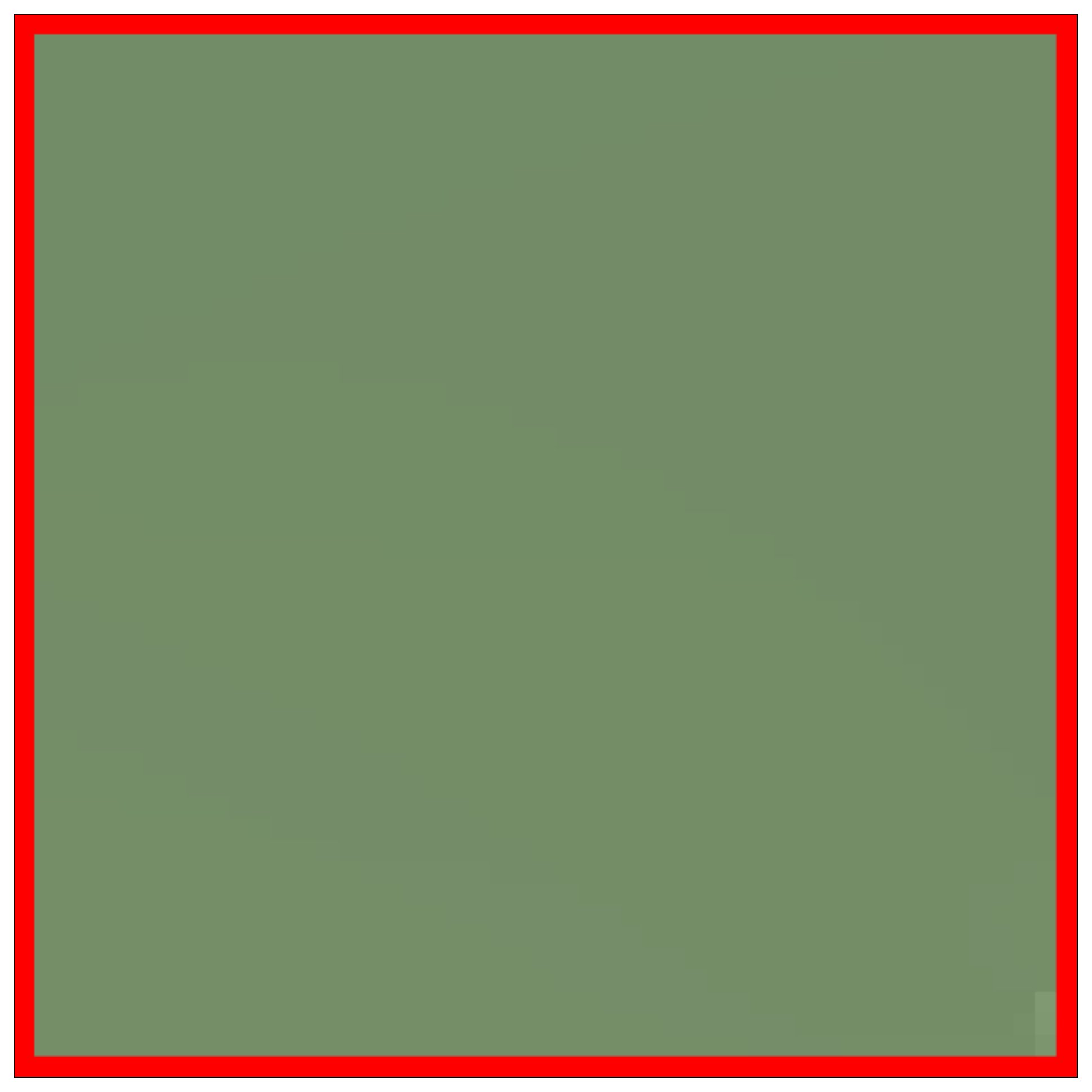}} 
\addtocounter{subfigure}{-1}
\subfloat[DJF \cite{li2016deep}]{\includegraphics[width=0.32\textwidth]{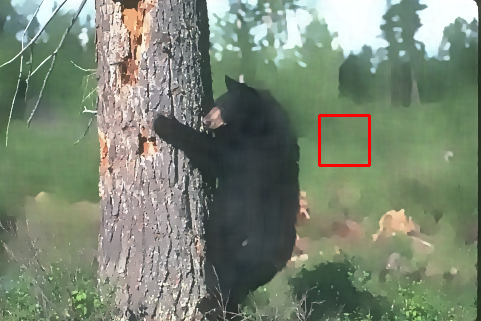}}
\hspace{-0.1\textwidth}\subfloat{\includegraphics[width=0.1\textwidth]{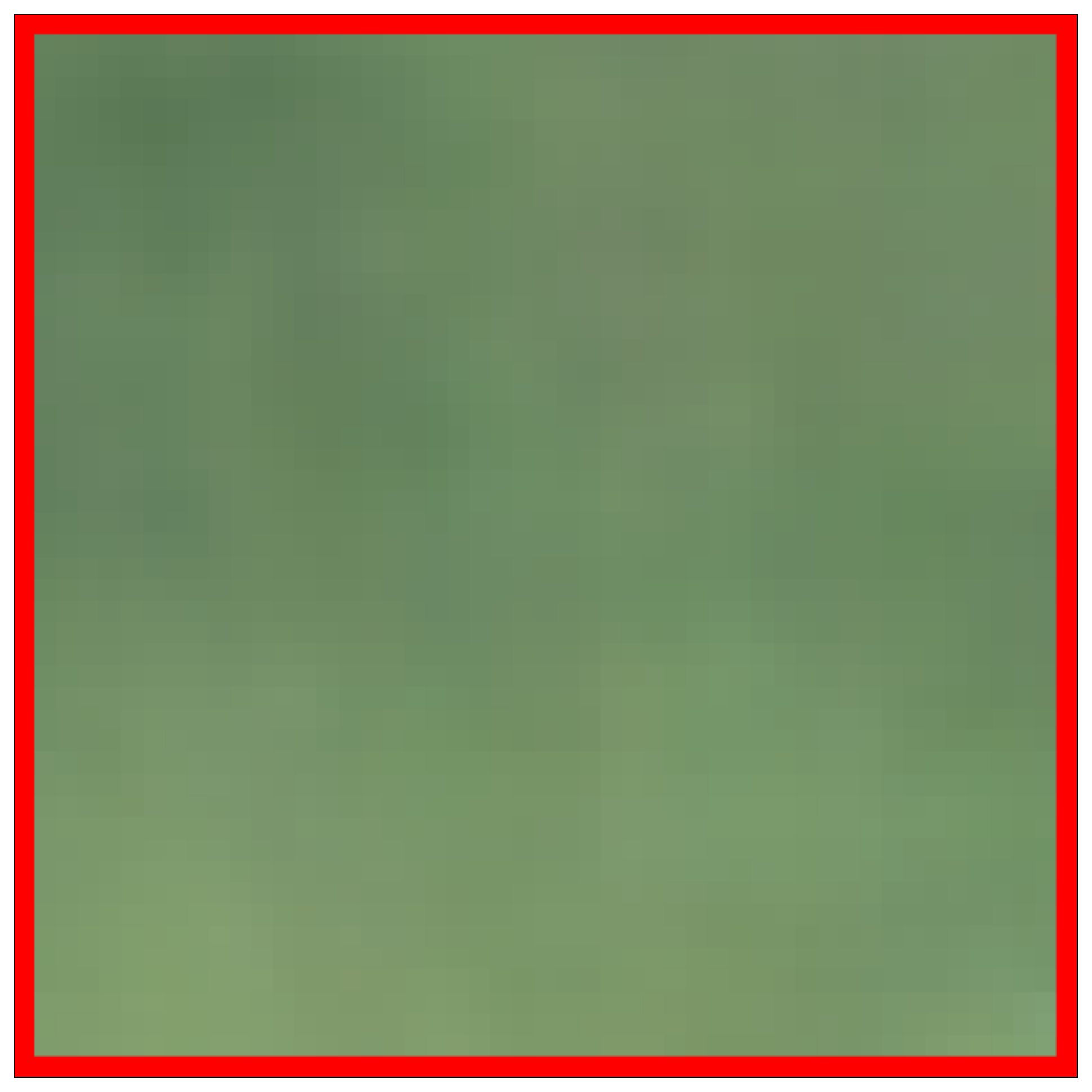}} 
\addtocounter{subfigure}{-1}
\subfloat[CEILNet \cite{fan2017generic}]{\includegraphics[width=0.32\textwidth]{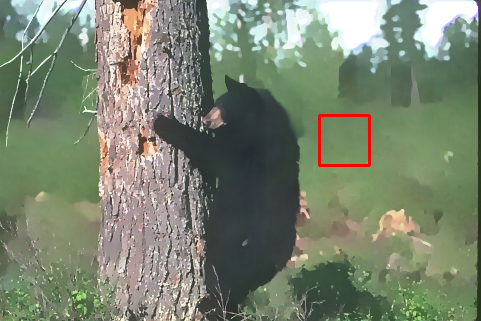}}
\hspace{-0.1\textwidth}\subfloat{\includegraphics[width=0.1\textwidth]{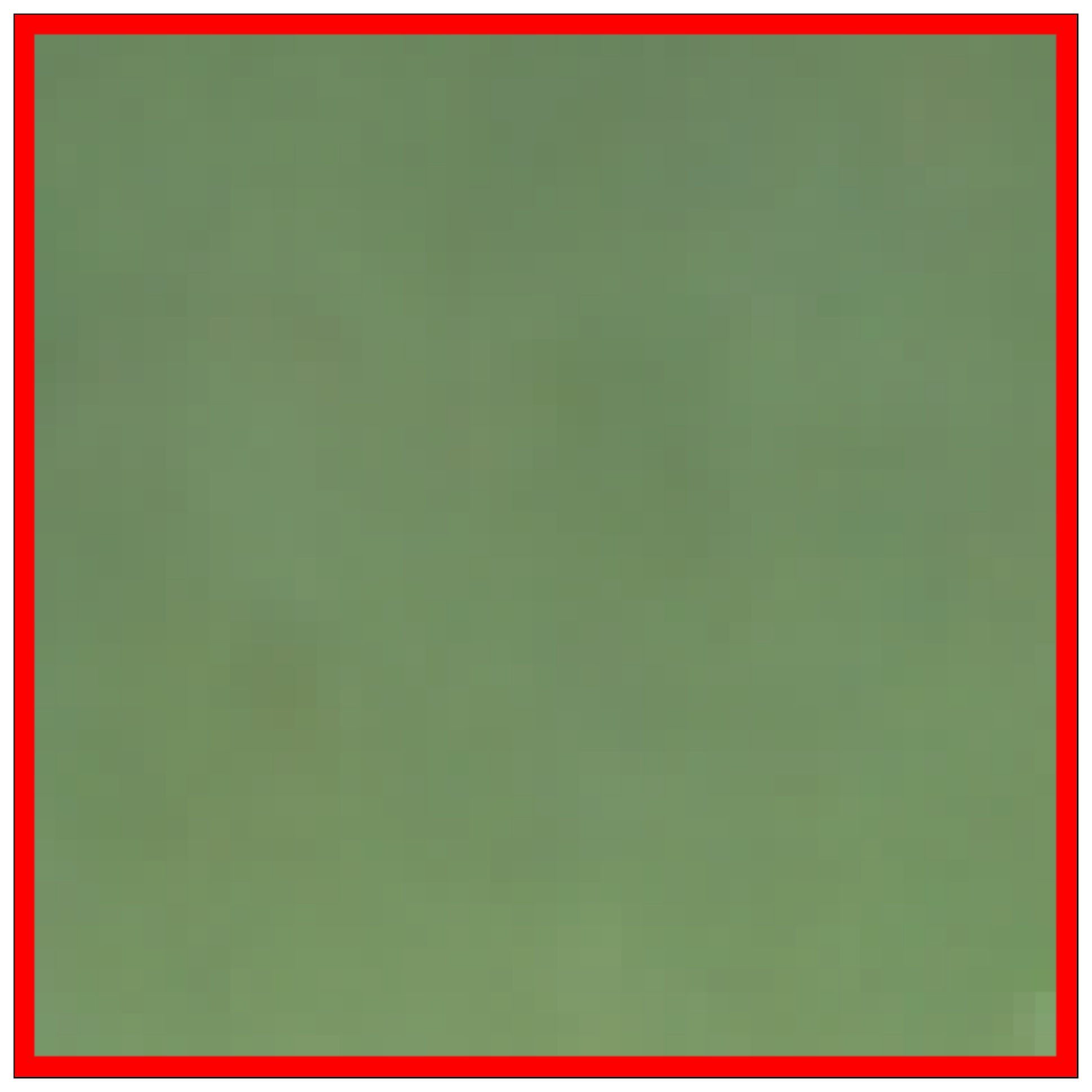}} 
\addtocounter{subfigure}{-1}
\subfloat[ResNet \cite{zhu2019benchmark}]{\includegraphics[width=0.32\textwidth]{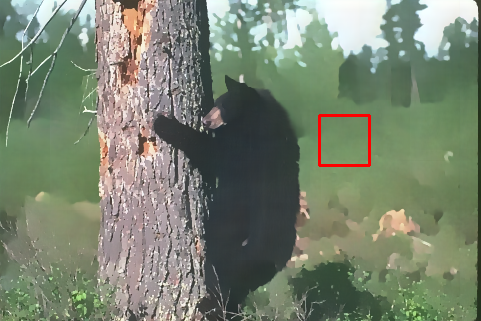}} 
\hspace{-0.1\textwidth}\subfloat{\includegraphics[width=0.1\textwidth]{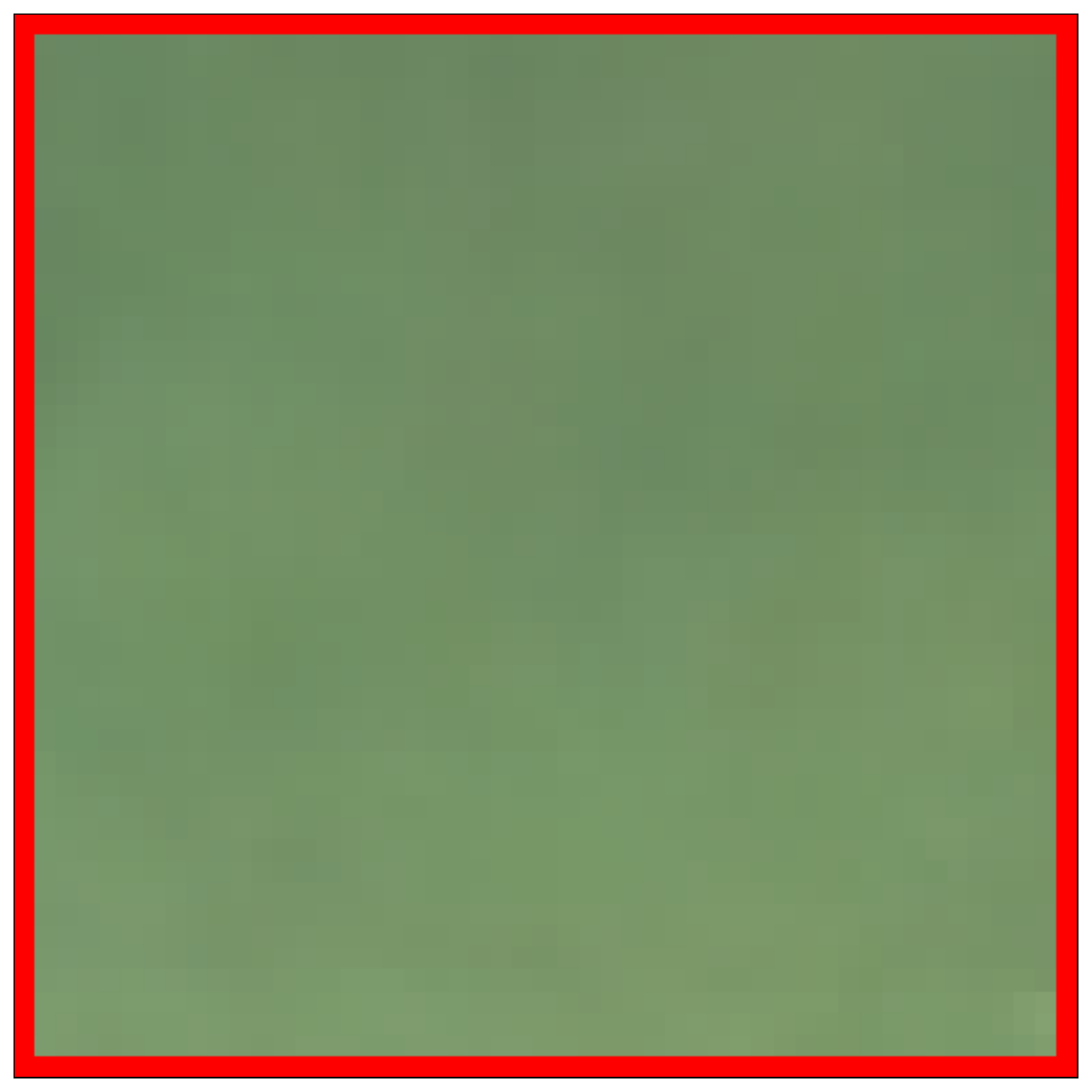}}
\addtocounter{subfigure}{-1}
\subfloat[Our PNEN]{\includegraphics[width=0.32\textwidth]{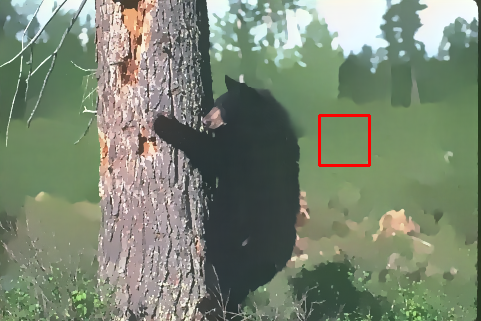}}
\hspace{-0.1\textwidth}\subfloat{\includegraphics[width=0.1\textwidth]{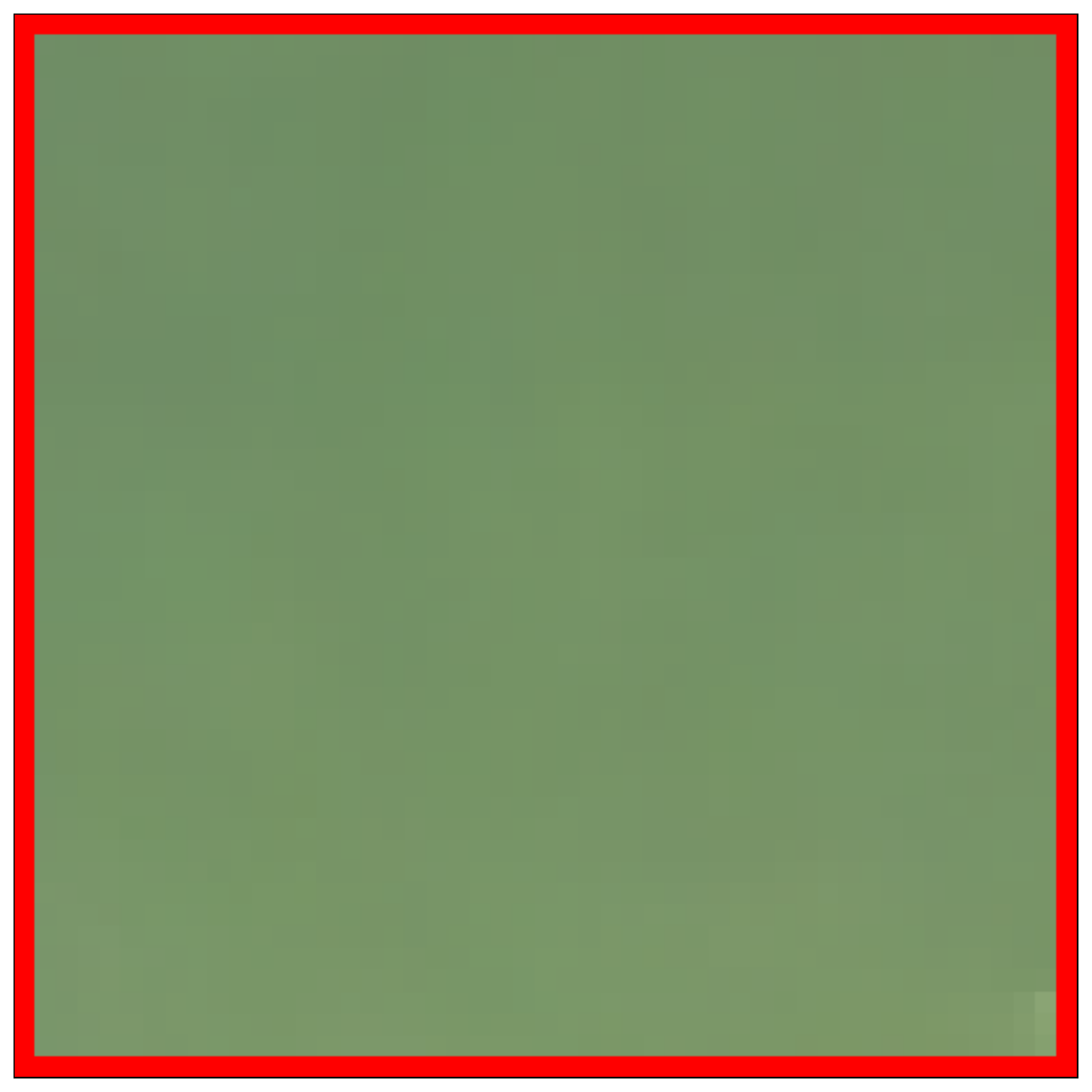}} 
\caption{Visual comparison of learning $L_0$ smoothing filter. The result produced by our proposed PNEN is clearer and flatter than results of other methods.}
\label{fig:StateOfTheArt}
\end{figure*}
\begin{figure*}[!t]
\captionsetup[subfigure]
{subrefformat=simple, listofformat=subsimple,farskip=1pt,justification=centering}
\centering
\subfloat{\includegraphics[width=0.21\textwidth]{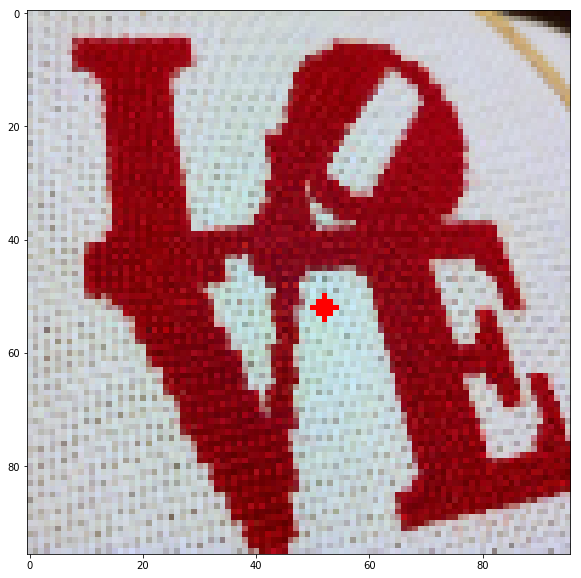}}
\subfloat{\includegraphics[width=0.24\textwidth]{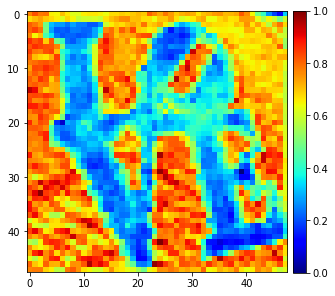}}
\subfloat{\includegraphics[width=0.24\textwidth]{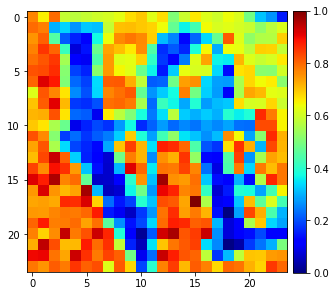}}
\subfloat{\includegraphics[width=0.24\textwidth]{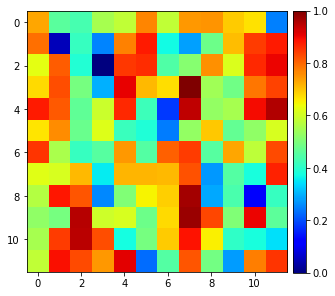}}\\
\addtocounter{subfigure}{-4}
\subfloat[Input Image]{\includegraphics[width=0.21\textwidth]{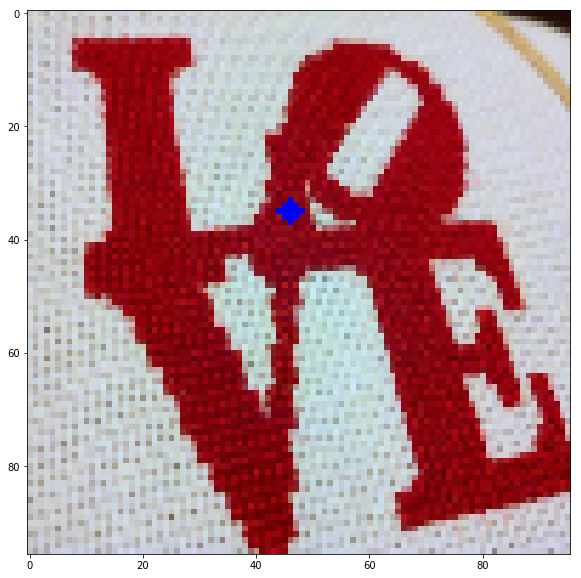}}
\subfloat[Scale $2$]{\includegraphics[width=0.24\textwidth]{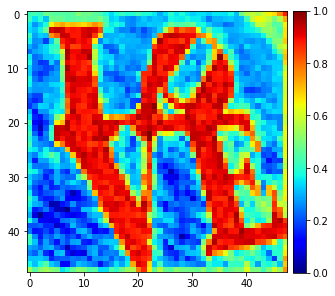}}
\subfloat[Scale $4$]{\includegraphics[width=0.24\textwidth]{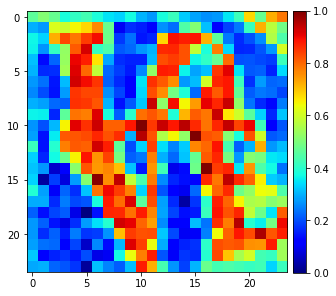}}
\subfloat[Scale $8$]{\includegraphics[width=0.24\textwidth]{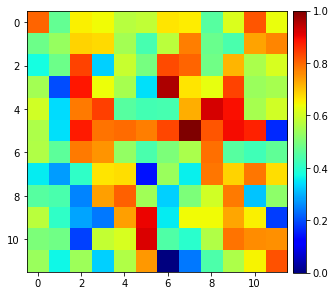}}
\caption{Correlation maps of the pyramid non-local operation at different scales. The first row shows correlation maps computed at the red point. The second row shows correlation maps computed at the blue point. }
\label{fig:CorrelationMatrix}
\end{figure*}

\subsection{Discussion}
\label{sec:Discussion}
The benefits of our proposed pyramid non-local block are three folds: 1) The pyramidal strategy adopts multiple convolutions to produce a pyramid of key and embedding features. This facilitates the correlation estimation across texels with different spatial scales.
2) In the filed of low-level image processing, most existing deep models based on non-local modules are implemented via connecting all pairs of pixels in the feature map \cite{wang2018non} or limiting the nonlocal dependencies within a constant neighborhood size \cite{liu2018non}. 
The former kind of method merely plugs non-local modules after high-level feature maps with small resolution, because of limited memory resource. The later kind of method inevitably neglects valuable correlations from pixels outside the fixed neighborhood.  We solve the problem ingeniously through embedding the input feature into a query feature map with full resolution and multiple reference feature maps with downscaled resolutions. In such a manner, the computation burden can be greatly relieved without reducing the resolution of feature representations.
3) The pyramid non-local block can be easily incorporated into existing CNN-based models proposed for other low-level image processing tasks, such as MemNet \cite{Tai-MemNet-2017} and RDN\cite{zhang2020residual}.

\begin{figure*}[t]
\captionsetup[subfigure]{labelformat=empty,farskip=1pt}
\centering
\subfloat[Ground Truth]{\includegraphics[width=0.23\textwidth]{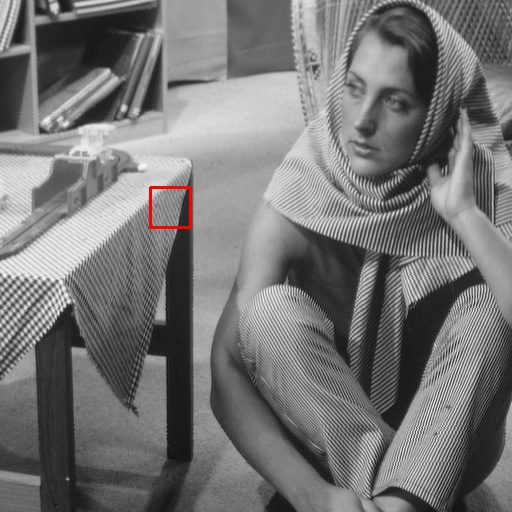}}
\hspace{-0.1\textwidth}\subfloat{\includegraphics[width=0.1\textwidth]{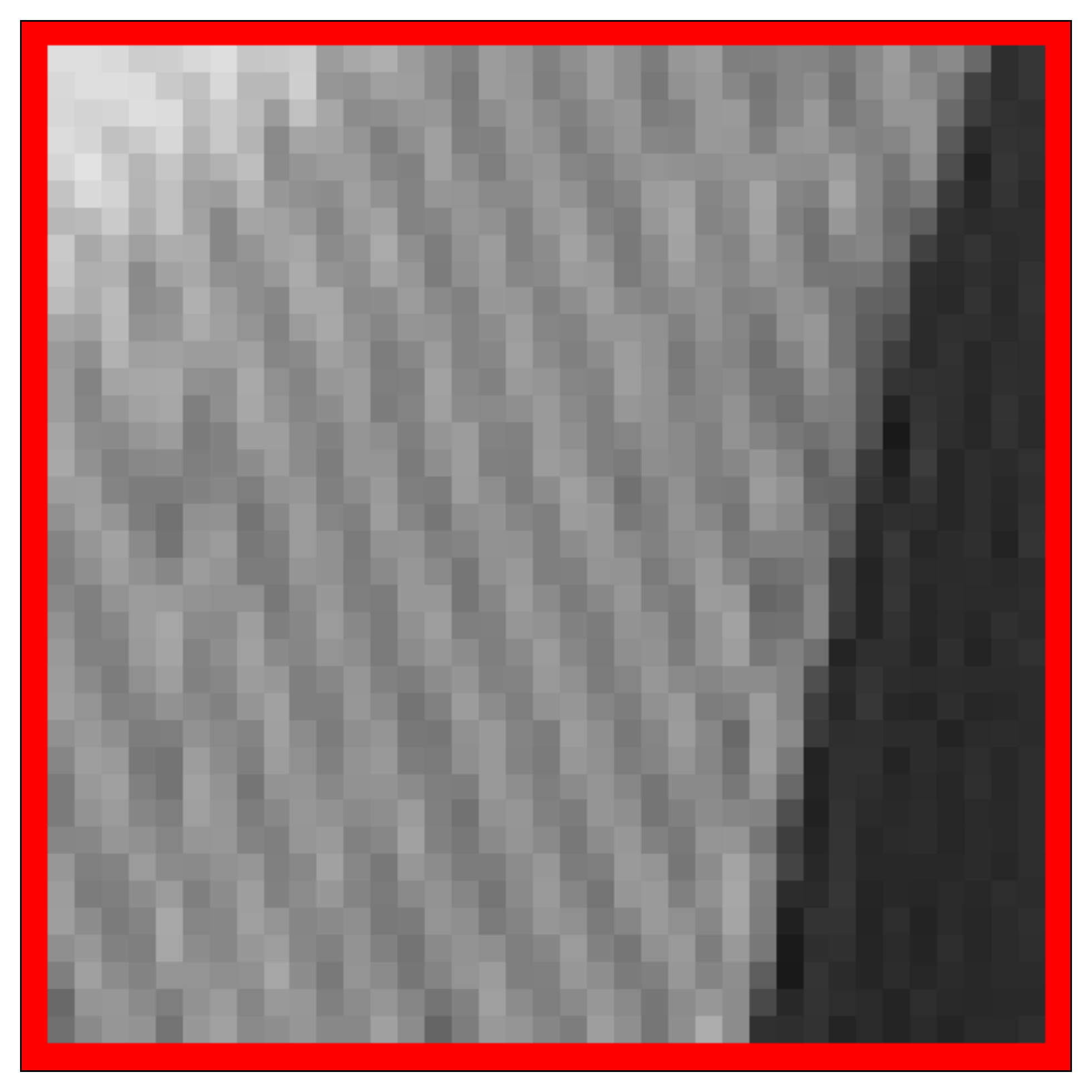}}~
\subfloat[Noisy $\sigma=30$]{\includegraphics[width=0.23\textwidth]{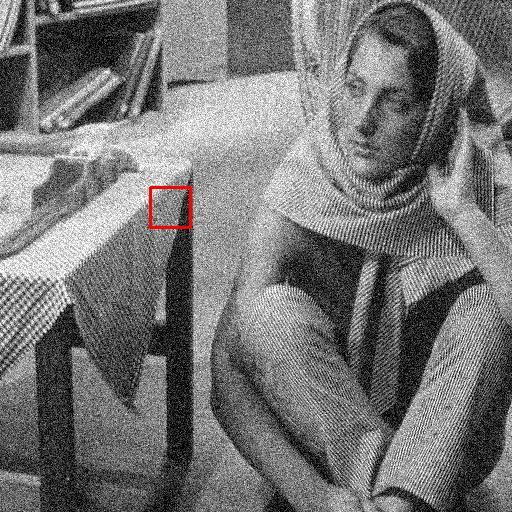}}
\hspace{-0.1\textwidth}\subfloat{\includegraphics[width=0.1\textwidth]{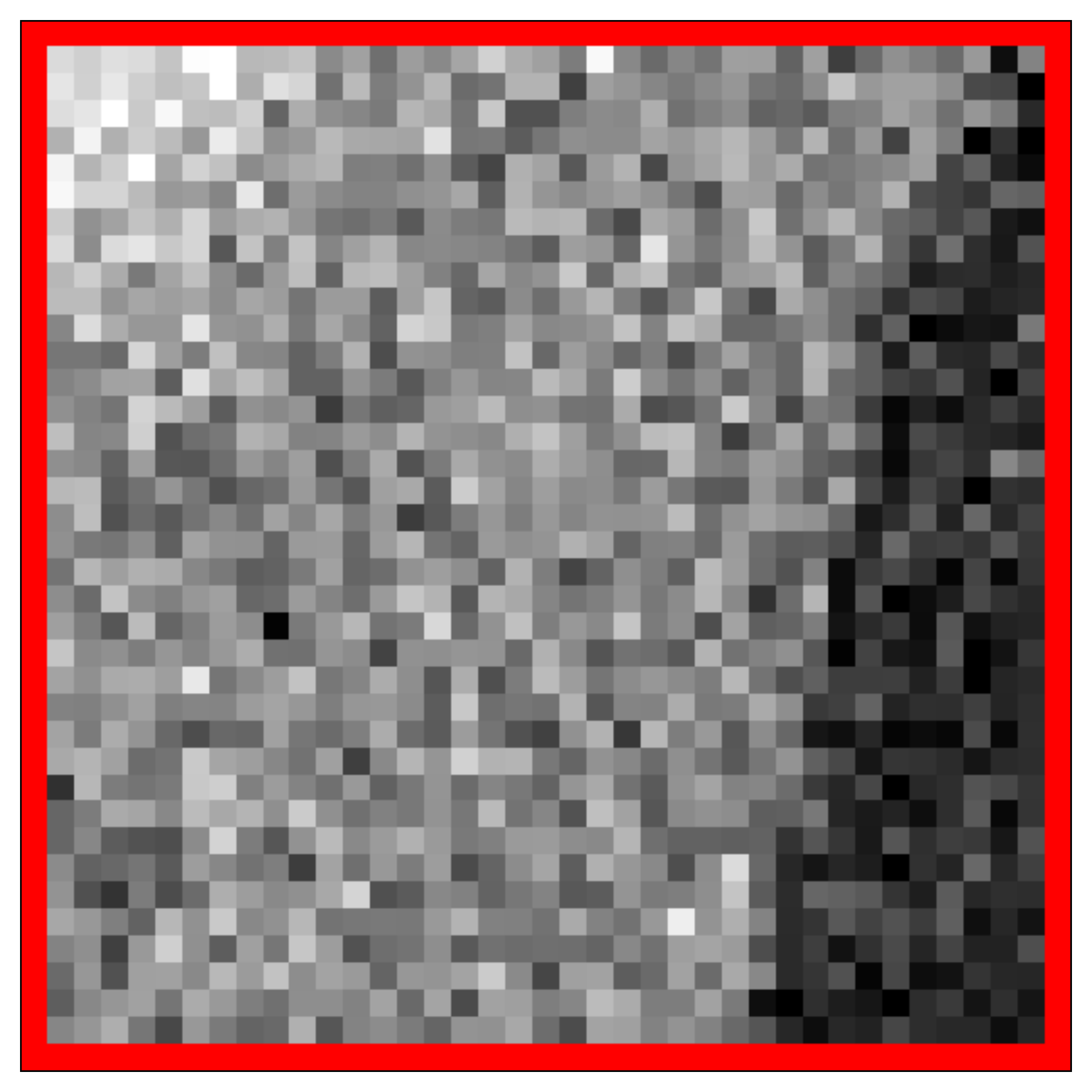}}~
\subfloat[RDN]{\includegraphics[width=0.23\textwidth]{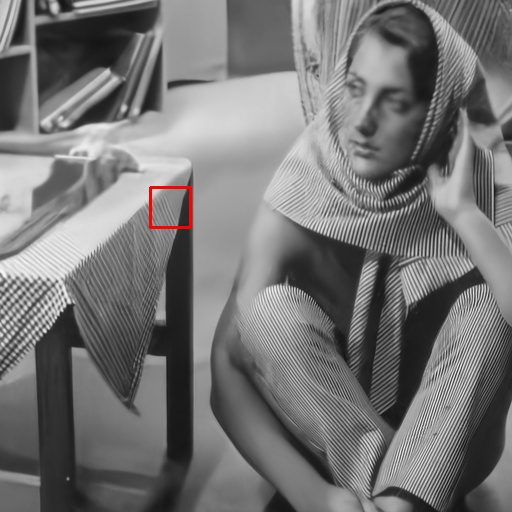}}
\hspace{-0.1\textwidth}\subfloat{\includegraphics[width=0.1\textwidth]{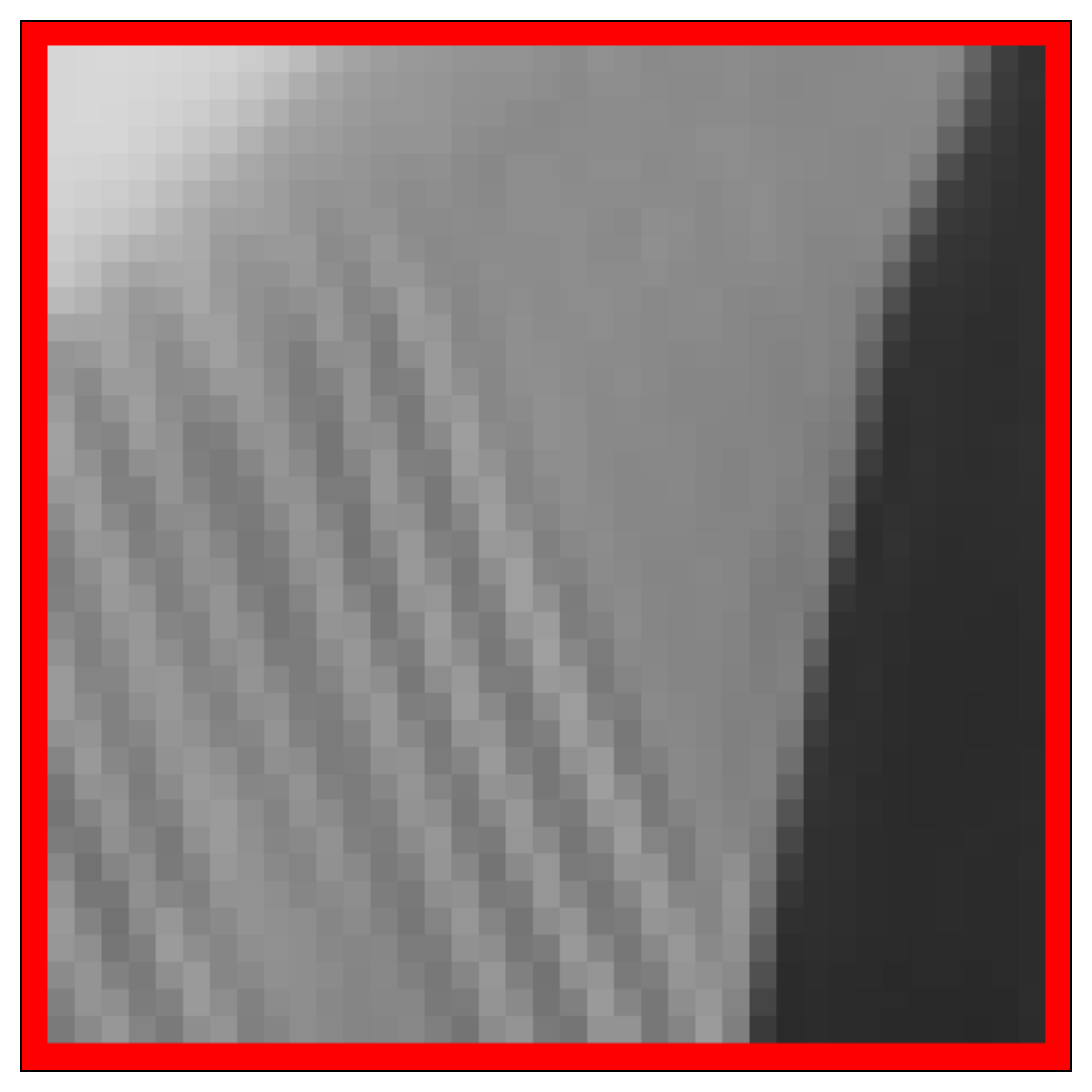}}~
\subfloat[RDN(w/ PNB)]{\includegraphics[width=0.23\textwidth]{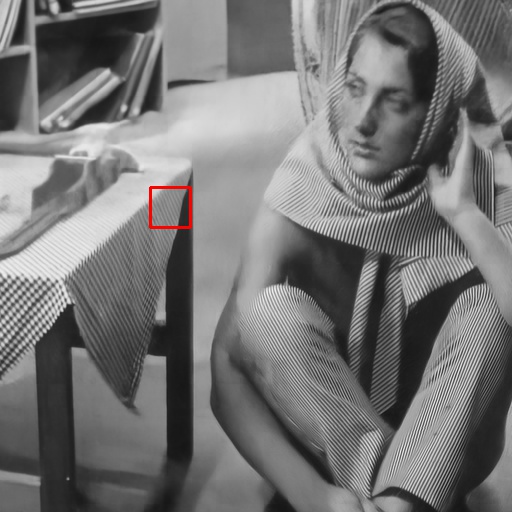}}
\hspace{-0.1\textwidth}\subfloat{\includegraphics[width=0.1\textwidth]{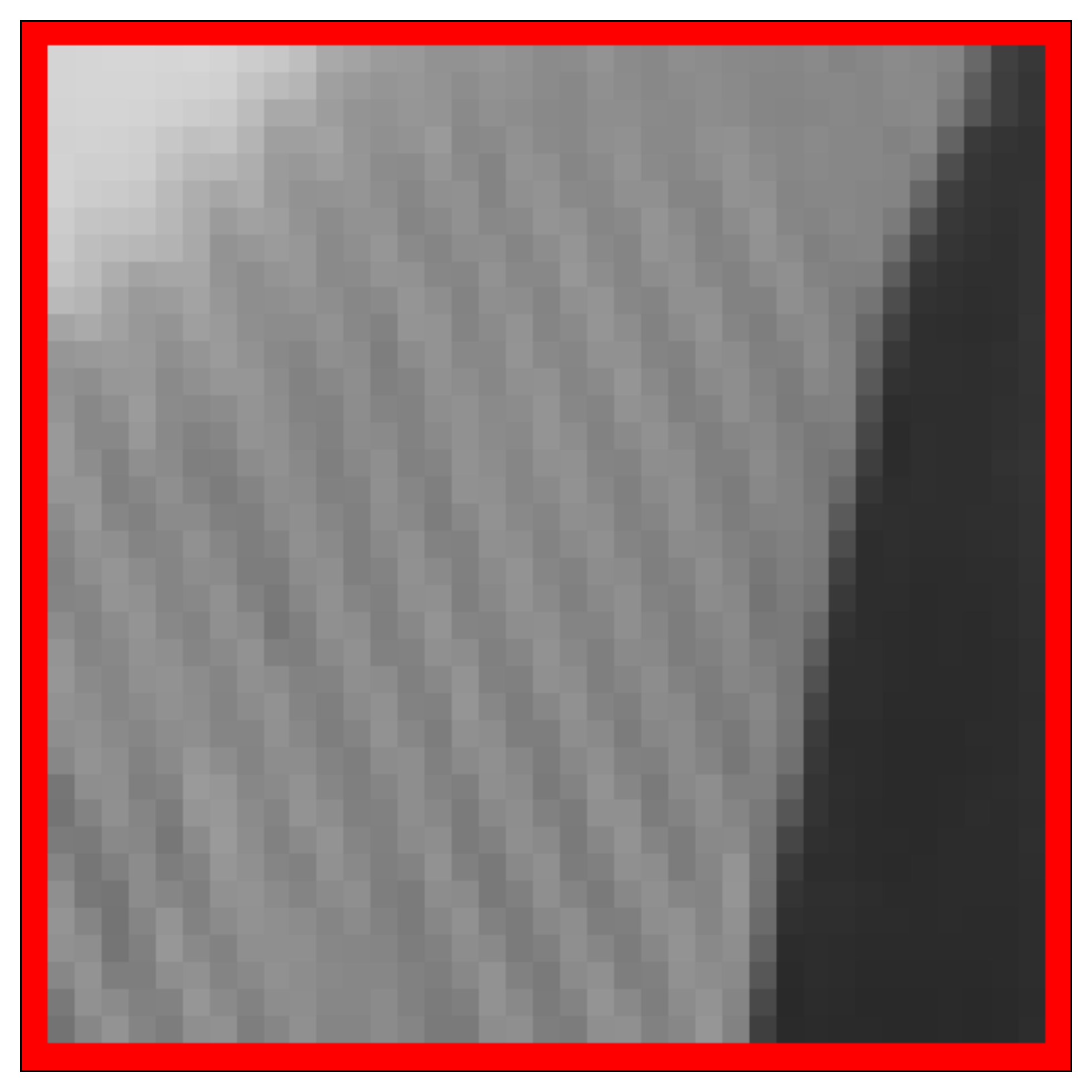}}\
\subfloat{\includegraphics[width=0.23\textwidth]{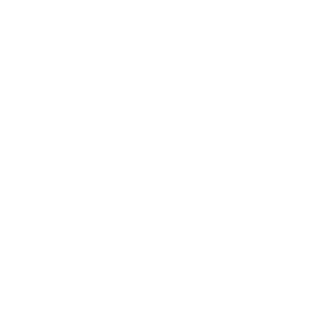}}
\subfloat[Noisy $\sigma=50$]{\includegraphics[width=0.23\textwidth]{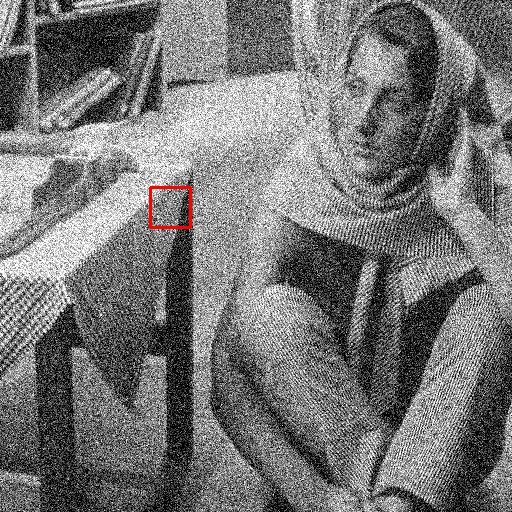}}
\hspace{-0.1\textwidth}\subfloat{\includegraphics[width=0.1\textwidth]{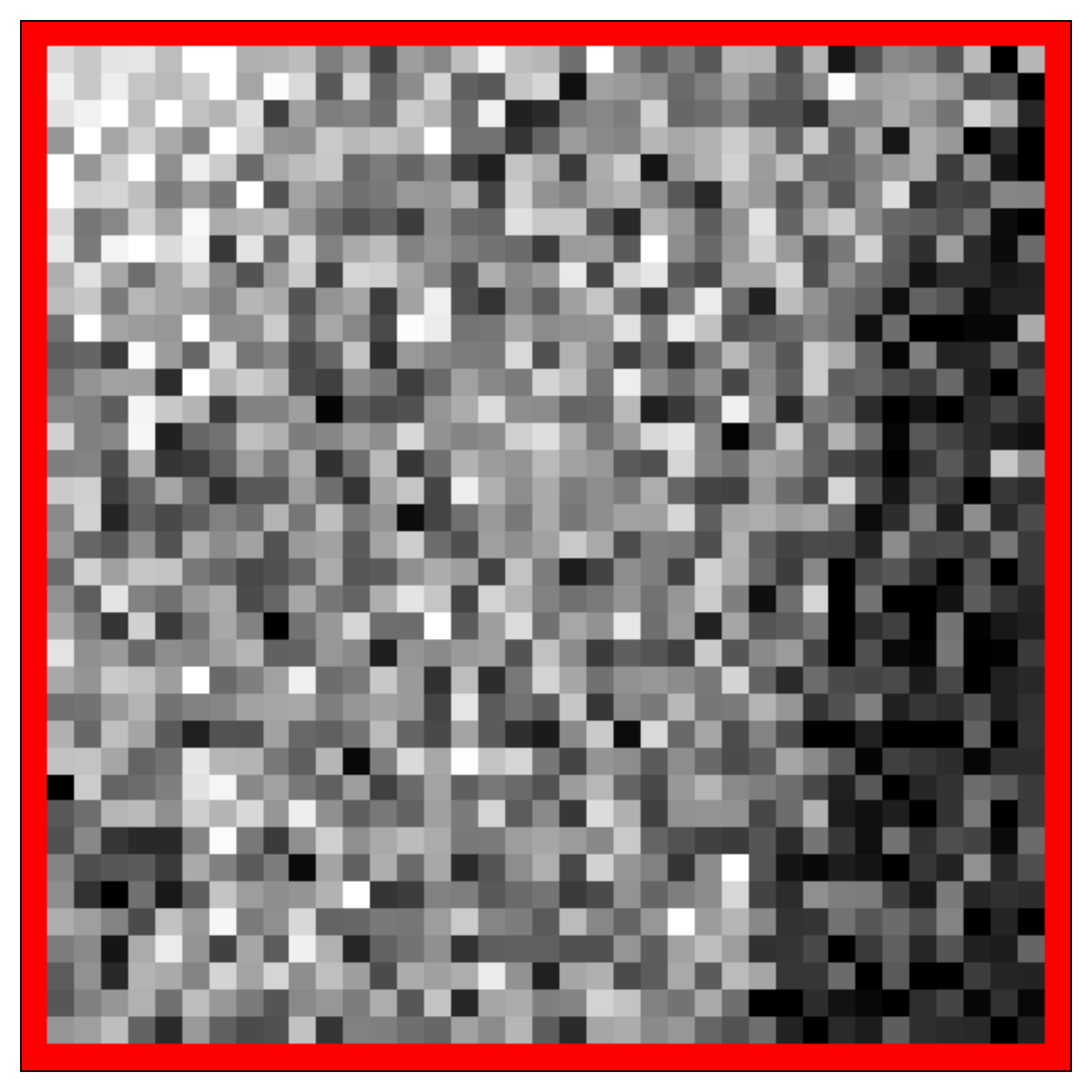}}~
\subfloat[RDN]{\includegraphics[width=0.23\textwidth]{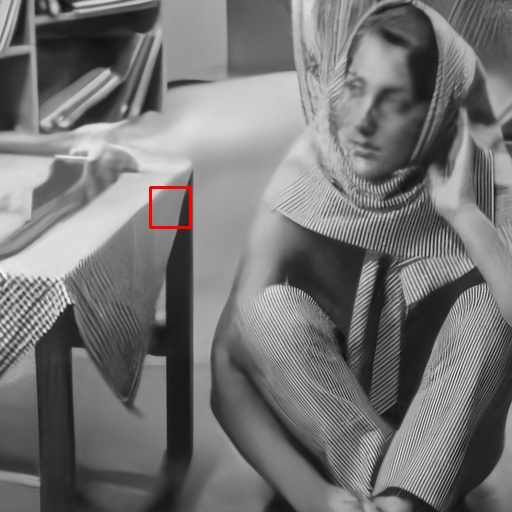}}
\hspace{-0.1\textwidth}\subfloat{\includegraphics[width=0.1\textwidth]{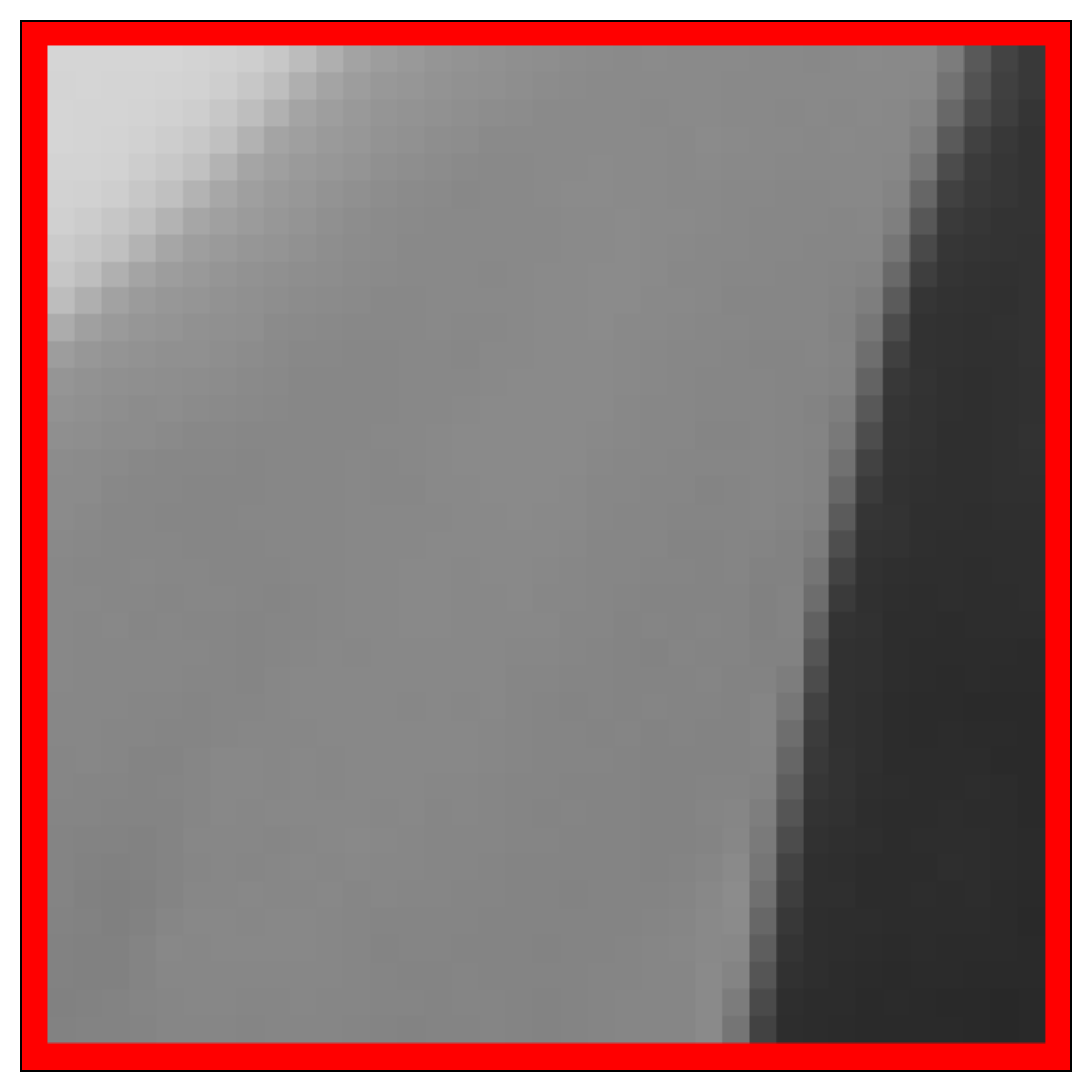}}~
\subfloat[RDN(w/ PNB)]{\includegraphics[width=0.23\textwidth]{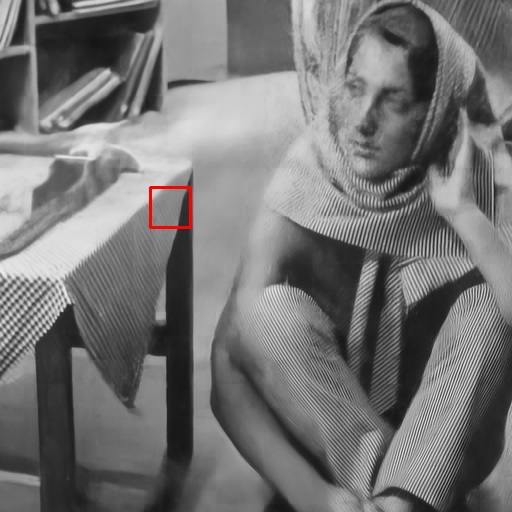}}
\hspace{-0.1\textwidth}\subfloat{\includegraphics[width=0.1\textwidth]{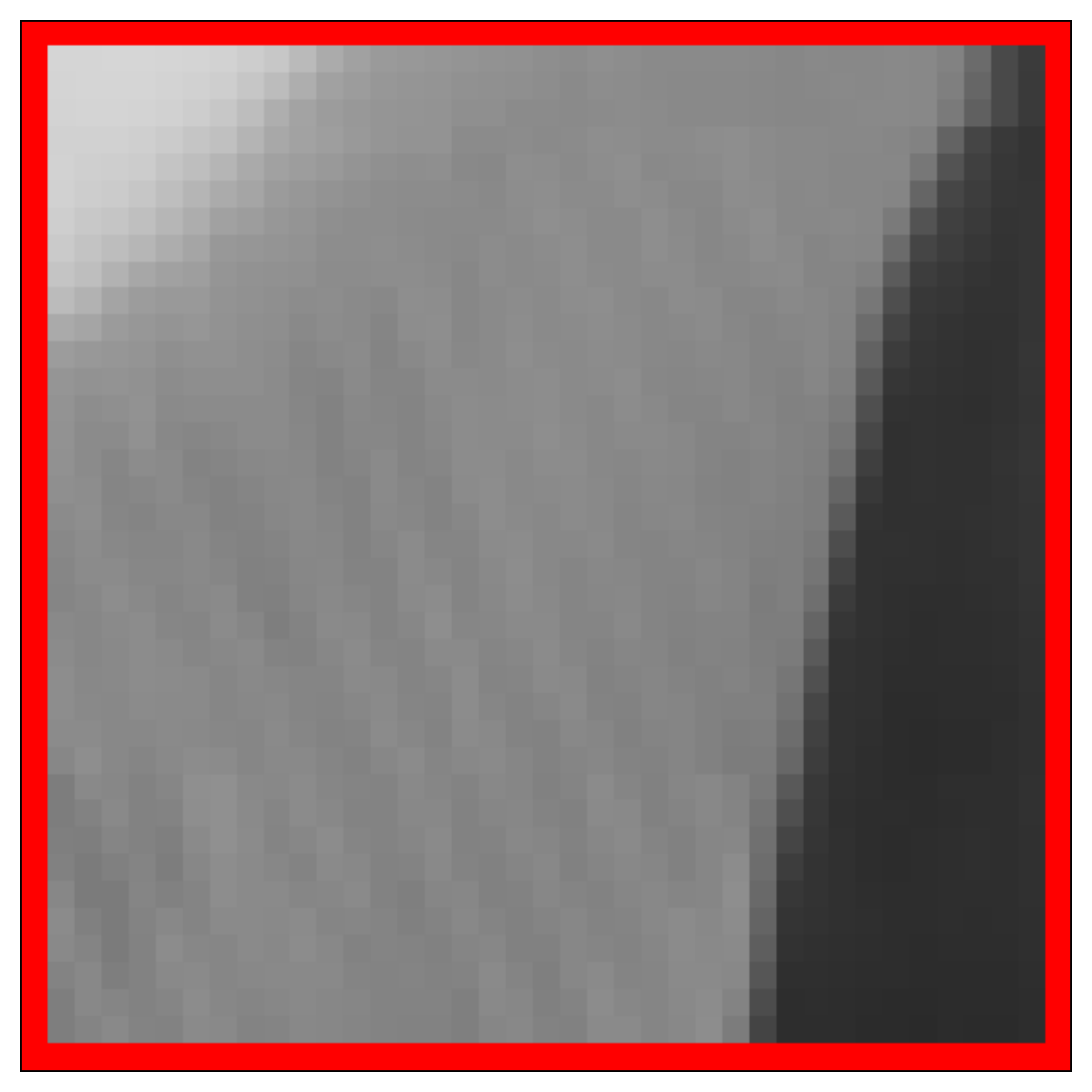}}\
\centering
\subfloat[Ground Truth]{\includegraphics[width=0.23\textwidth]{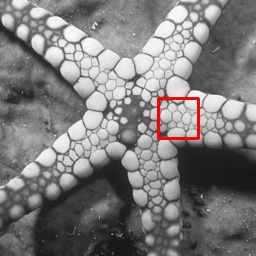}}
\hspace{-0.1\textwidth}\subfloat{\includegraphics[width=0.1\textwidth]{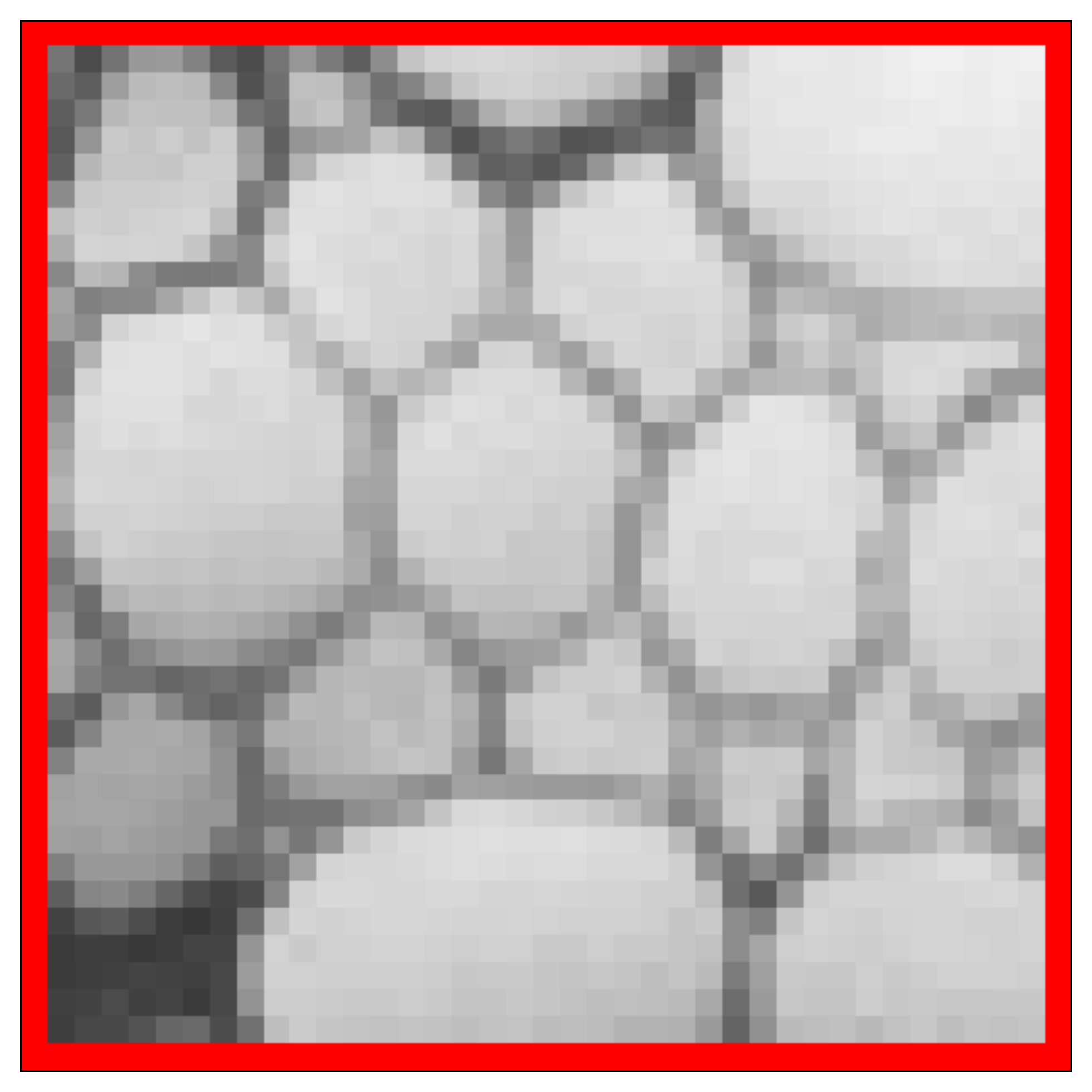}}~
\subfloat[Noisy $\sigma=30$]{\includegraphics[width=0.23\textwidth]{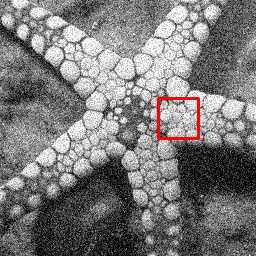}}
\hspace{-0.1\textwidth}\subfloat{\includegraphics[width=0.1\textwidth]{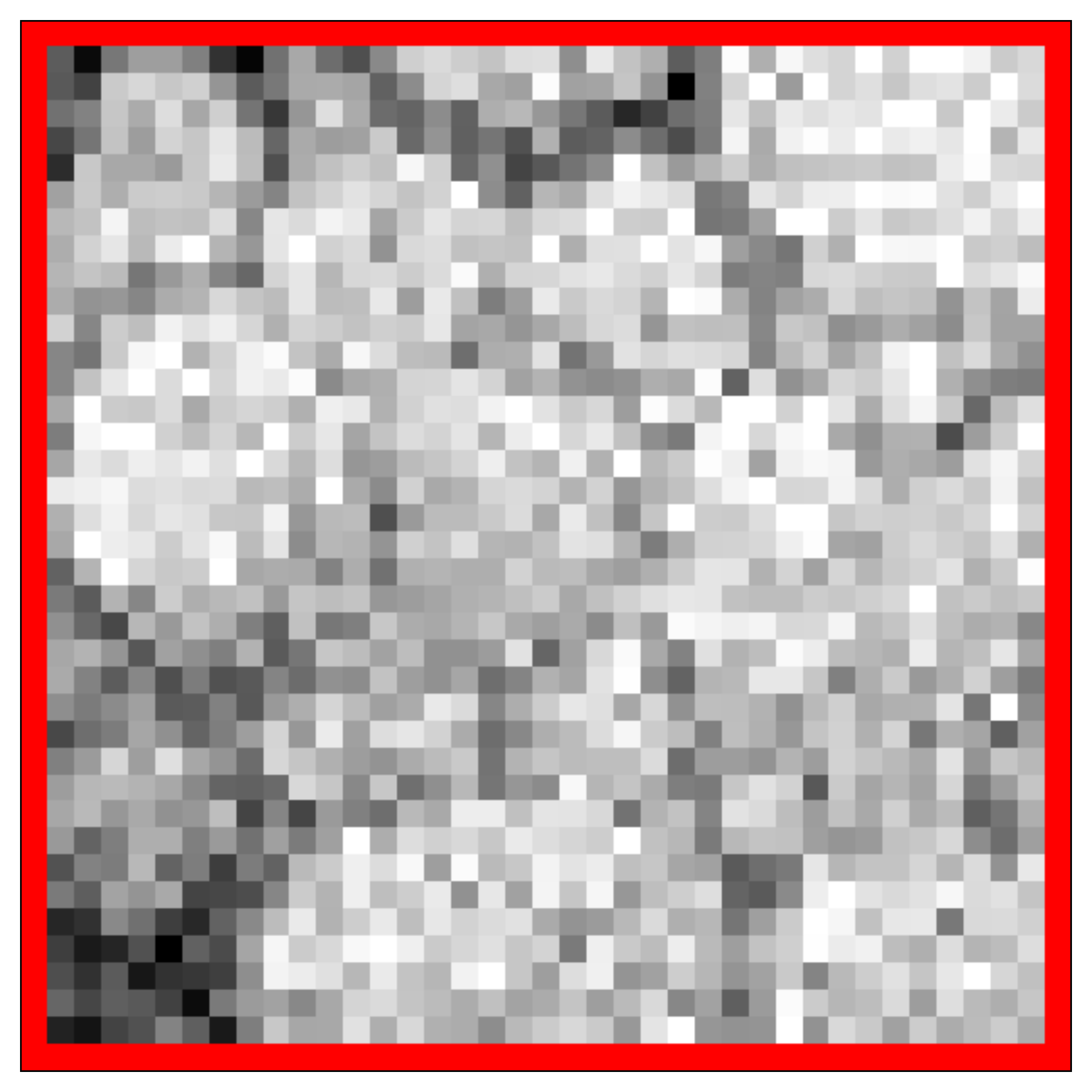}}~
\subfloat[RDN]{\includegraphics[width=0.23\textwidth]{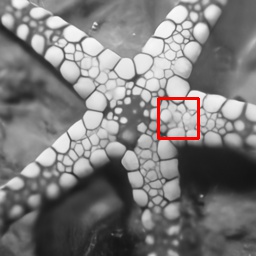}}
\hspace{-0.1\textwidth}\subfloat{\includegraphics[width=0.1\textwidth]{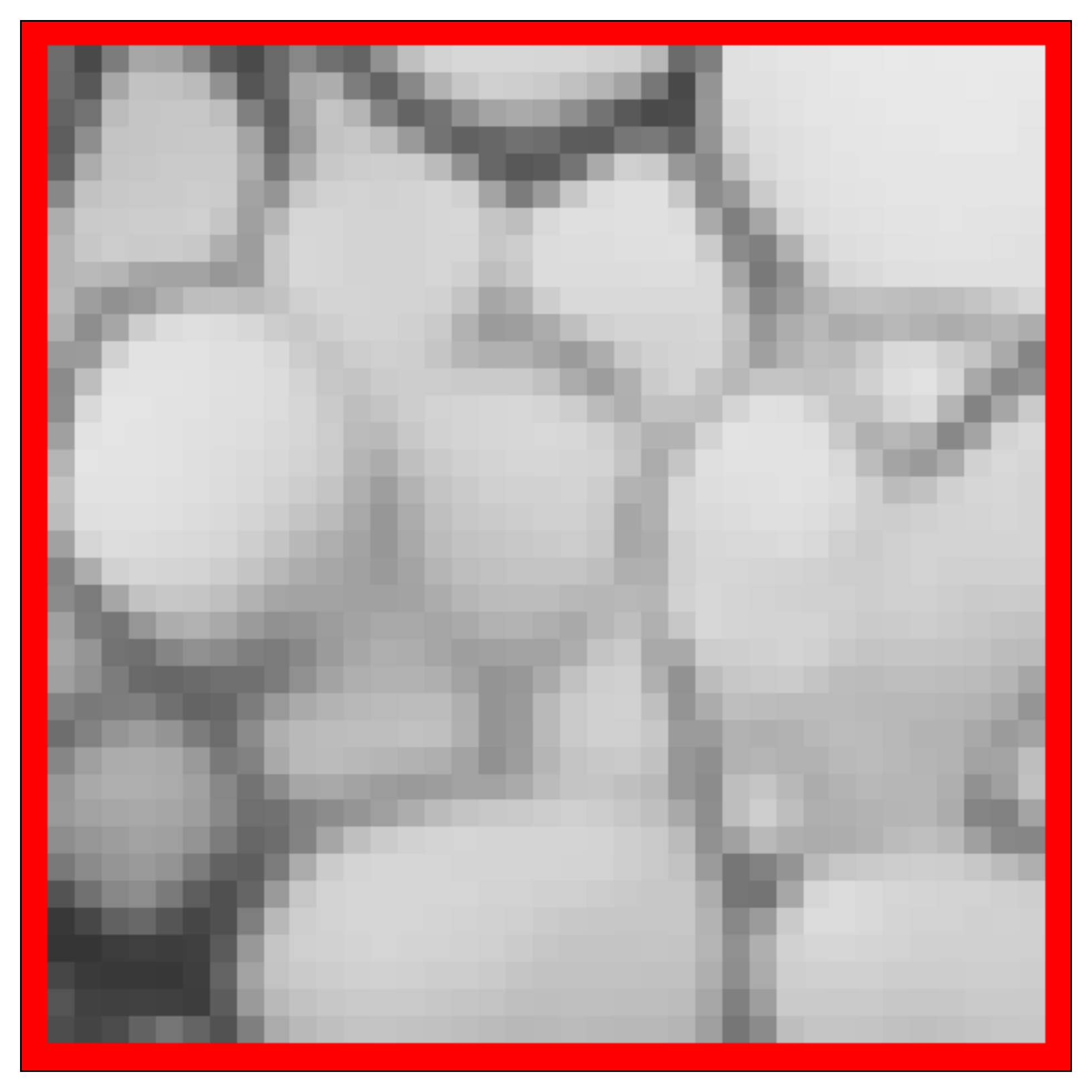}}~
\subfloat[RDN(w/ PNB)]{\includegraphics[width=0.23\textwidth]{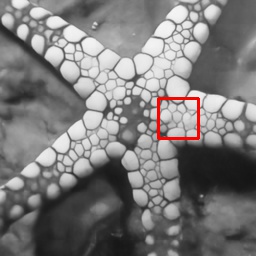}}
\hspace{-0.1\textwidth}\subfloat{\includegraphics[width=0.1\textwidth]{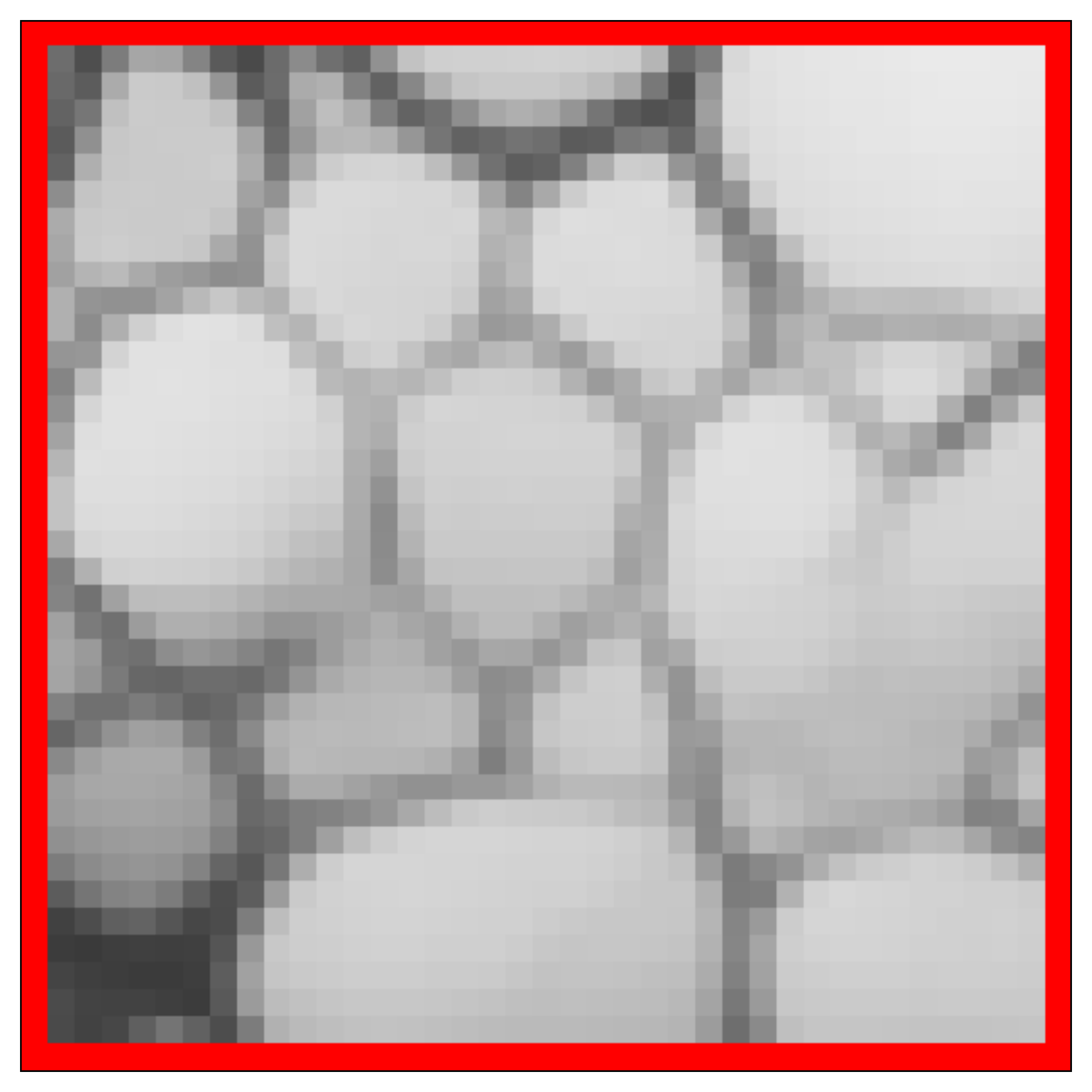}}\
\subfloat{\includegraphics[width=0.23\textwidth]{figures/blank.png}}
\subfloat[Noisy $\sigma=50$]{\includegraphics[width=0.23\textwidth]{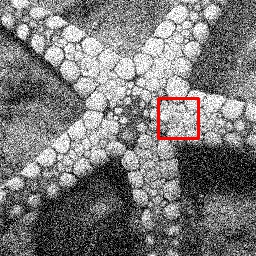}}
\hspace{-0.1\textwidth}\subfloat{\includegraphics[width=0.1\textwidth]{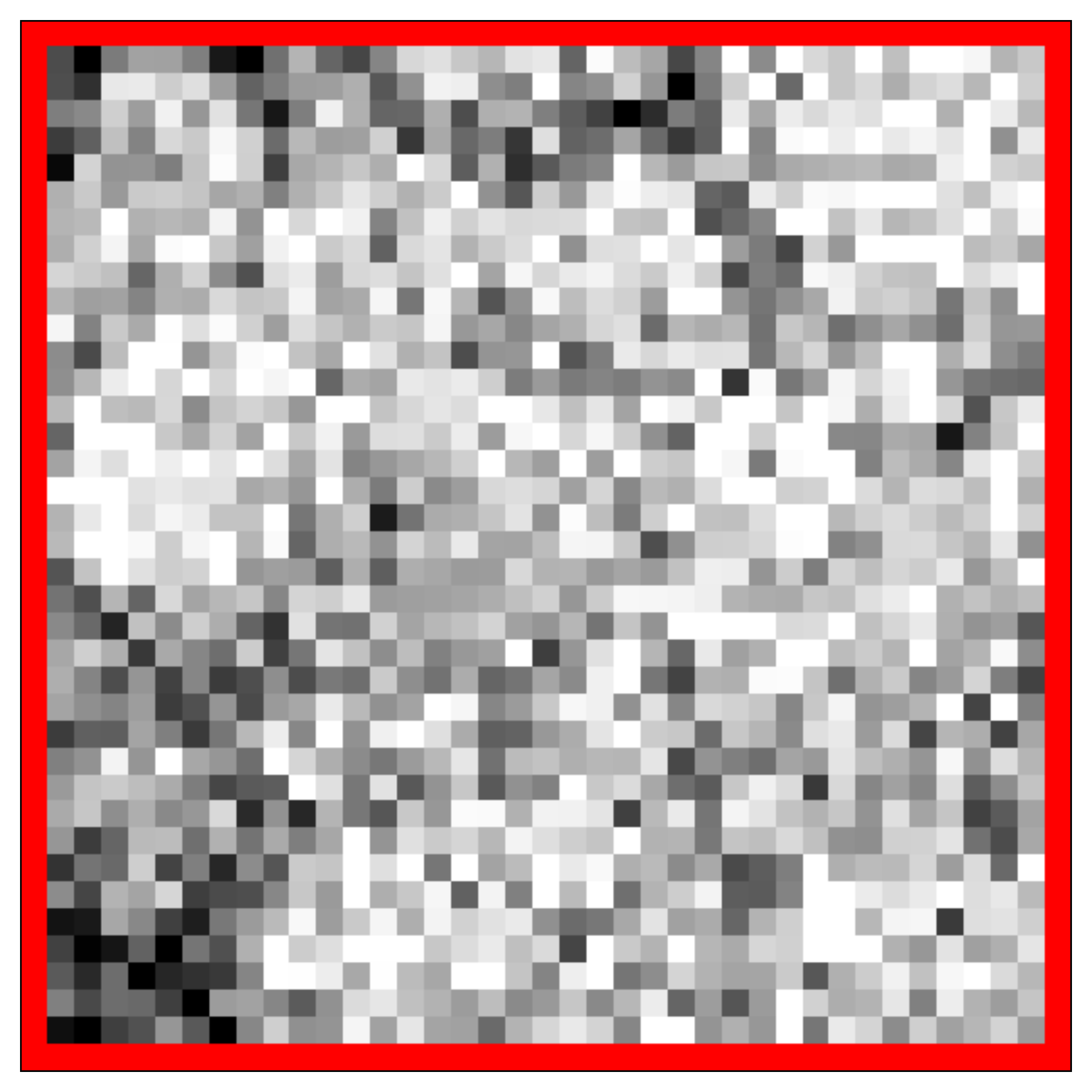}}~
\subfloat[RDN]{\includegraphics[width=0.23\textwidth]{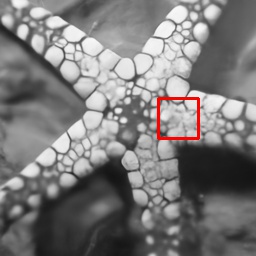}}
\hspace{-0.1\textwidth}\subfloat{\includegraphics[width=0.1\textwidth]{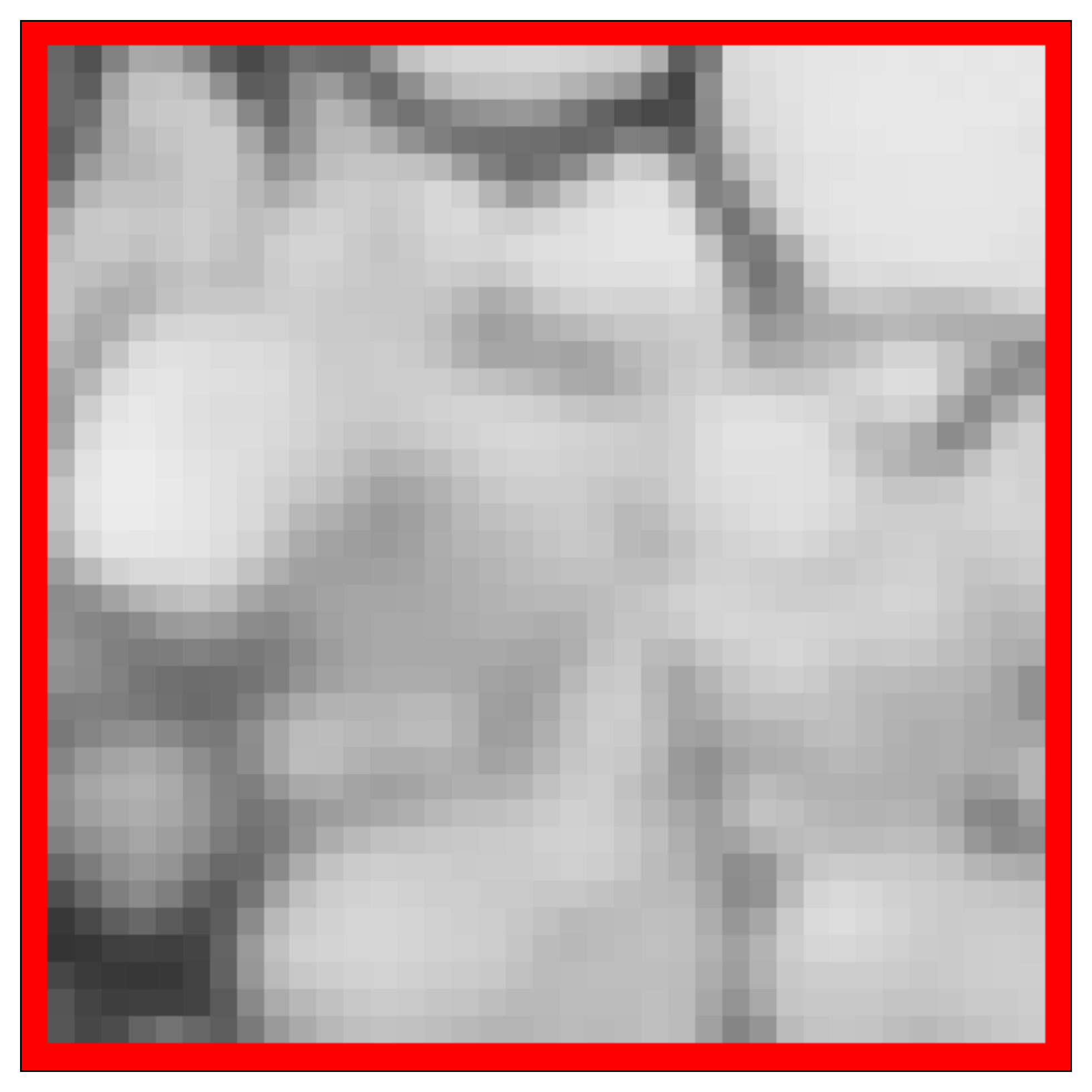}}~
\subfloat[RDN(w/ PNB)]{\includegraphics[width=0.23\textwidth]{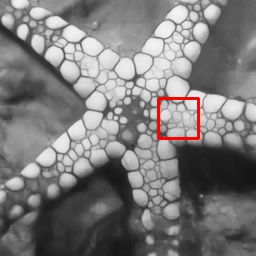}}
\hspace{-0.1\textwidth}\subfloat{\includegraphics[width=0.1\textwidth]{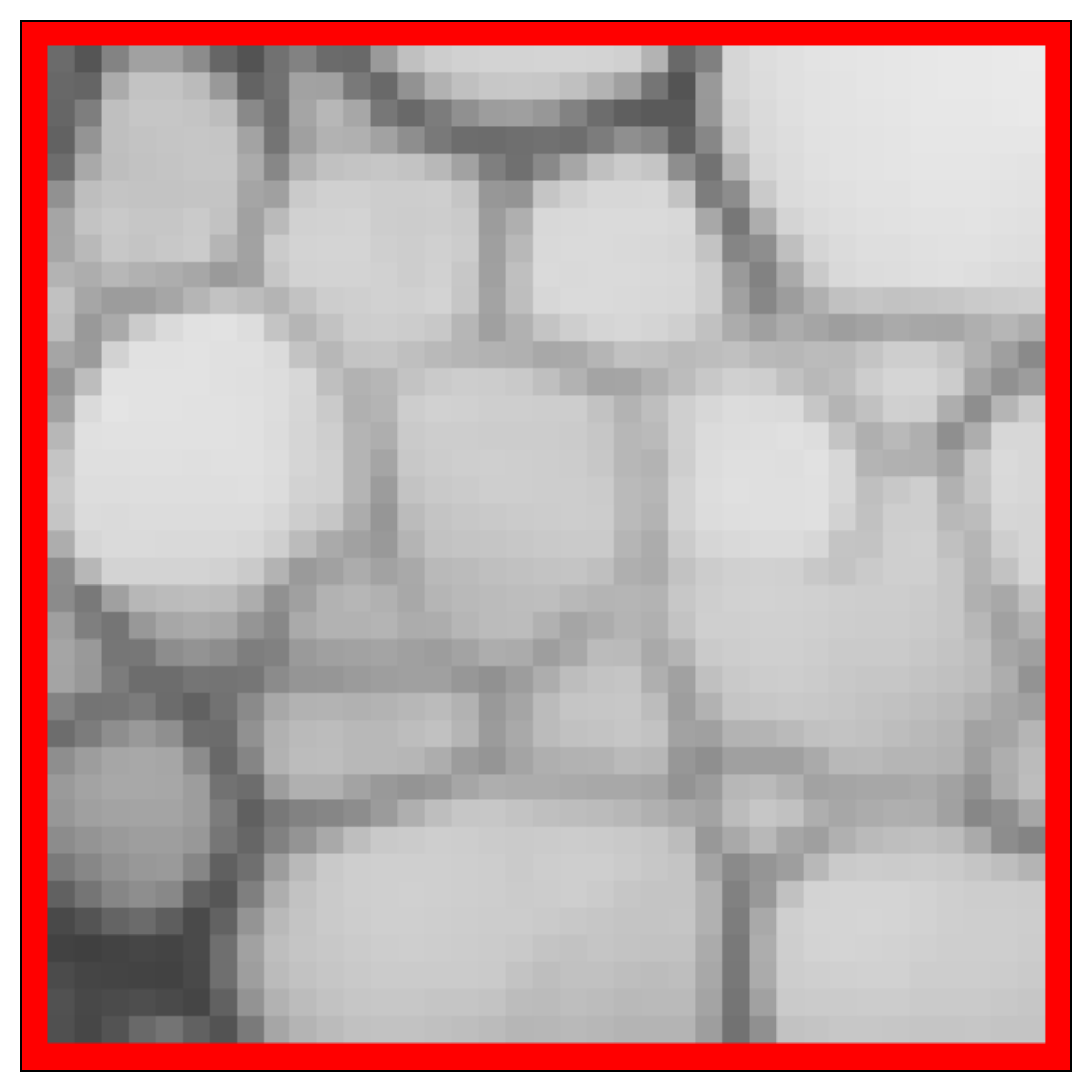}}\
\caption{Visual comparison of image denoising with noise level $\sigma=30$ and $\sigma=50$. From left to right: the clean images, the noisy images synthesized via adding white Gaussian noise, results produced by baseline RDN \cite{zhang2020residual} and results produced by PNB-enhanced RDN.}
\label{fig:denoising}
\end{figure*}

\section{Experiments in Edge-preserving Smoothing}

Edge-preserving smoothing is a fundamental topic in image processing. It should preserve major image structures while neglecting insignificant details, which is critical in many computer vision tasks such as image segmentation and contour detection.

\subsection{Dataset}
\label{dataset}

The dataset in \cite{zhu2019benchmark}, consisting of 500 images of clear structures and visible details, are used to validate image smoothing algorithms. These images are split into 400 for training and 100 for testing. We train models to reproduce three representative filters in our experiments, including weighted median filter $(r=10, \sigma=50)$ \cite{zhang2014100+}, $L_0$ smoothing $(\lambda=0.02,\kappa=2)$ \cite{xu2011image} and SD Filter $(\lambda=15)$ \cite{ham2017robust}.

\noindent \textbf{Implementation Details}  Without specification, all convolution layers have $64$ filters with kernel size $3 \times 3$. We stack three PNB-s and DRB-s consecutively as the feature extractor, resulting in a 37-layer deep network. During the training
stage, random horizontal flip and rotation are applied for data augmentation. Training images are decomposed into $96 \times 96$ patches. The mini-batch size is set to 8. The Adam optimizer \cite{kingma2014adam} with $\beta_1=0.9, \beta_2=0.999, \epsilon=10^{-8}$ is used for optimization. The initial learning rate is set to $5\times10^{-4}$ and reduced by half when the training loss stops decreasing, until it is reduced to $10^{-4}$. It takes around 2 days to train a model on one TITAN Xp GPU. For inference, the model takes 1.2 seconds to process a testing image with $500 \times 400$ pixels. 

\begin{table}[!t]
\centering
\caption{Quantitative comparison in terms of PSNR/SSIM in reproducing three image smoothing methods. Network depth and number of parameters are also indicated.}
\label{table:state_of_the_art}
\resizebox{0.5\textwidth}{!}
{
\renewcommand{\arraystretch}{1.2}
\begin{tabular}{c|cccc}\specialrule{.1em}{0em}{0em}
              & DJF \cite{li2016deep} & CEILNet \cite{fan2017generic} & ResNet \cite{zhu2019benchmark} & PNEN  \\ \hline 
WMF           & 34.31/0.9647 & 37.73/0.9773 & 38.30/0.9813 & \textbf{39.45/0.9846}     \\ 
$L_0$            & 30.20/0.9458 & 31.30/0.9519 & 32.30/0.9671 & \textbf{33.44/0.9741}     \\ 
SD Filter     & 30.95/0.9264 & 32.67/0.9452 & 33.21/0.9532 & \textbf{34.19/0.9646}     \\ \hline
Max depth     & 6            & 32           & 37           & 37    \\
\#Params      & 99k          & 1113k        & 1961k        & 1875k \\ \specialrule{.1em}{0em}{0em}
\end{tabular}
}
\end{table}

\subsection{Comparison with the State-of-the-art}
Piles of CNN-based approaches \cite{xu2015deep,liu2016learning,li2016deep,fan2017generic,zhu2019benchmark} have been proposed to reproduce edge-preserving smoothing filters. We compare our proposed PNEN against three state-of-the-art methods, including Deep Joint Filter (DJF) \cite{li2016deep},  Cascaded Edge and Image Learning Network (CEILNet) \cite{fan2017generic} and Residual Networks (ResNet) \cite{zhu2019benchmark}. For fair comparison, the networks are re-trained from scratch on the same training dataset as described above. We evaluate the quality of the generated images using two metrics, including Peak Signal-to-Noise Ratio (PSNR) and Structure Similarity Index (SSIM) \cite{wang2004image}.

Quantitative results are reported in Table \ref{table:state_of_the_art}. When reproducing WMF, $L_0$ smoothing and SD Filter, our proposed method (PNEN) outperforms the second best method ResNet \cite{zhu2019benchmark} by $1.15$dB, $1.14$dB and $0.98$dB in PSNR metric respectively. Higher SSIM values also indicate that our method can give rise to smoothed images with better structural information. A visual comparison of learning $L_0$ smoothing filter is provided in Fig. \ref{fig:StateOfTheArt}. As can be seen in close-ups of the selected patch (inside the red box), the region smoothed by our proposed PNEN appears to be cleaner and flatter than results of other methods. Our result is closer to the ground-truth image.

As shown in Fig. \ref{fig:CorrelationMatrix}, we visualize the similarity maps derived from the last pyramid non-local block at two locations, marked by red and blue respectively. Fig. \ref{fig:CorrelationMatrix} (b), (c) and (d) show the similarity map at `Scale $2$', `Scale $4$' and `Scale $8$', respectively. We can see that pixels with similar features show high correlation in these maps. Thus, the pixelwise feature representation is non-locally enhanced by exploiting long-range dependencies. Particularly, for the pixel marked by red, there exist noises when estimating correlation coefficients with similar texels (left-bottom area) in `Scale 2' while the estimation in `Scale 4' performs well on these texels. It indicates that the adoption of pyramid non-local operations can benefit the robustness of estimating correlations between different scales of texels.

\begin{figure*}[t]
\captionsetup[subfigure]{labelformat=empty,farskip=1pt}
\centering
\subfloat[Original Image]{\includegraphics[width=0.24\textwidth]{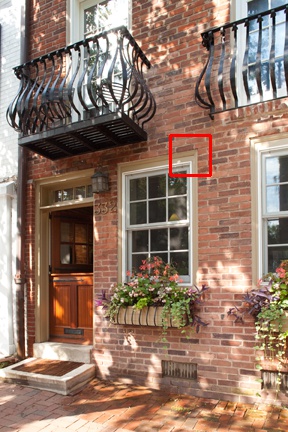}}
\hspace{-0.11\textwidth}\subfloat{\includegraphics[width=0.11\textwidth]{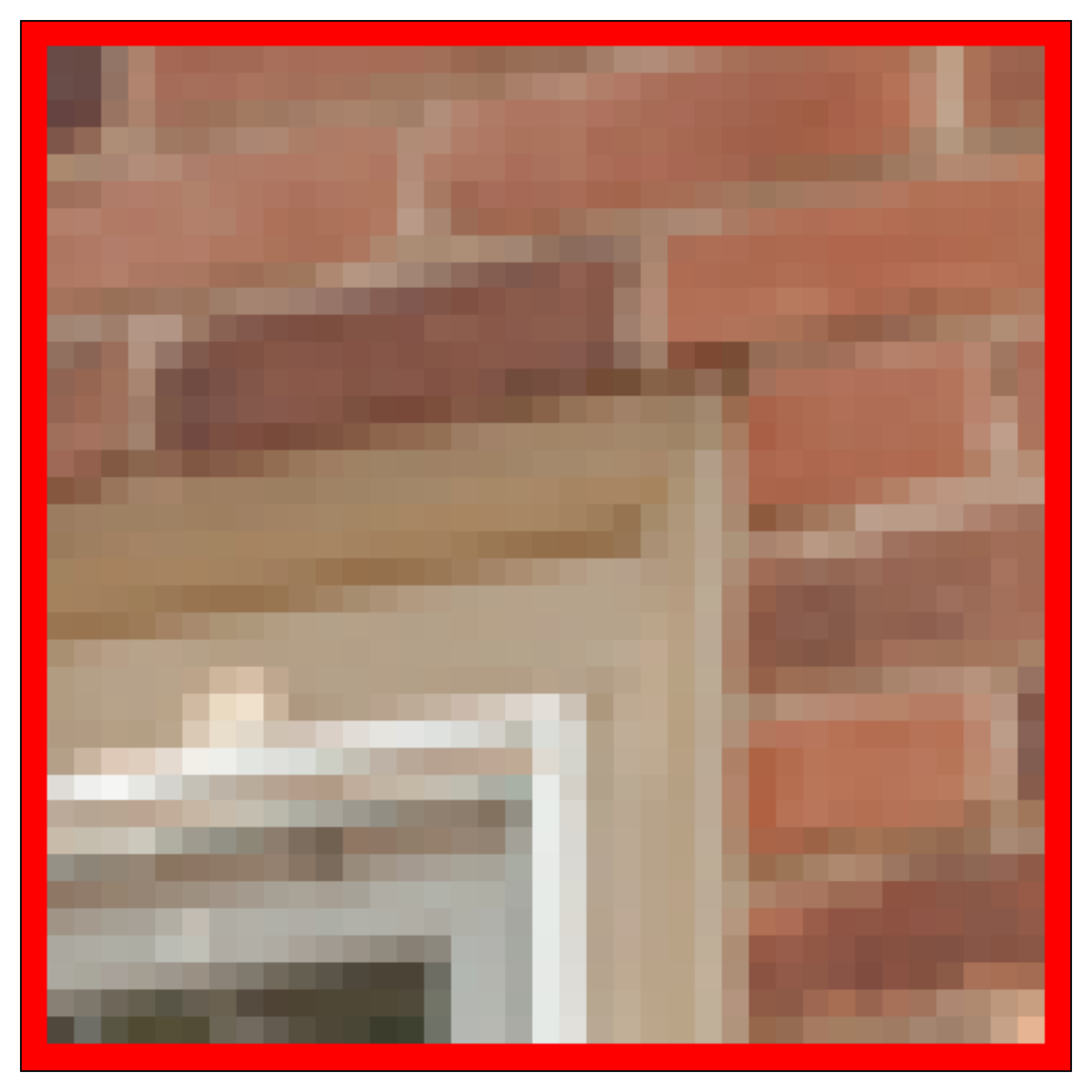}}~
\subfloat[Ground Truth]{\includegraphics[width=0.24\textwidth]{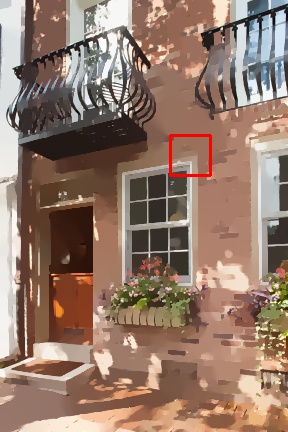}}
\hspace{-0.11\textwidth}\subfloat{\includegraphics[width=0.11\textwidth]{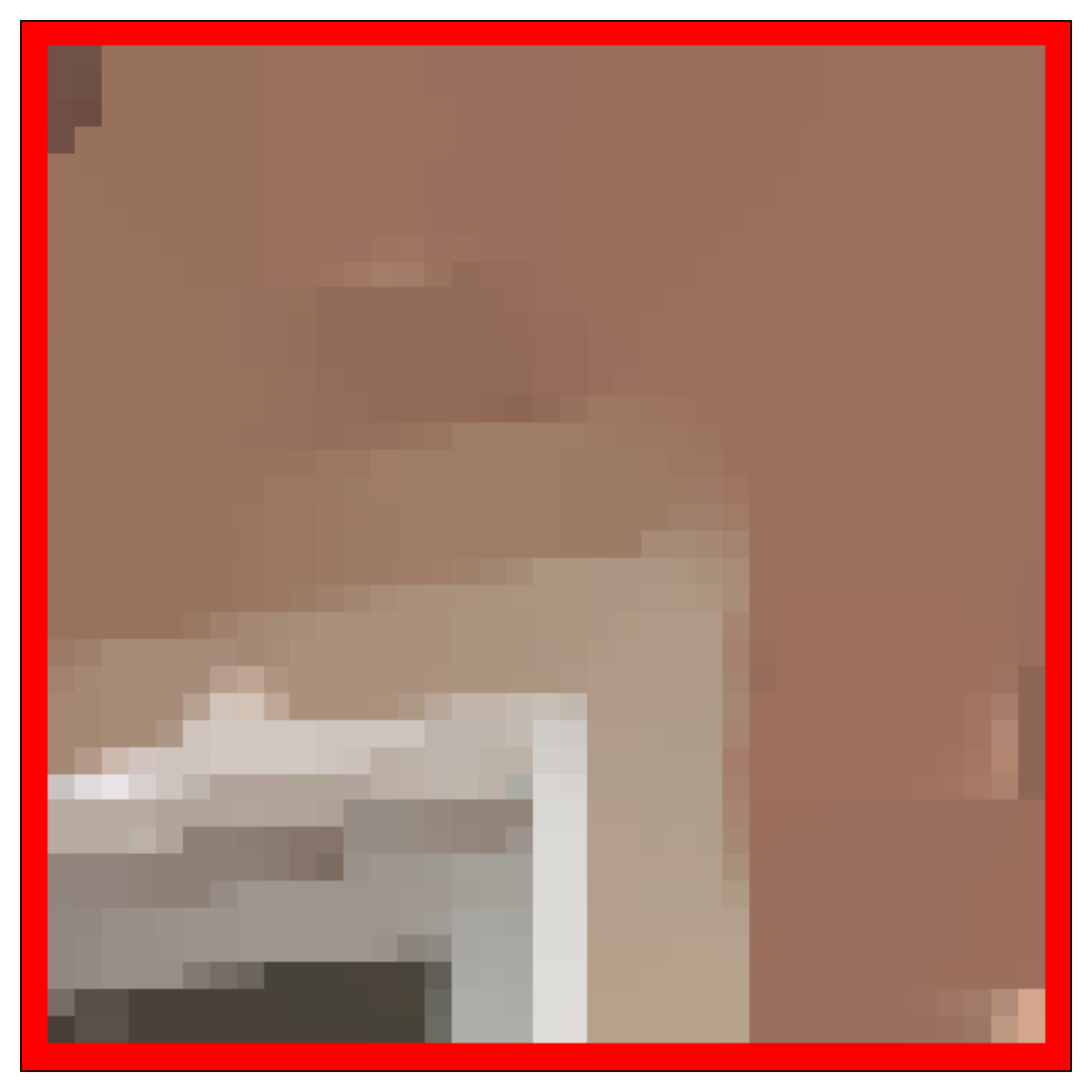}}~
\subfloat[PNEN w/ APNB \cite{zhu2019asymmetric}]{\includegraphics[width=0.24\textwidth]{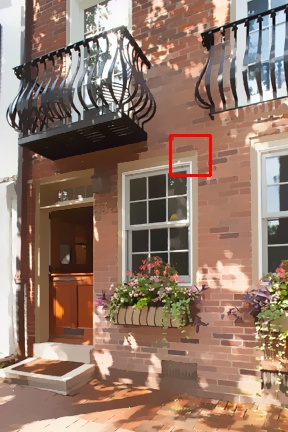}}
\hspace{-0.11\textwidth}\subfloat{\includegraphics[width=0.11\textwidth]{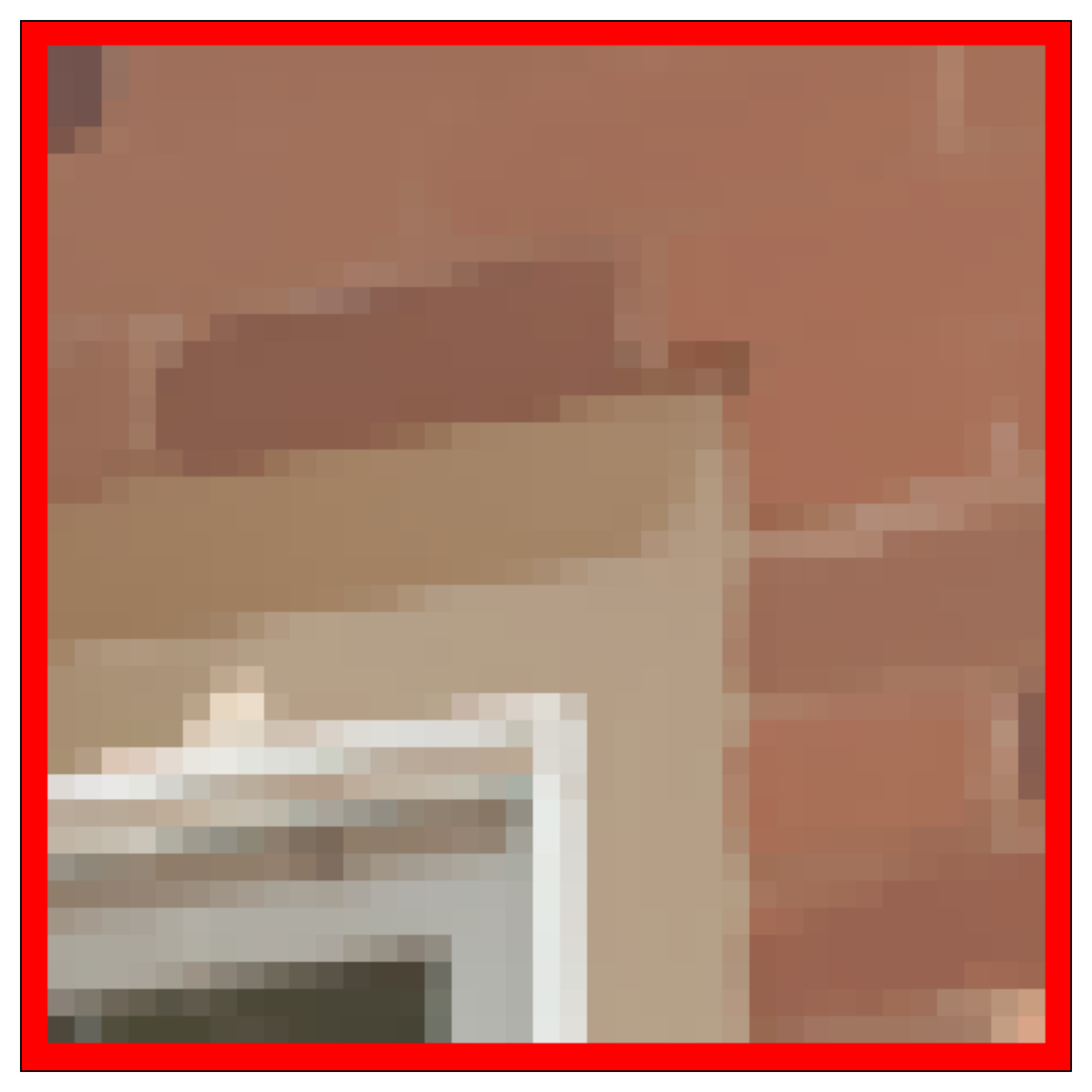}}~
\subfloat[PNEN w/ PNB]{\includegraphics[width=0.24\textwidth]{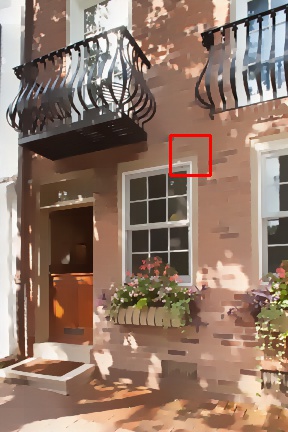}}
\hspace{-0.11\textwidth}\subfloat{\includegraphics[width=0.11\textwidth]{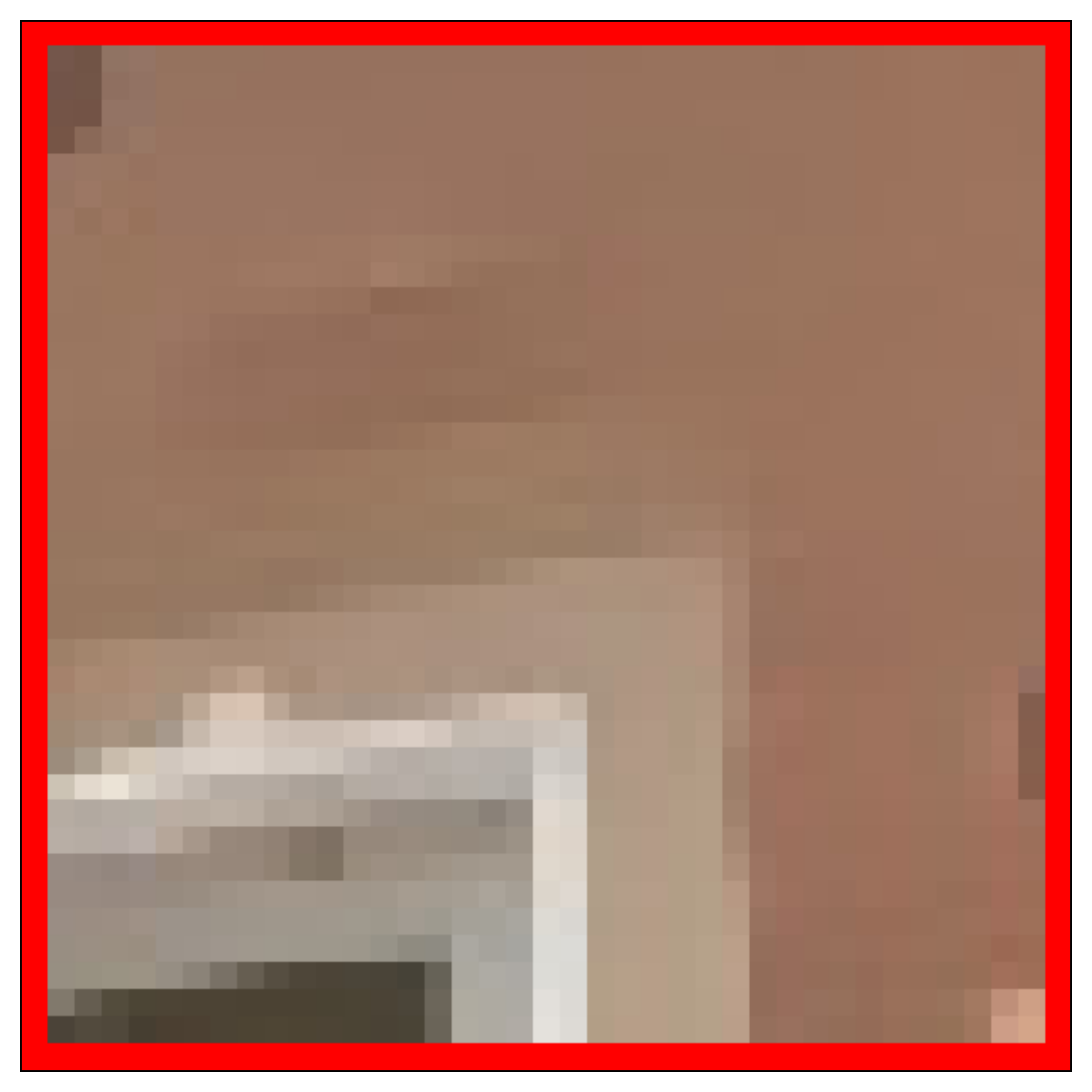}}\
\caption{Visual comparison of learning $L_0$ smoothing filter. APNB \cite{zhu2019asymmetric} can not remove the 'brick' textures as effectively as our method. }
\label{fig:apnb_pnb}
\end{figure*}

\subsection{Ablation study}
\label{sec:Ablation study}

\begin{table}[t]
\centering
\caption{Ablation study of using different settings for the pyramid non-local block. }
\label{table:ablation1}
\renewcommand{\arraystretch}{1.2}
\begin{tabular}{c|cccccc} \specialrule{.1em}{0em}{0em}
\hline
                &    1       & 2          & 3          & 4          & 5          & 6           \\ \hline
Scale 2        &            & \checkmark &            &            & \checkmark & \checkmark  \\
Scale 4        &            &            & \checkmark &            & \checkmark & \checkmark  \\
Scale 8        &            &            &            & \checkmark &            & \checkmark  \\ \hline
PSNR            & 31.95      &  32.57     &    33.17   &     32.87  &    33.21   &   \textbf{33.44}     \\
SSIM            & 0.9665     &  0.9706    &    0.9708  &     0.9707 &    0.9720  &   \textbf{0.9741}    \\ \specialrule{.1em}{0em}{0em}
\end{tabular}
\end{table}

\begin{table}[t]
\centering
\caption{Comparison of algorithm efficiency in term of floating point operations (FLOPs) and memory consumption.}
\label{table:ablation2}
\renewcommand{\arraystretch}{1.2}
\begin{tabular}{l|c|c|c|c} \specialrule{.1em}{0em}{0em}
NL Type                           &FLOPs &Memory &\#Params &PSNR   \\ \hline
w/o                               &40.6G &1.1GB  &3401k    &31.95 \\
w/ NLB~\cite{wang2018non}         &46.2G &9.6GB  &3524k    &32.40  \\ 
w/ APNB~\cite{zhu2019asymmetric}  &41.7G &2.1GB  &3418k    &32.10  \\ \hline
w/ Ours-PNB                      &42.5G &4.3GB  &3771k    &33.44   \\ \specialrule{.1em}{0em}{0em}
\end{tabular}
\end{table}

\begin{figure*}[t]
\captionsetup[subfigure]{labelformat=empty,farskip=1pt}
\centering
\subfloat{\includegraphics[width=0.24\textwidth]{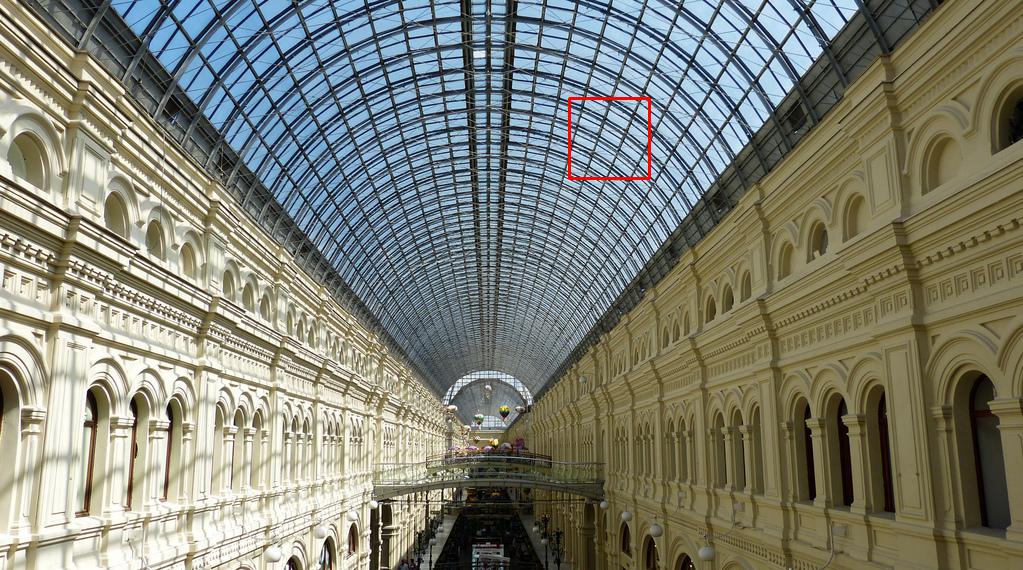}}
\hspace{-0.09\textwidth}\subfloat{\includegraphics[width=0.09\textwidth]{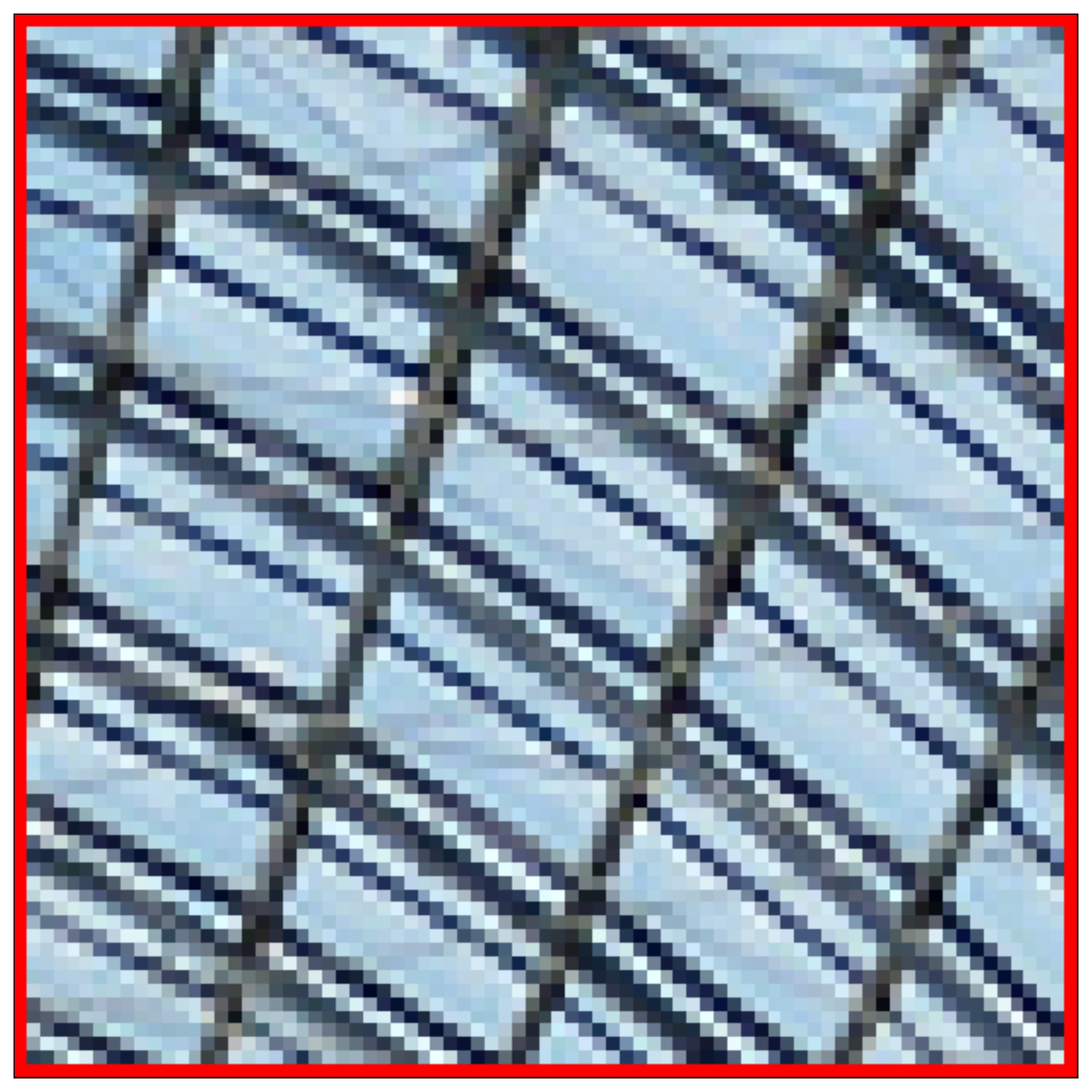}}~
\subfloat{\includegraphics[width=0.24\textwidth]{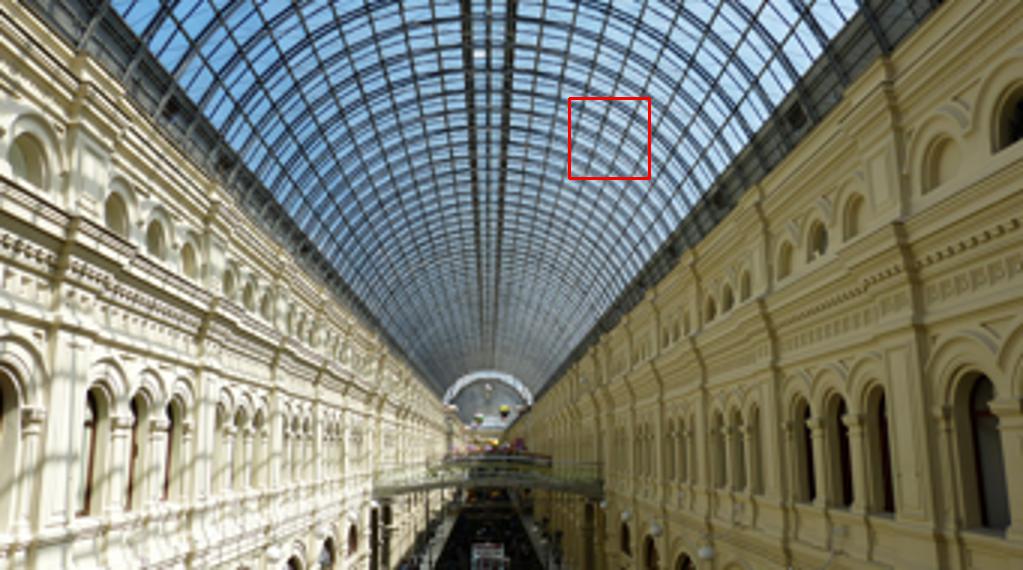}}
\hspace{-0.09\textwidth}\subfloat{\includegraphics[width=0.09\textwidth]{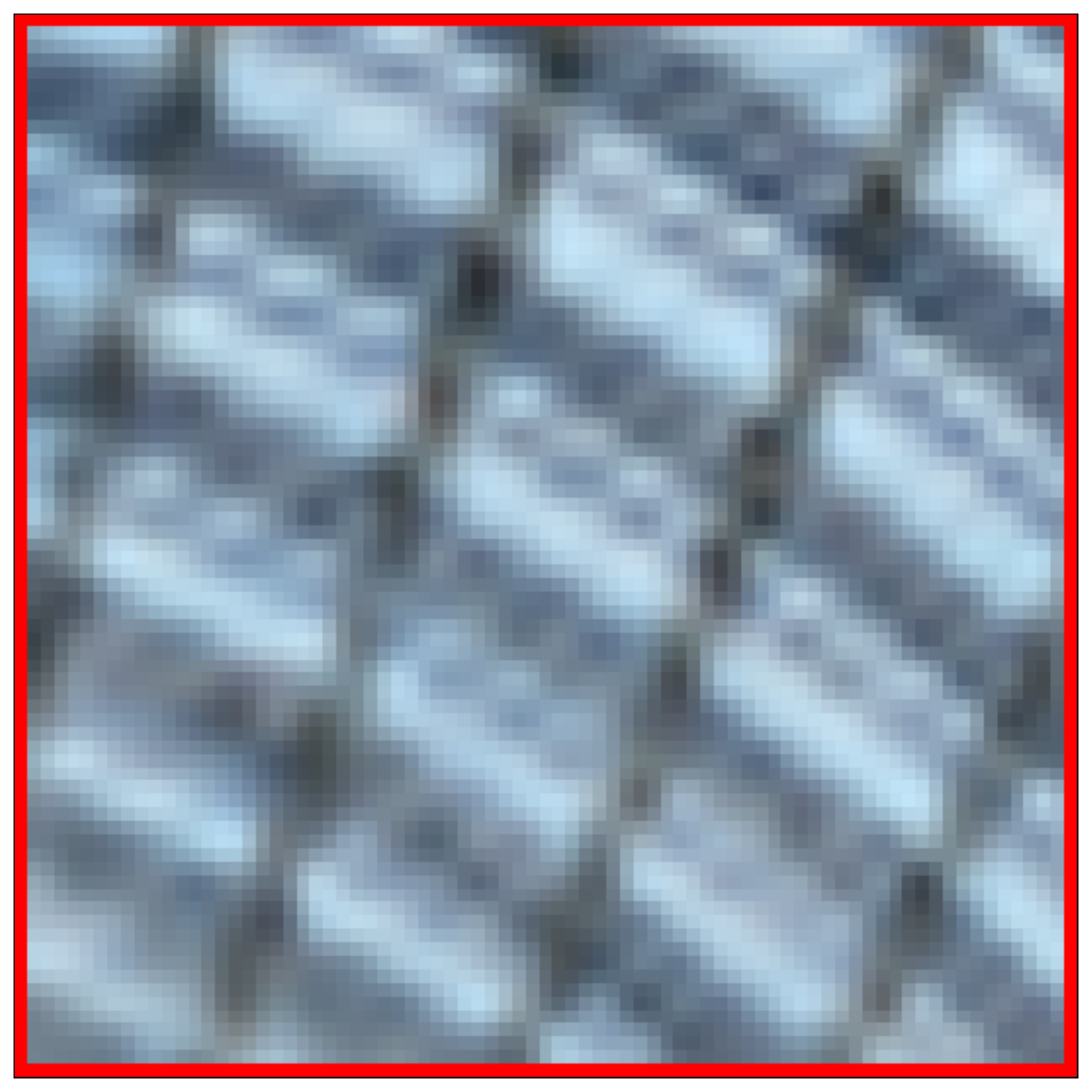}}~
\subfloat{\includegraphics[width=0.24\textwidth]{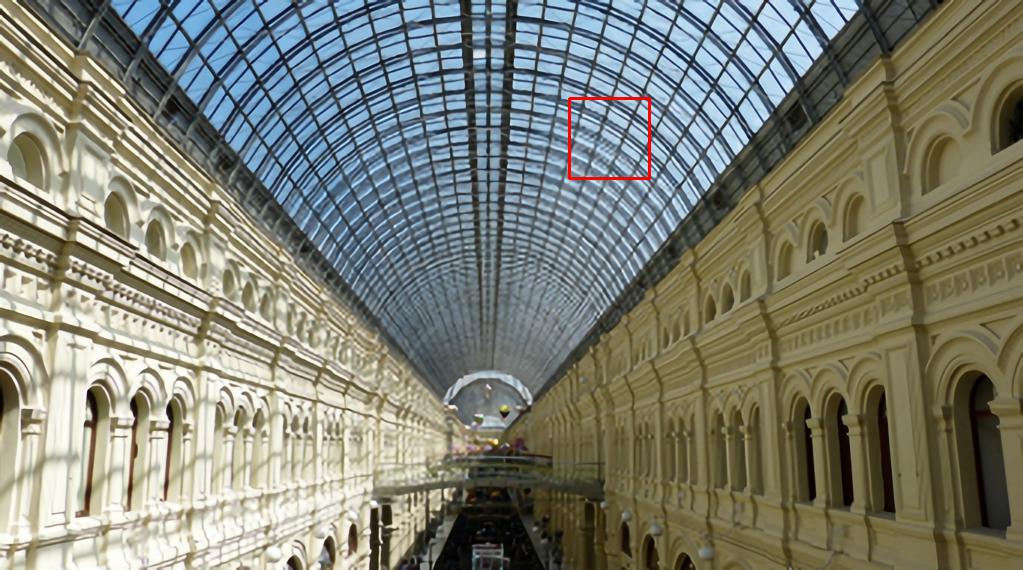}}
\hspace{-0.09\textwidth}\subfloat{\includegraphics[width=0.09\textwidth]{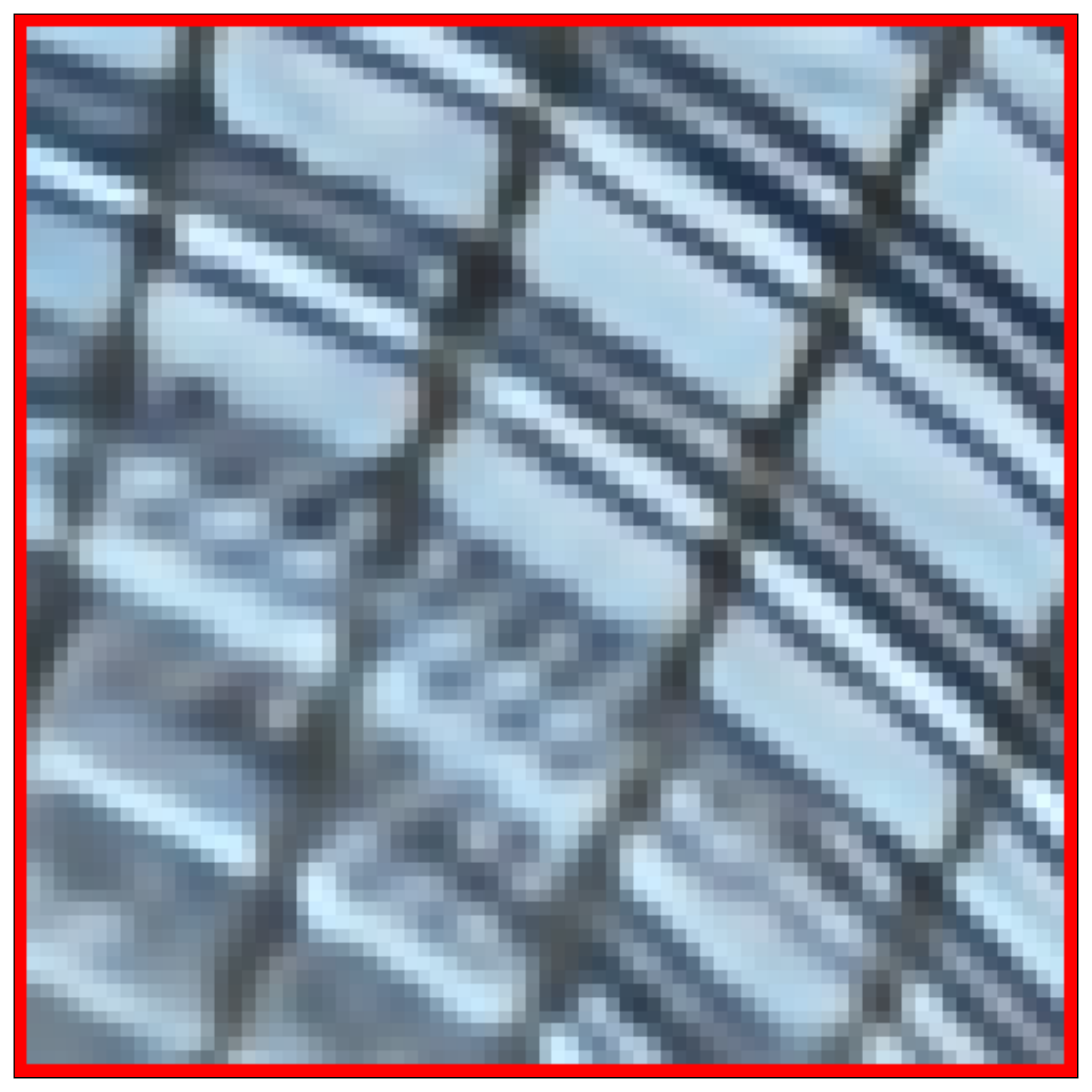}}~
\subfloat{\includegraphics[width=0.24\textwidth]{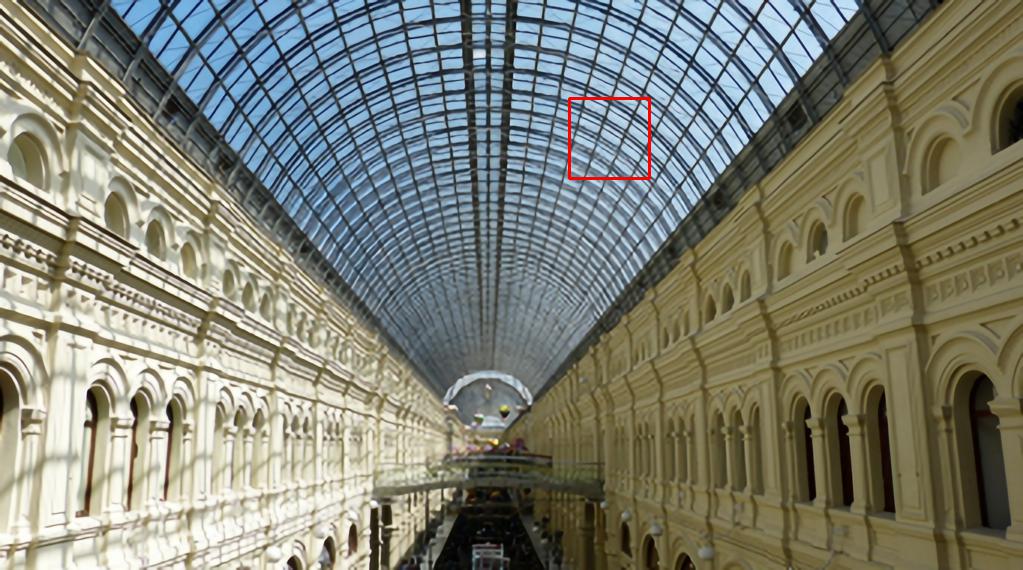}}
\hspace{-0.09\textwidth}\subfloat{\includegraphics[width=0.09\textwidth]{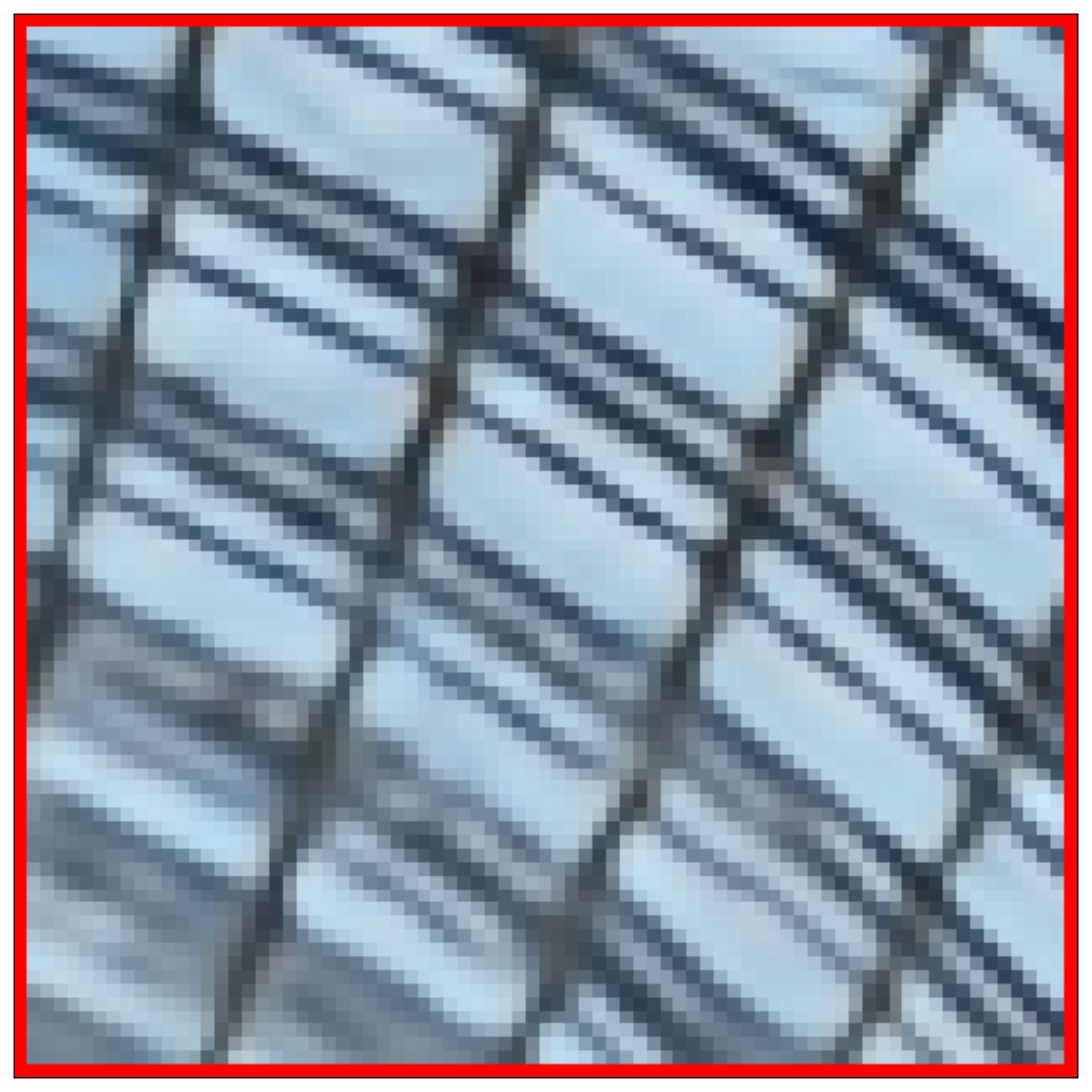}}\
\subfloat[Ground Truth]{\includegraphics[width=0.24\textwidth]{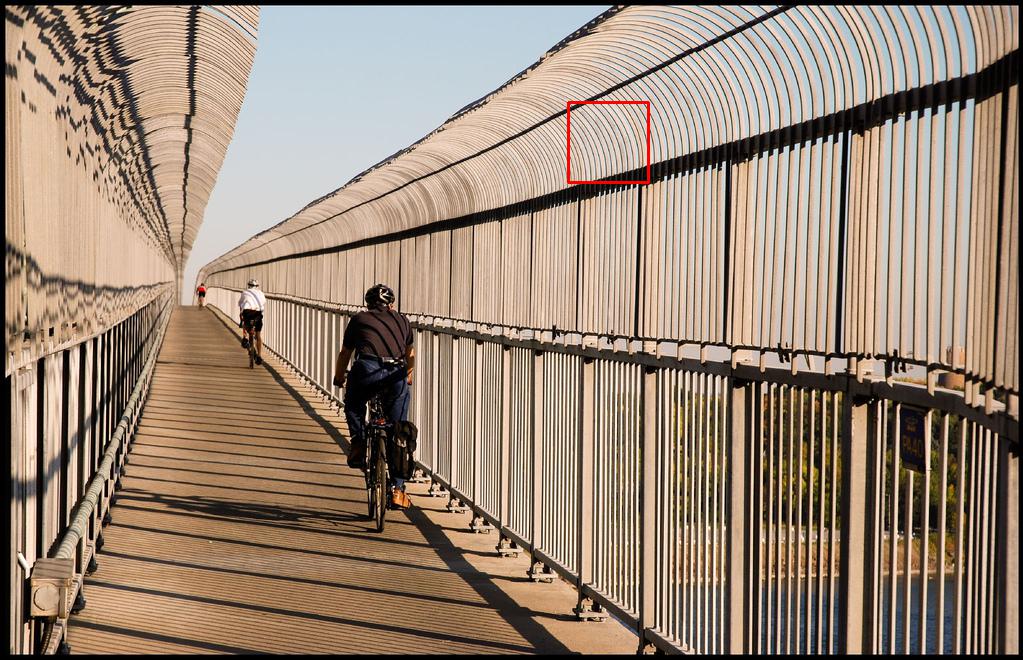}}
\hspace{-0.09\textwidth}\subfloat{\includegraphics[width=0.09\textwidth]{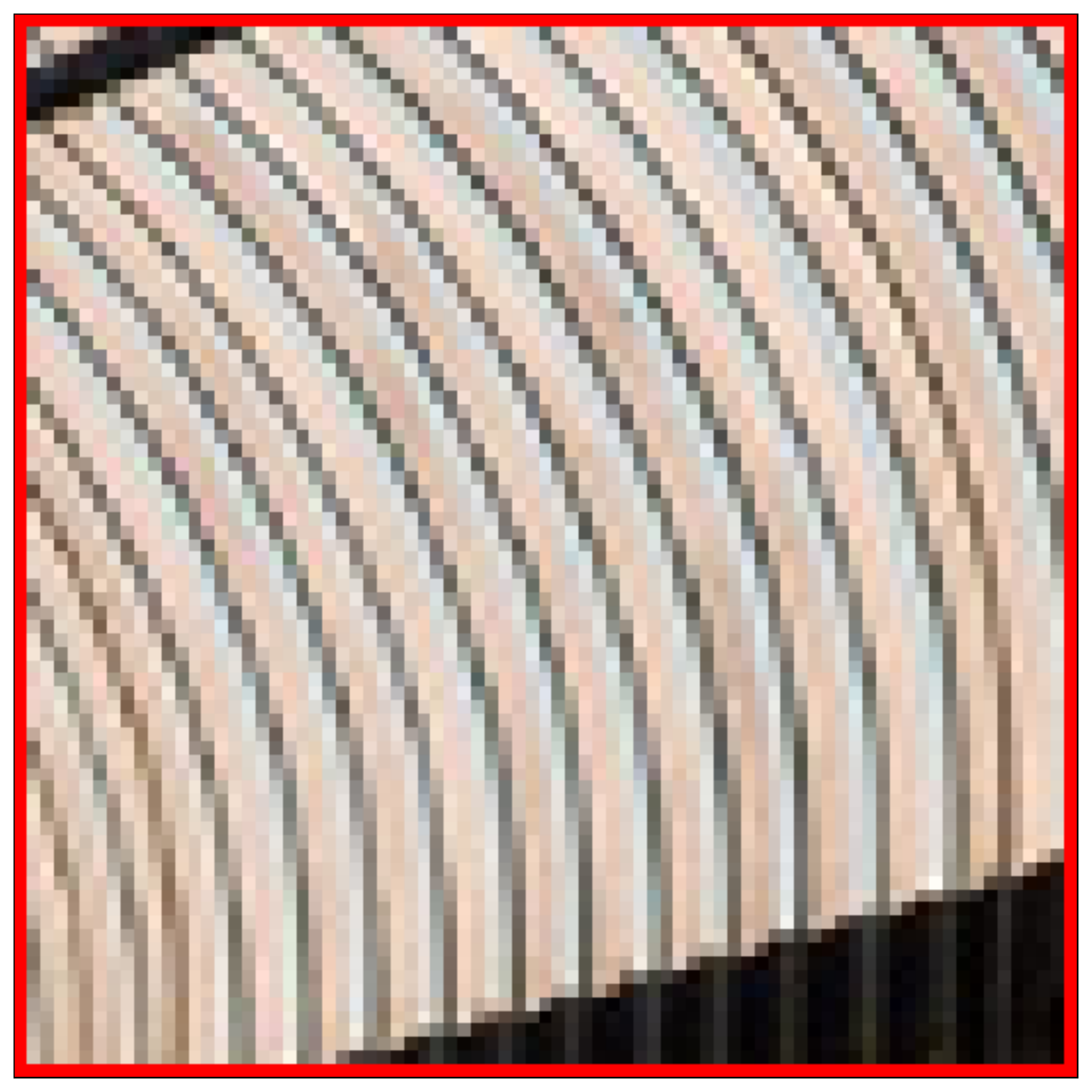}}~
\subfloat[Bicubic]{\includegraphics[width=0.24\textwidth]{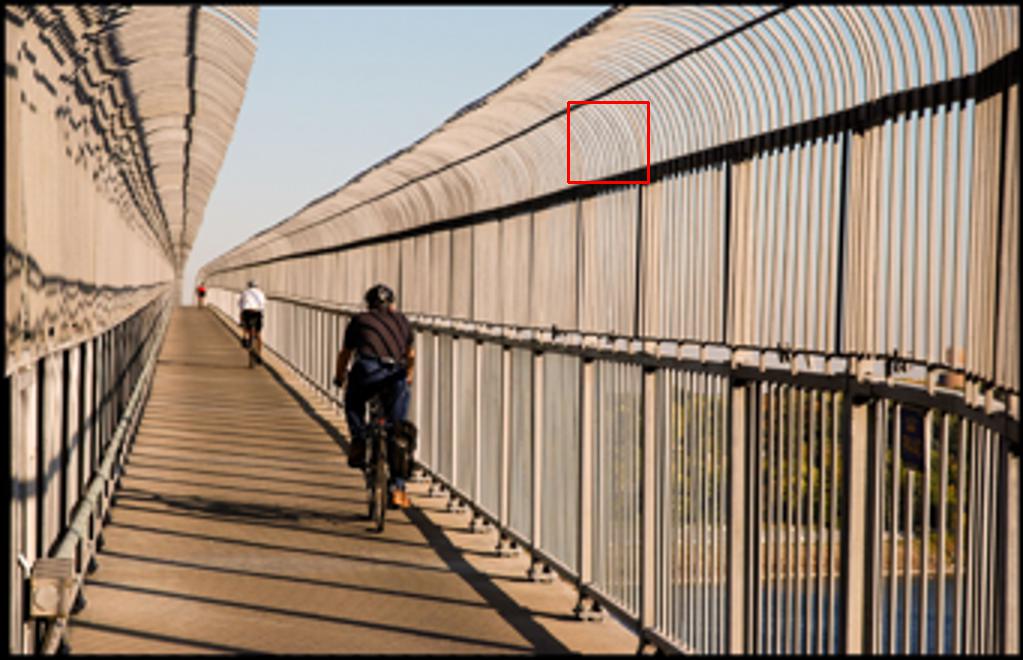}}
\hspace{-0.09\textwidth}\subfloat{\includegraphics[width=0.09\textwidth]{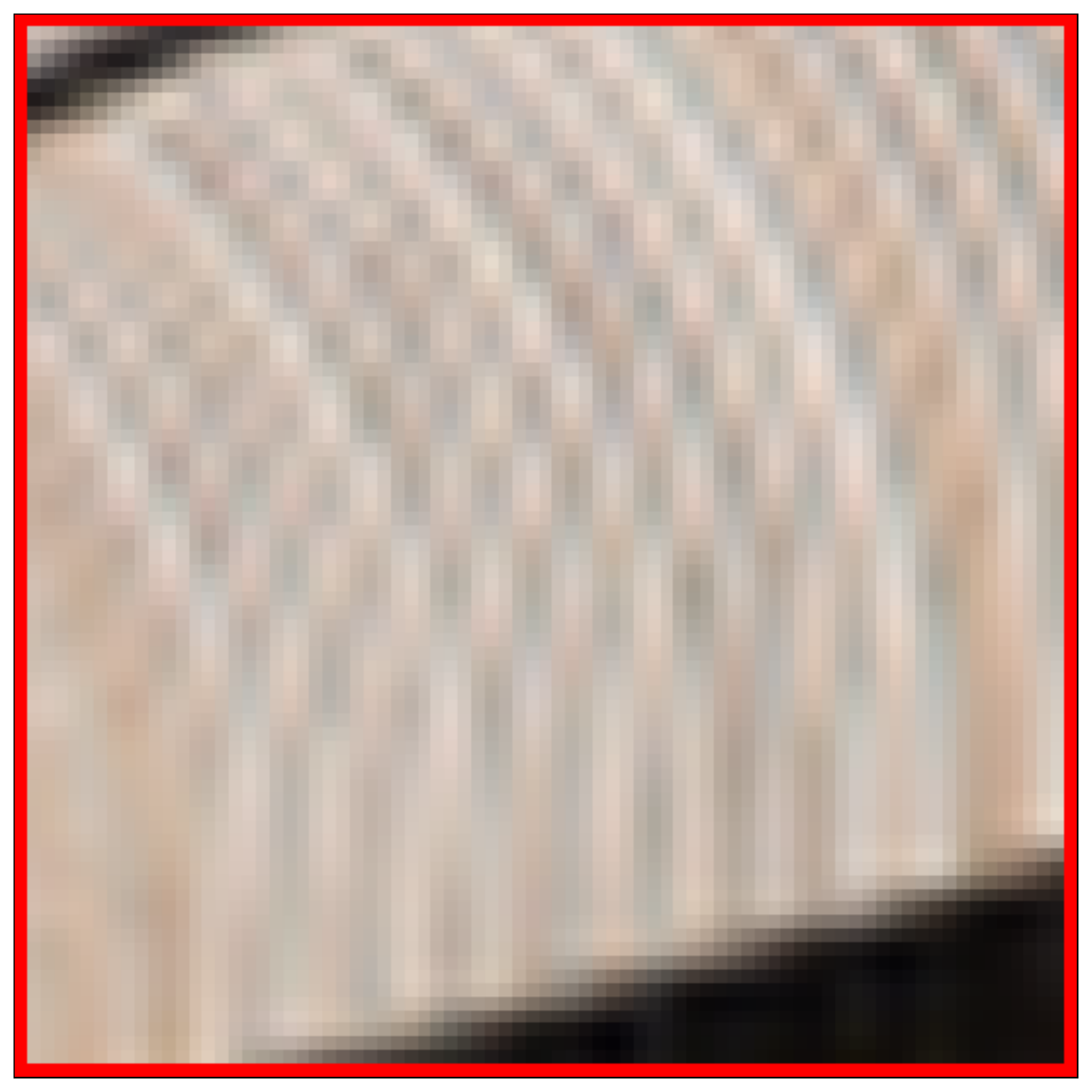}}~
\subfloat[MemNet]{\includegraphics[width=0.24\textwidth]{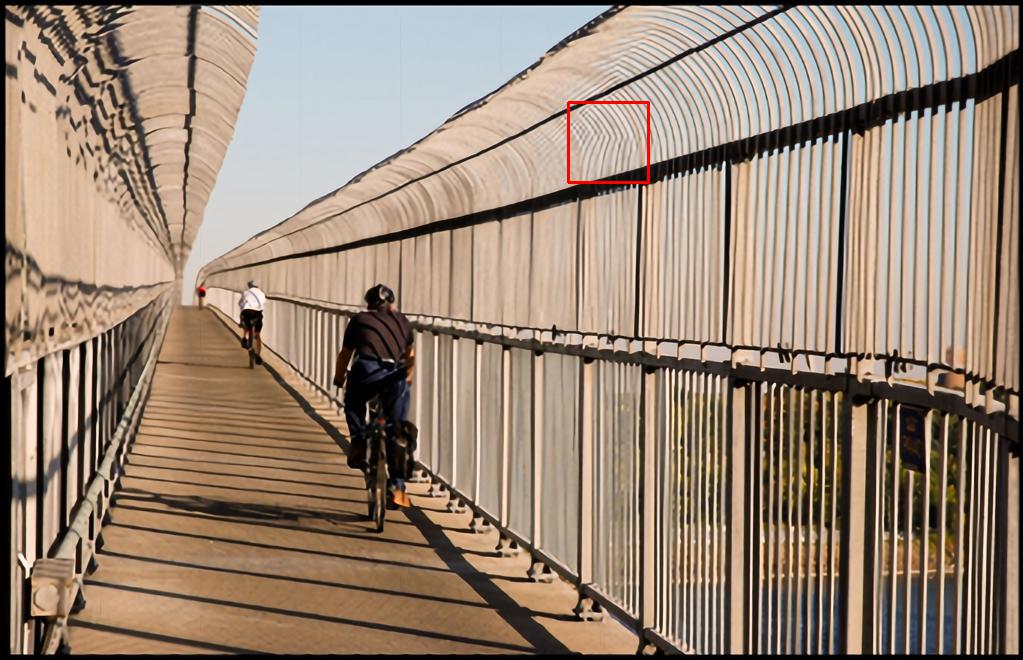}}
\hspace{-0.09\textwidth}\subfloat{\includegraphics[width=0.09\textwidth]{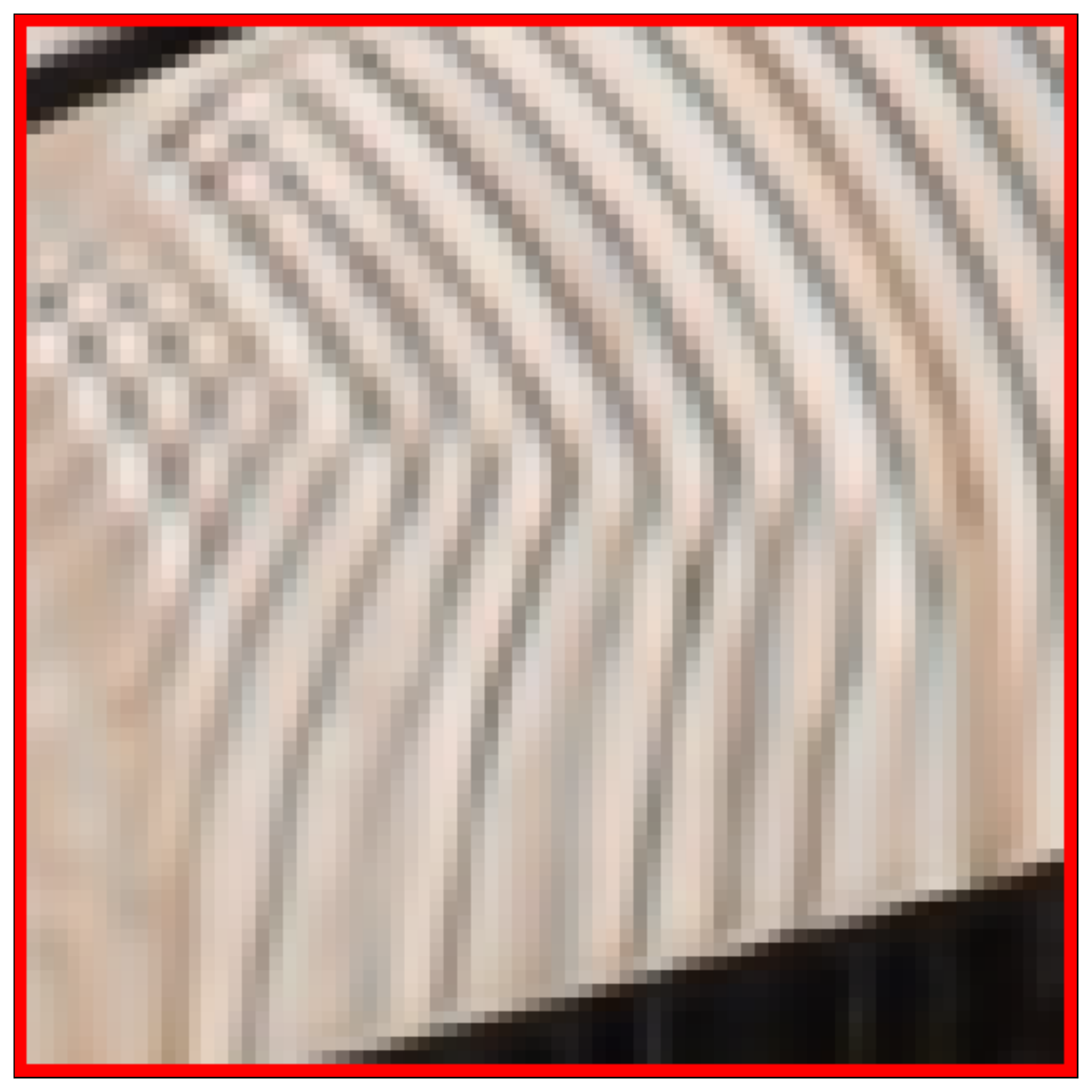}}~
\subfloat[MemNet (w/ PNB)]{\includegraphics[width=0.24\textwidth]{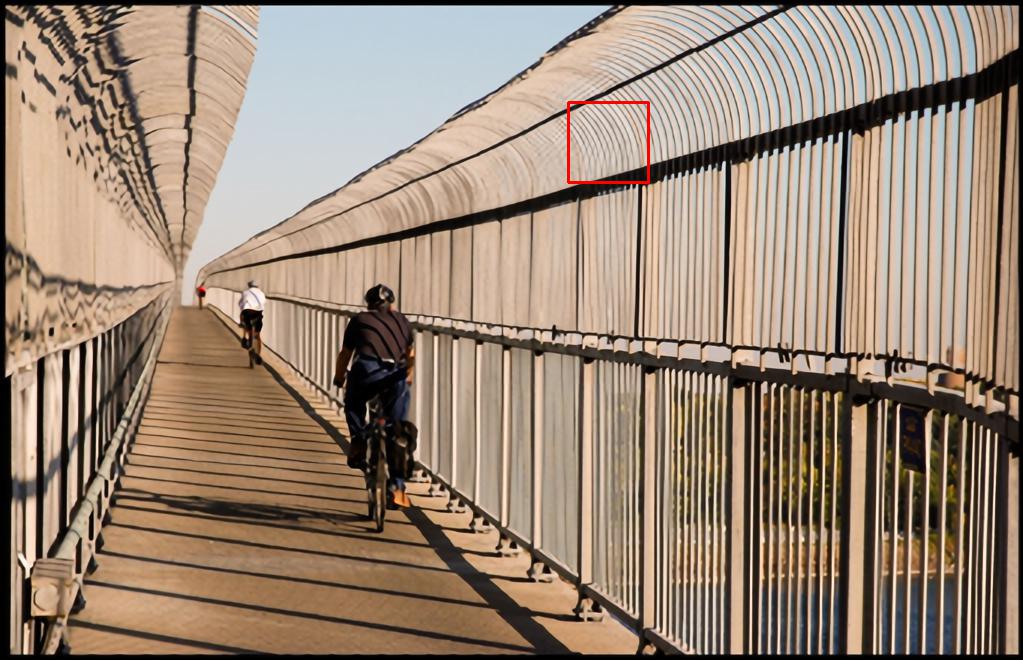}}
\hspace{-0.09\textwidth}\subfloat{\includegraphics[width=0.09\textwidth]{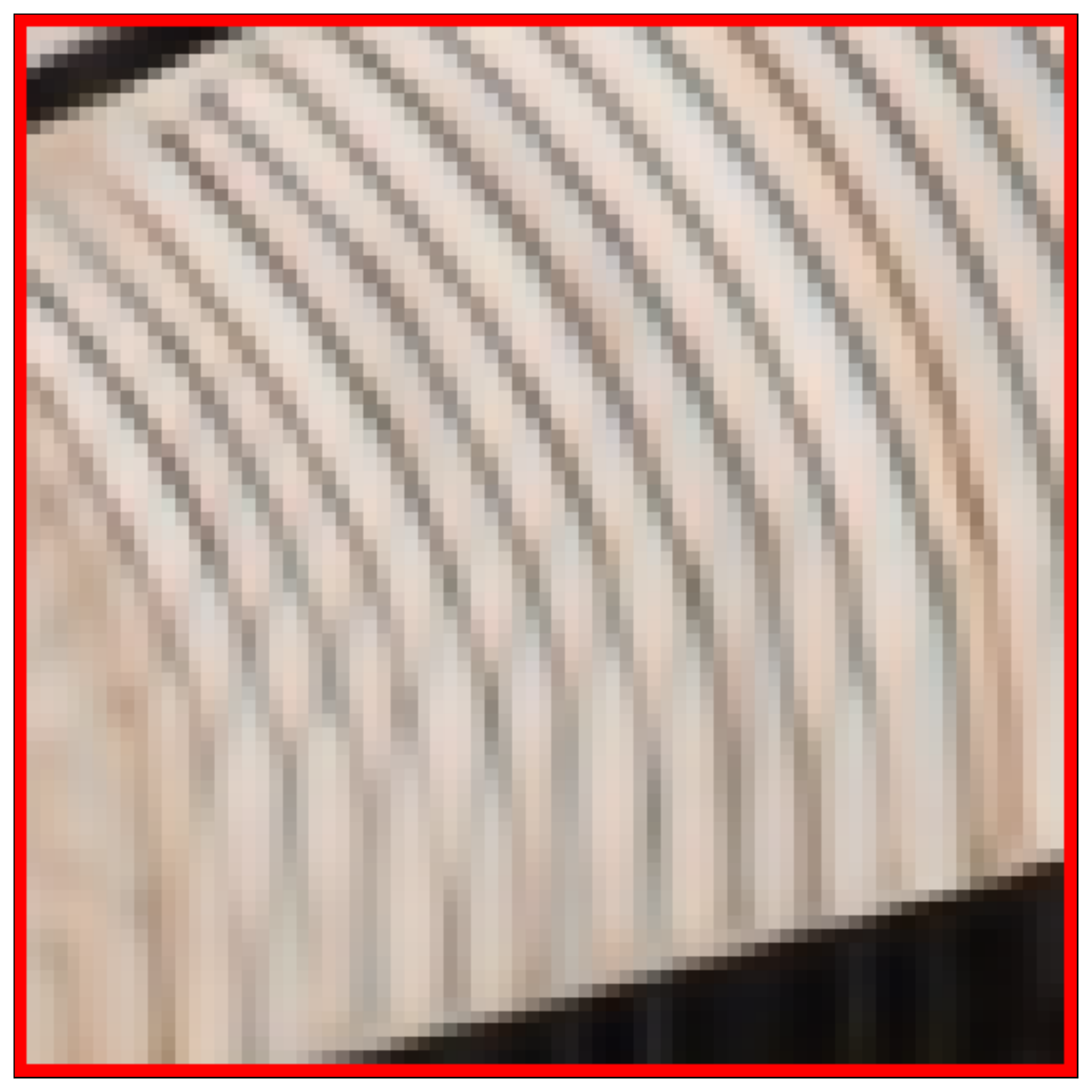}}\
\caption{Visual comparison of image super-resolution with scaling factor $\times 3$. From left to right: the HR images, the LR images synthesized via bicubic degradation, results produced by baseline MemNet \cite{Tai-MemNet-2017} and results produced by PNB-enhanced MemNet. }
\label{fig:sisr_memnet}
\end{figure*}

\begin{figure*}[t]
\captionsetup[subfigure]{labelformat=empty,farskip=1pt}
\centering
\subfloat{\includegraphics[width=0.24\textwidth]{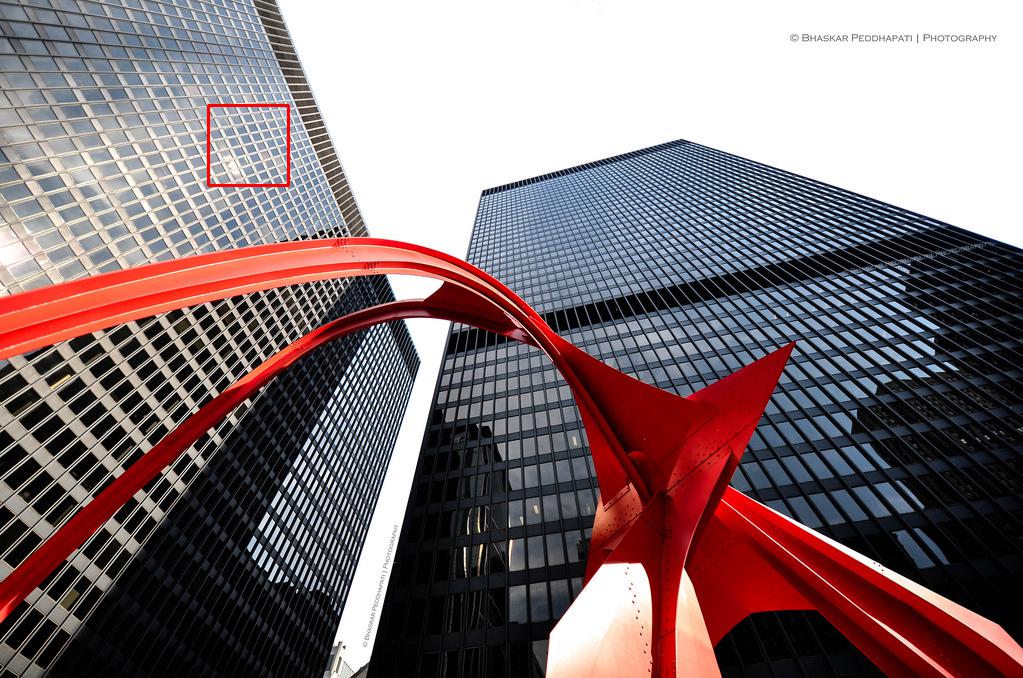}}
\hspace{-0.09\textwidth}\subfloat{\includegraphics[width=0.09\textwidth]{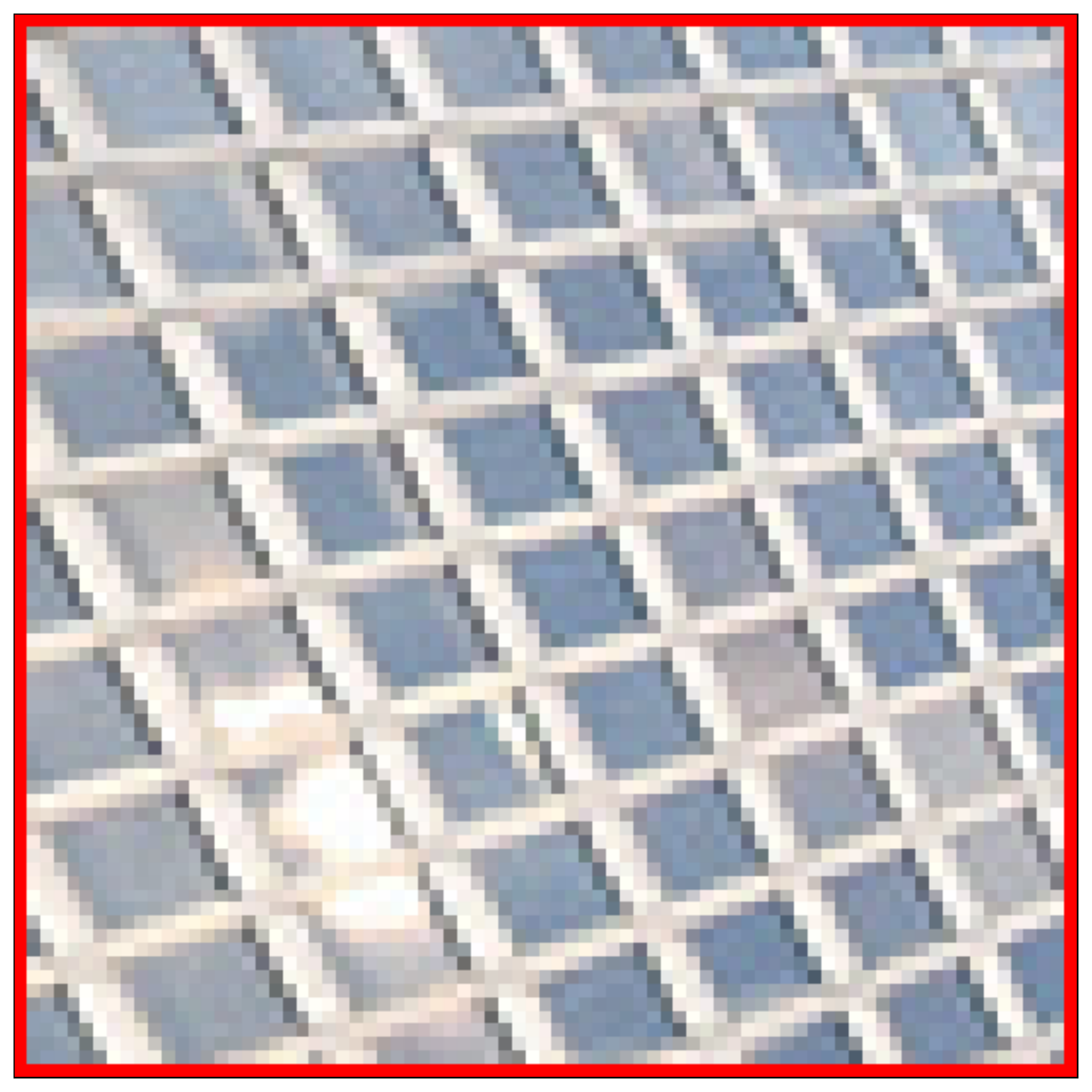}}~
\subfloat{\includegraphics[width=0.24\textwidth]{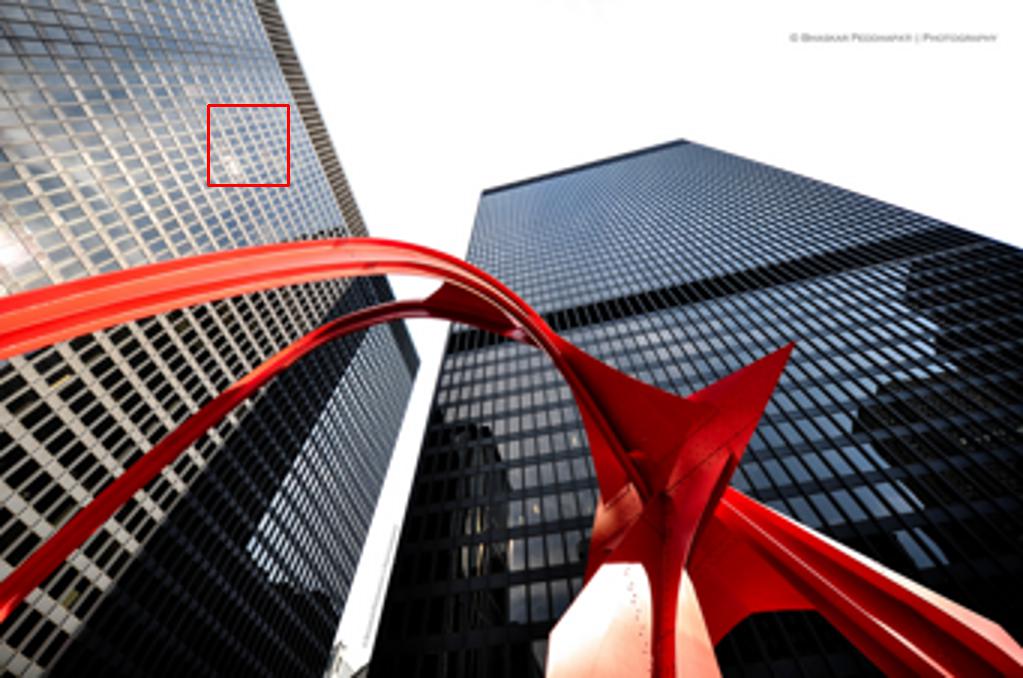}}
\hspace{-0.09\textwidth}\subfloat{\includegraphics[width=0.09\textwidth]{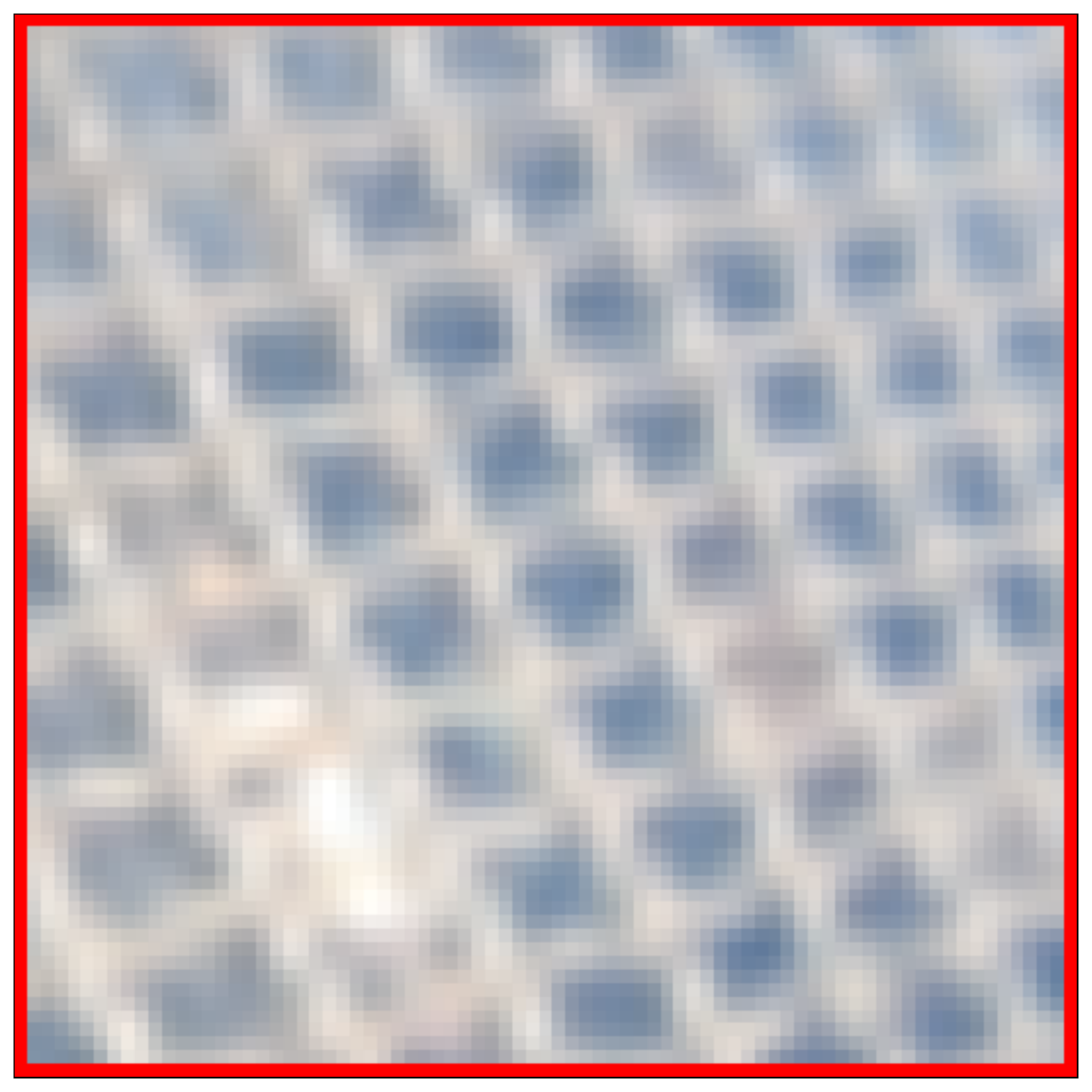}}~
\subfloat{\includegraphics[width=0.24\textwidth]{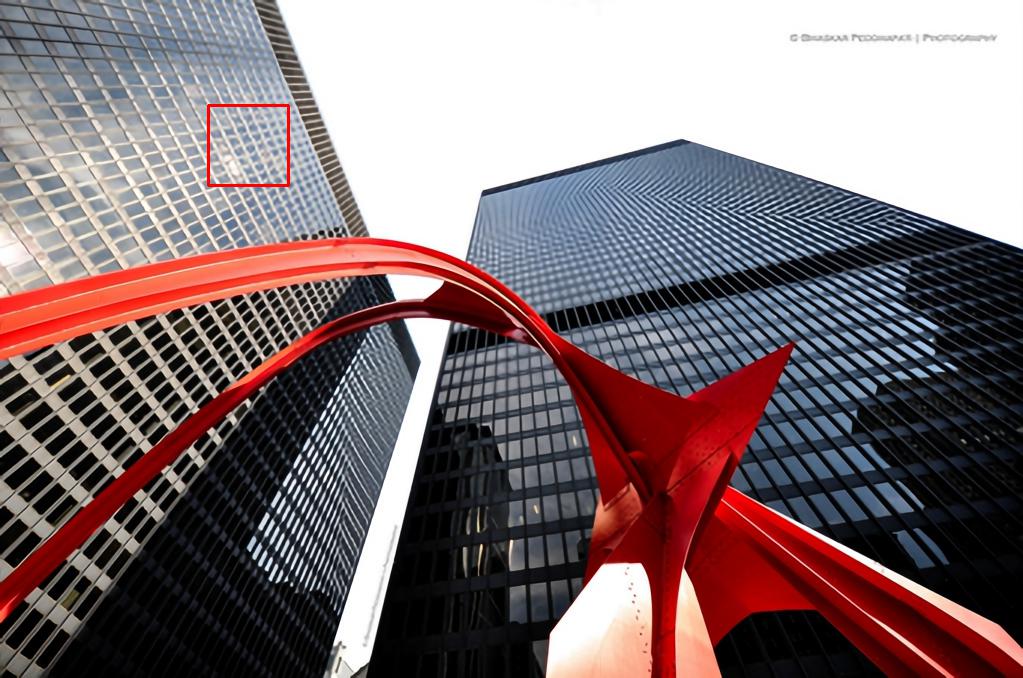}}
\hspace{-0.09\textwidth}\subfloat{\includegraphics[width=0.09\textwidth]{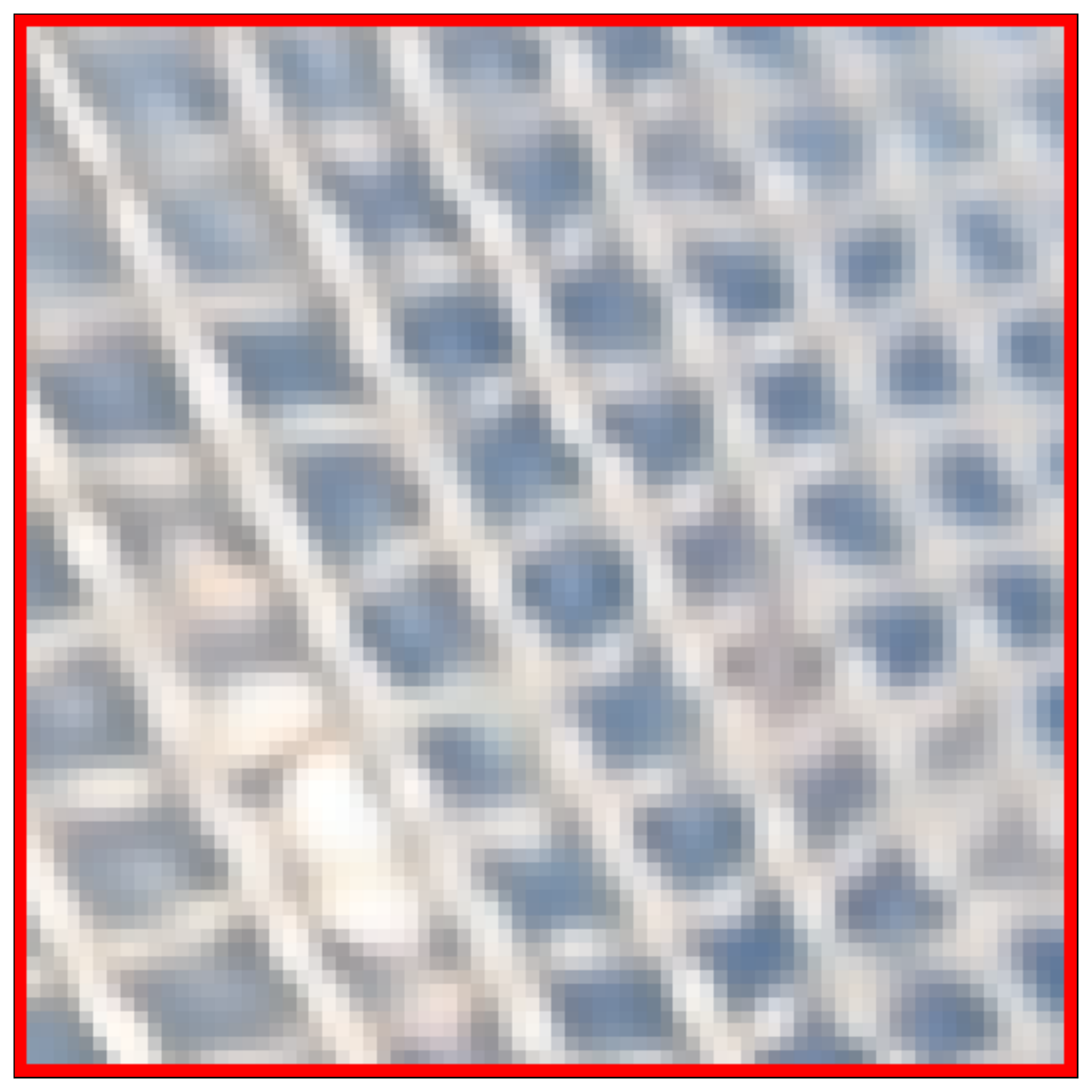}}~
\subfloat{\includegraphics[width=0.24\textwidth]{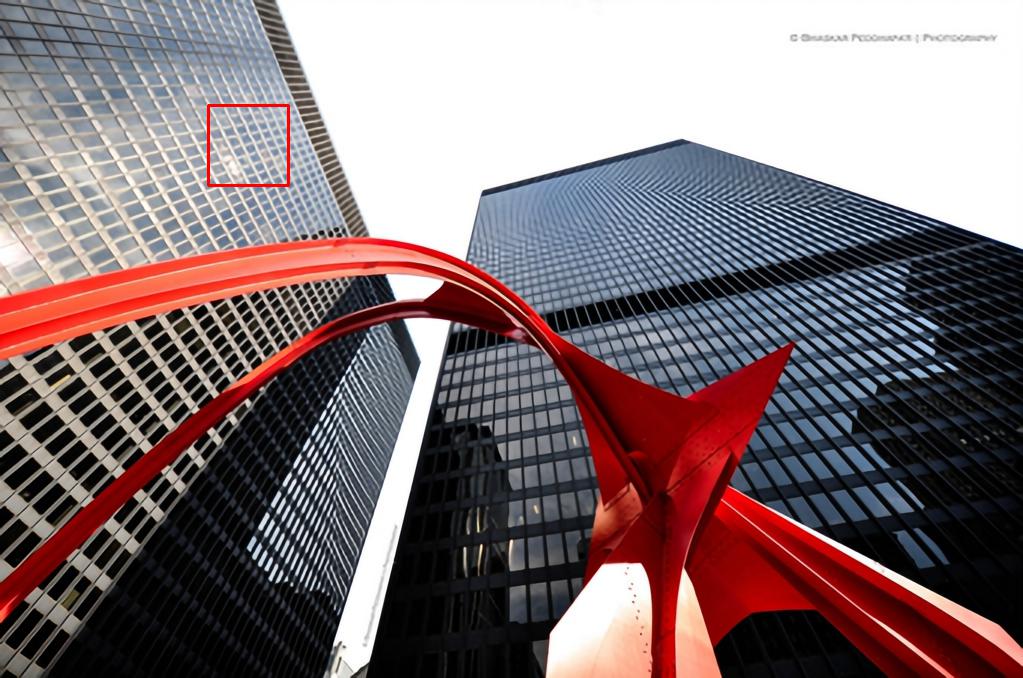}}
\hspace{-0.09\textwidth}\subfloat{\includegraphics[width=0.09\textwidth]{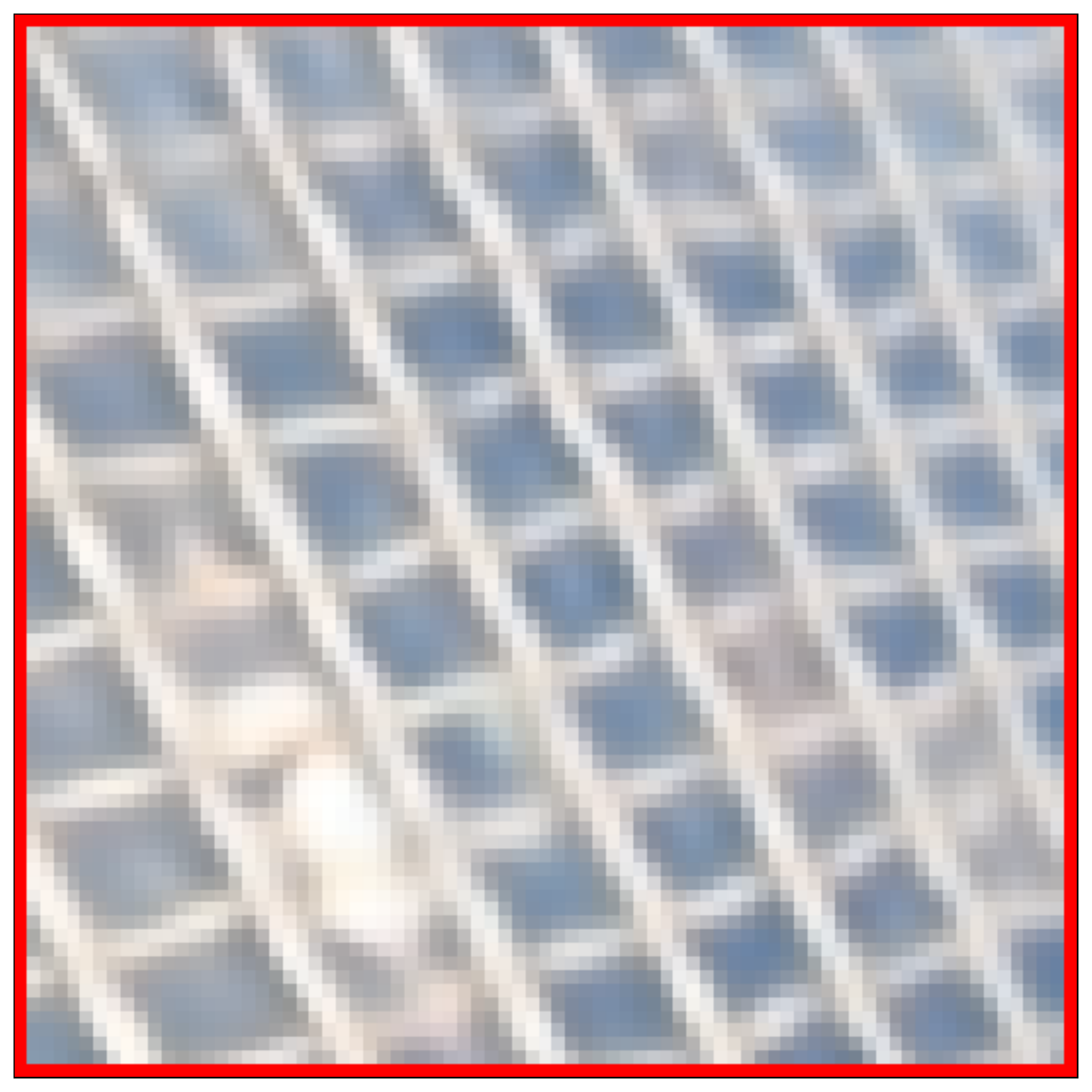}}\
\subfloat[Ground Truth]{\includegraphics[width=0.24\textwidth]{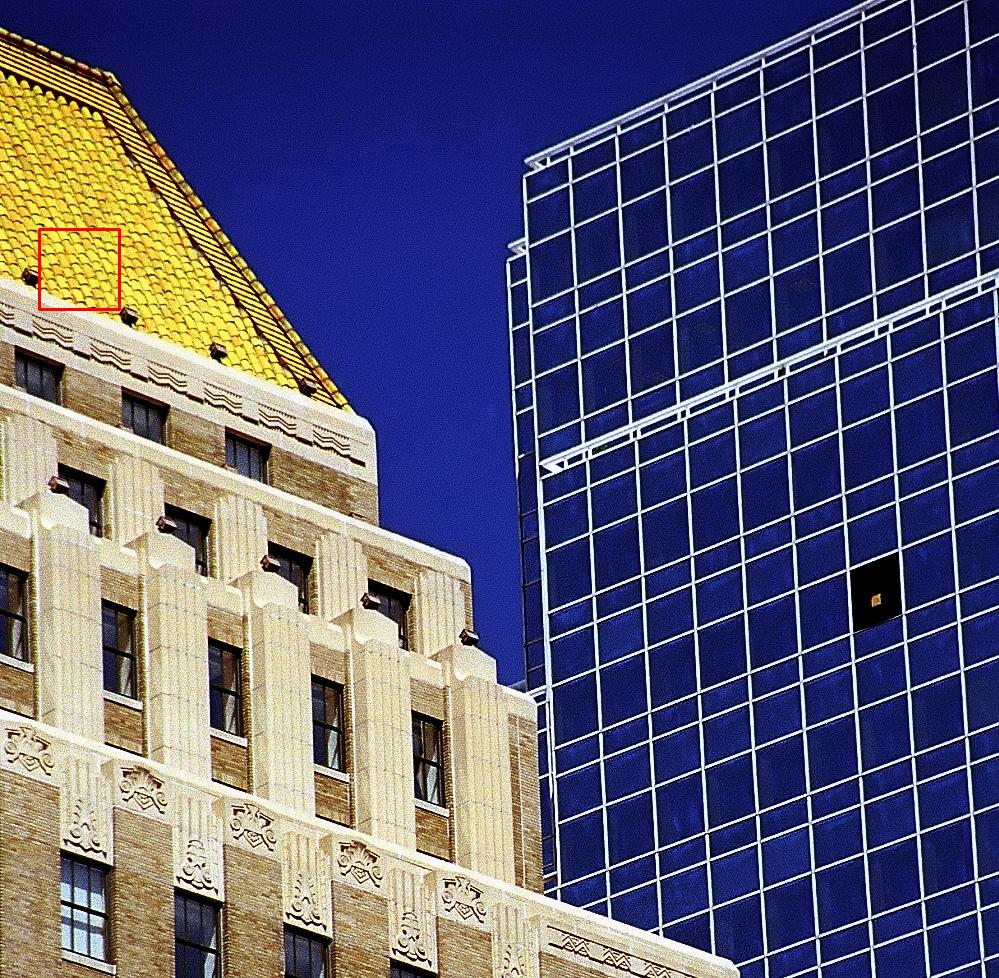}}
\hspace{-0.09\textwidth}\subfloat{\includegraphics[width=0.09\textwidth]{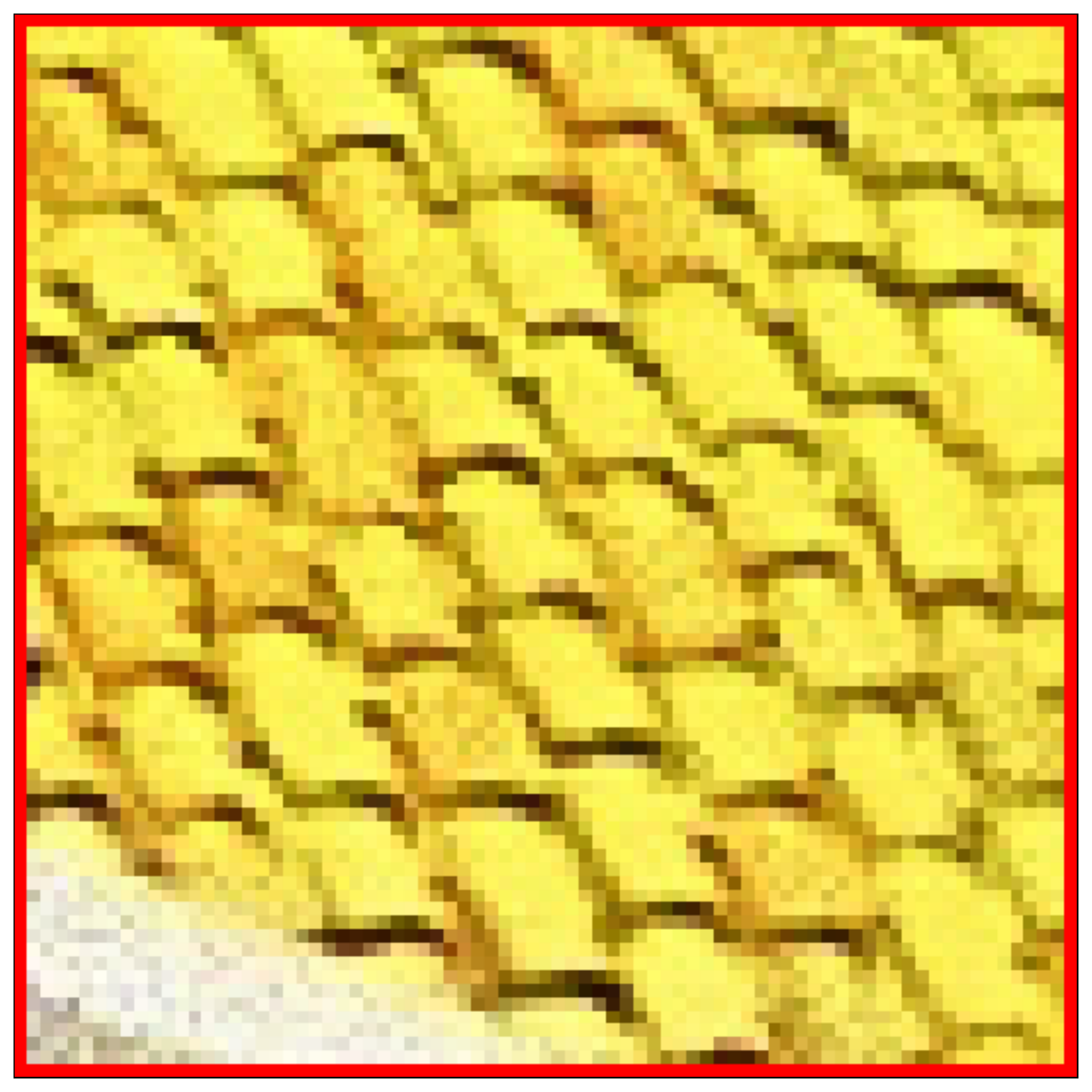}}~
\subfloat[Bicubic]{\includegraphics[width=0.24\textwidth]{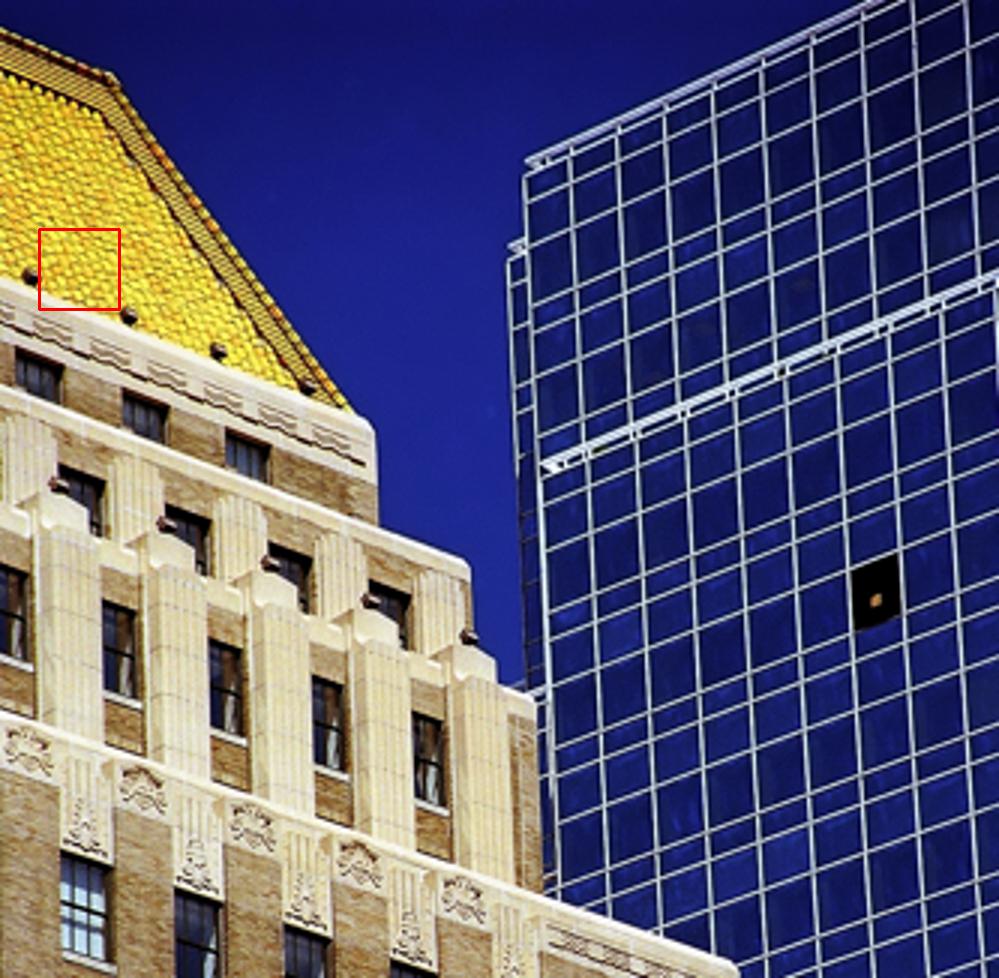}}
\hspace{-0.09\textwidth}\subfloat{\includegraphics[width=0.09\textwidth]{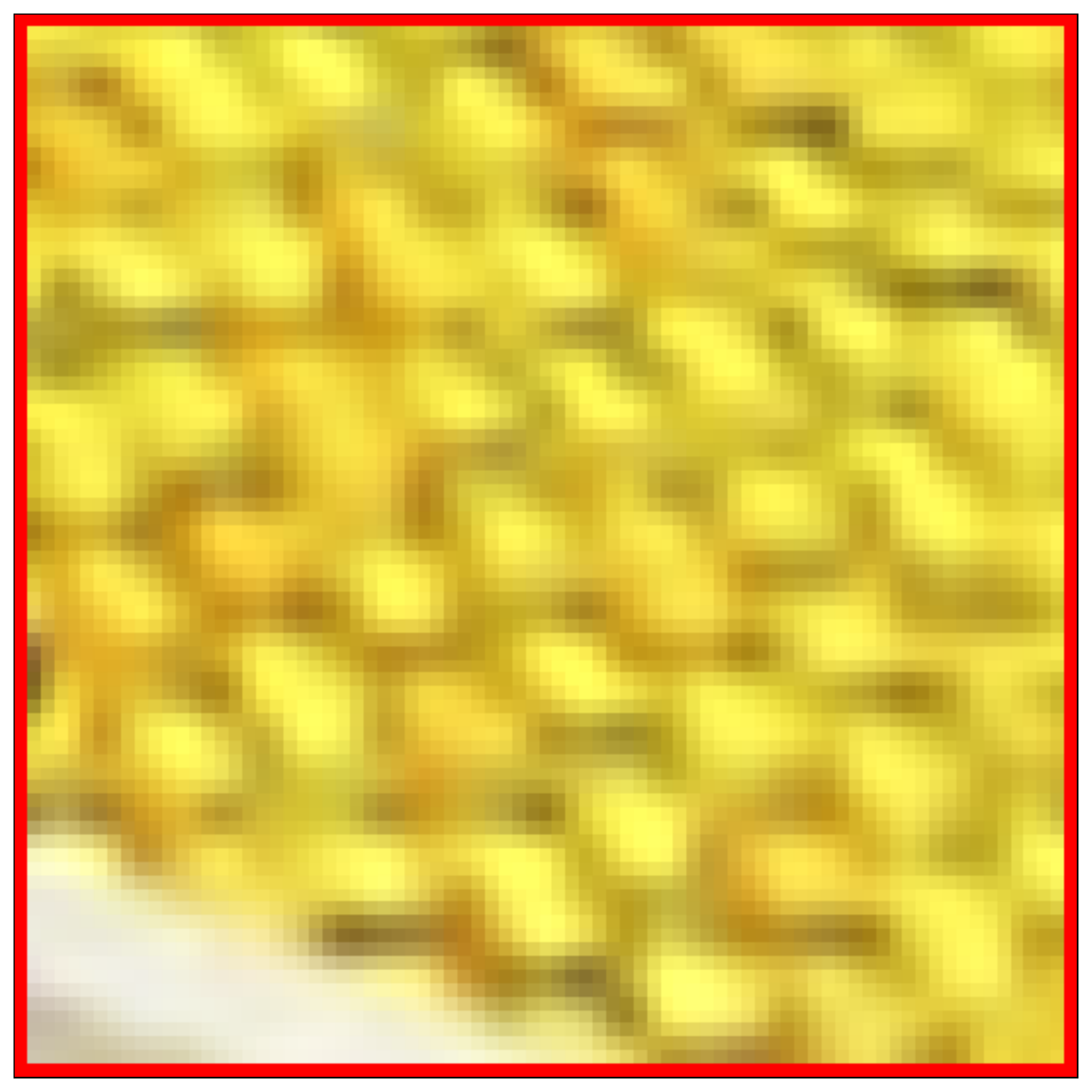}}~
\subfloat[RDN]{\includegraphics[width=0.24\textwidth]{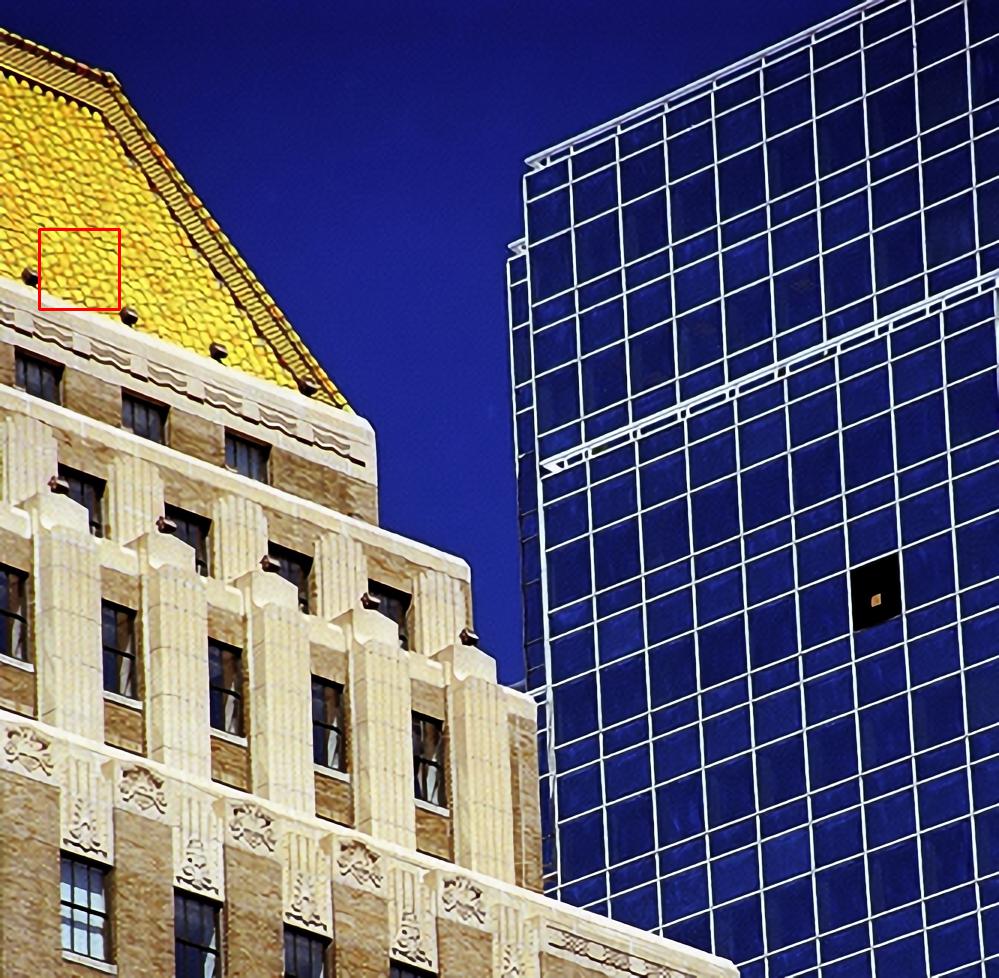}}
\hspace{-0.09\textwidth}\subfloat{\includegraphics[width=0.09\textwidth]{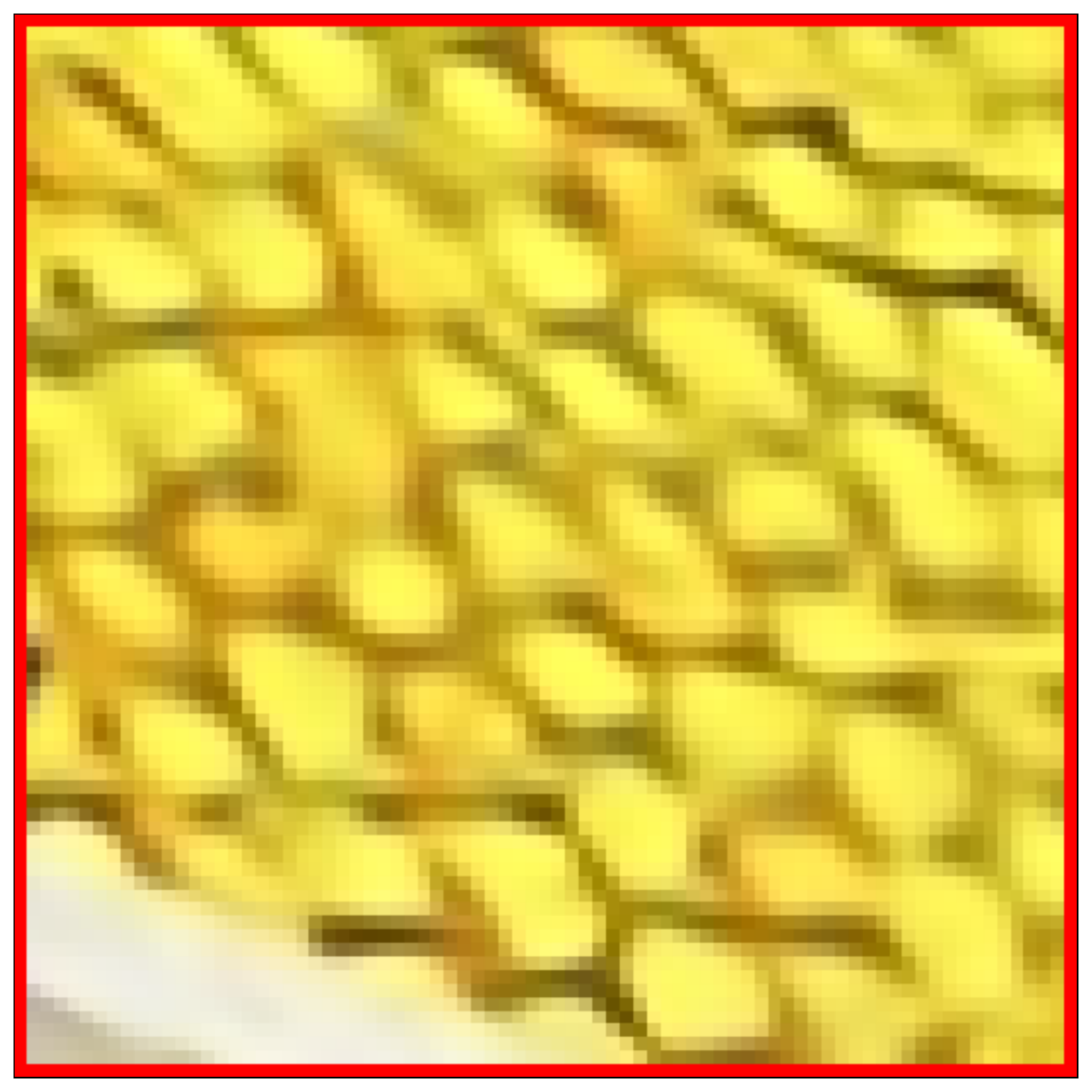}}~
\subfloat[RDN (w/ PNB)]{\includegraphics[width=0.24\textwidth]{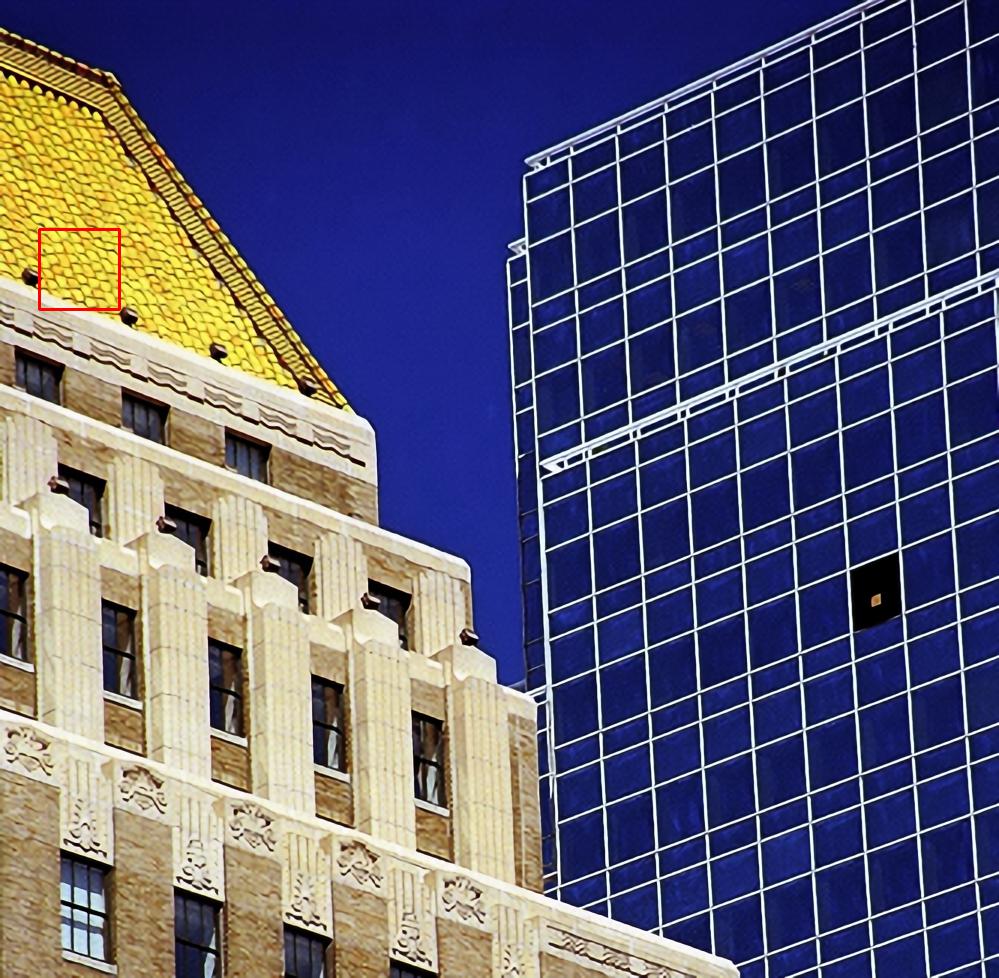}}
\hspace{-0.09\textwidth}\subfloat{\includegraphics[width=0.09\textwidth]{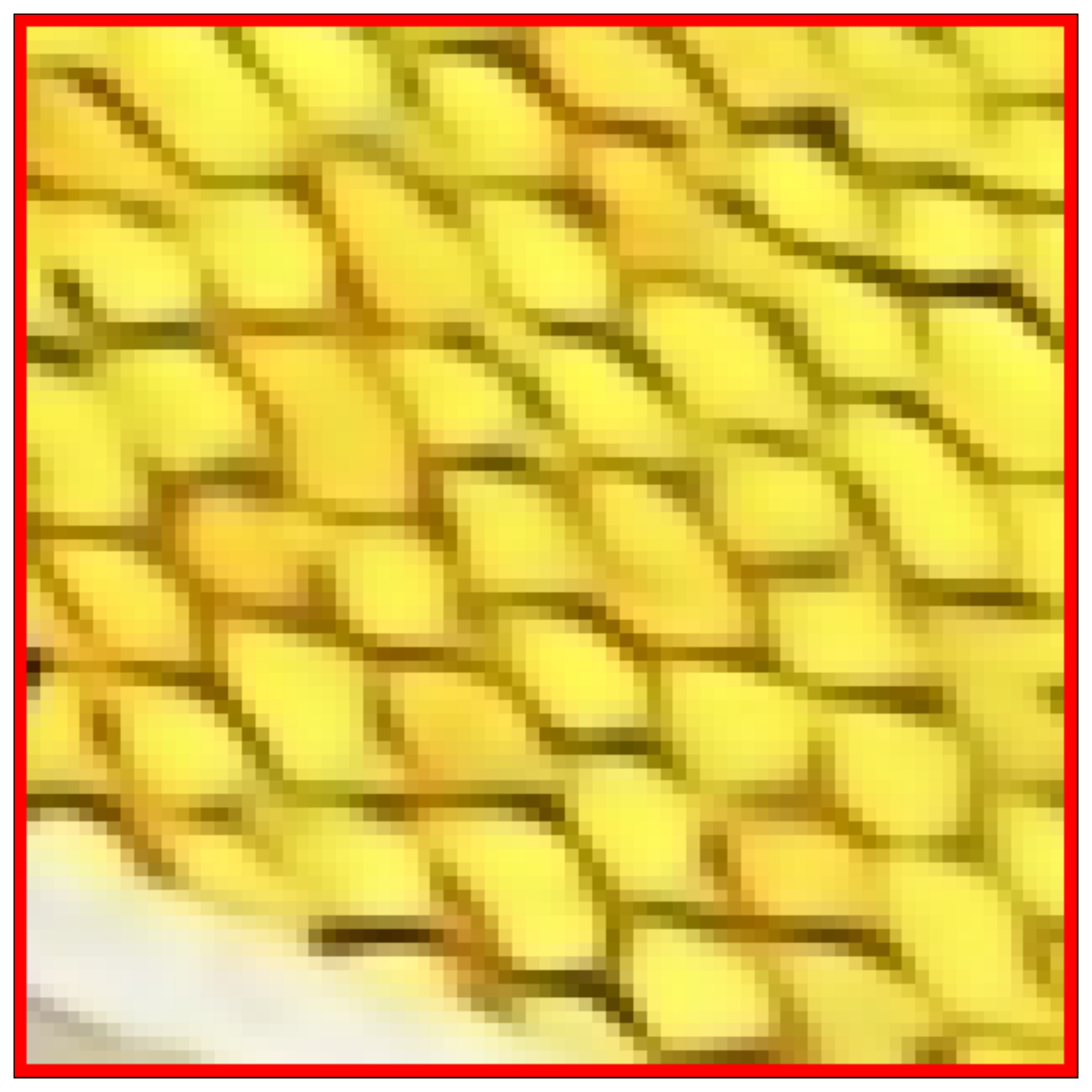}}\
\caption{Visual comparison of image super-resolution with scaling factor $\times 3$. From left to right: the HR images, the LR images synthesized via bicubic degradation, results produced by baseline RDN \cite{zhang2020residual} and results produced by PNB-enhanced RDN. }
\label{fig:sisr_rdn}
\end{figure*}


To validate the effectiveness and necessity of pyramidal strategy, we exhaustively compare PNB with its variants on learning $L_0$ smoothing filter. The performance variations in terms of PSNR and SSIM are shown in Table \ref{table:ablation1}. The multi-scale pyramid non-local networks outperform single-scale non-local networks with a significant margin. It is noteworthy that the design of PNB is flexible according to the trade-off between computation efficiency and accuracy. 

The required computational resources, in terms of floating point operations (FLOPs) and memory consumption, are summarized in Table \ref{table:ablation2}. The performance is obtained by testing methods on a $96\times 96$ patch. We compare our proposed PNB against the original non-local block (NLB) \cite{wang2018non}, and asymmetric pyramid non-local block (APNB) \cite{zhu2019asymmetric} proposed for image semantic segmentation. The original non-local operation brings dramatic increase of memory consumption since it requires to compute a large correlation matrix. In contrast, the additional memory requirement of non-local operation in our method (3.2G) is 62.4\% less than that (8.5G) of \cite{wang2018non}. This means that our method allows larger training patch/batch size or larger receptive field under the same GPU memory. The APNB adopts spatial pyramid pooling to generate global and multi-scale representations. We follow their setting to set the output sizes of adaptive pooling layers as $(1 \times 1, 3 \times 3, 6 \times 6$ and $8 \times 8)$, respectively. APNB is very computationally efficient, but performs even worse than the original NLB in this task. The reason is that the pooling operations disregard lots of texture information. An example of the visual comparison between our proposed PNB and APNB is shown in Fig. \ref{fig:apnb_pnb}. 

\begin{table*}[t]
\centering
\caption{Performance of image denoising methods. Average PSNR metrics are reported on Set12, BSD68 and Urban100 dataset.}
\label{table:denoise}
\setlength\tabcolsep{3pt}
\renewcommand{\arraystretch}{1.2}
\begin{tabular}{l|cc|cc|cc|c|c|c} \specialrule{.1em}{0em}{0em}
\multirow{2}{*}{Method}  &\multicolumn{2}{c|}{Set12}  &\multicolumn{2}{c|}{BSD68} &\multicolumn{2}{c|}{Urban100} &\multirow{2}{*}{FLOPs} &\multirow{2}{*}{Memory} &\multirow{2}{*}{\#Params} \\ \cline{2-7}
               &30           &50           &30           &50           &30           &50  &  & & \\ \hline
MemNet         &29.63 &27.38 &28.43 &26.35 &29.10 &26.65 &26.9G &1.6GB &677k \\ 
MemNet(deeper) &29.67 &27.42 &28.45 &26.36 &29.16 &26.72 &31.5G &1.9GB &1056k \\
MemNet(w/ NLB) &29.66 &27.41 &28.44 &26.37 &29.23 &26.77 &32.7G &10.2GB &791k \\ 
MemNet(w/ PNB)&\textbf{29.72} &\textbf{27.49} &\textbf{28.47} &\textbf{26.41} &\textbf{29.28} &\textbf{26.82} &28.9G &4.8GB &1047k \\ \hline
RDN            &29.94 &27.59 &28.56 &26.38 &30.01 &27.39 &202.9G &2.6GB &22.01M \\ 
RDN(deeper)    &29.94 &27.61 &28.57 &26.39 &30.03 &27.43 &215.5G &2.8GB &23.38M \\ 
RDN(w/ NLB)    &29.95 &27.62 &28.57 &26.41 &30.10 &27.52 &208.4G &11.3GB &22.13M \\
RDN(w/ PNB)   &\textbf{29.97} &\textbf{27.64} &\textbf{28.59} &\textbf{26.43} &\textbf{30.22} &\textbf{27.68} &205.0G &5.8GB &22.38M \\ \specialrule{.1em}{0em}{0em}
\end{tabular}
\end{table*}

\begin{table*}[t]
\centering
\caption{Performance of SISR methods. Average PSNR metrics are reported on Set5, Set14, BSD100 and Urban100 dataset.}
\label{table:sisr}
\setlength\tabcolsep{5pt}
\renewcommand{\arraystretch}{1.2}
\begin{tabular}{l|c|c|c|c|c|c|c} \specialrule{.1em}{0em}{0em}
Method         &Set5   &Set14 &BSD100 &Urban100 &FLOPs &Memory &\#Params \\ \hline
MemNet         &34.09 &30.00 &28.96 &27.56 &26.9G  &1.6GB &677k     \\
MemNet(deeper) &34.12 &30.04 &28.97 &27.65 &31.5G  &1.9GB &1056k  \\
MemNet(w/ NLB) &34.11 &30.07 &28.97 &27.72 &32.7G &10.2GB &791k  \\ 
MemNet(w/ PNB)&\textbf{34.18} &\textbf{30.12} &\textbf{29.02} &\textbf{27.88} &28.9G &4.8GB &1047k  \\ \hline
RDN            &34.76 &30.62 &29.31 &29.01 &205.7G &2.7GB &22.31M  \\ 
RDN(deeper)    &34.77 &30.64 &29.33 &29.05 &218.3G &2.9GB &23.68M  \\ 
RDN(w/ NLB)    &34.76 &30.64 &29.32 &29.10 &211.2G &11.5GB &22.43M  \\
RDN(w/ PNB)   &\textbf{34.80} &\textbf{30.67} &\textbf{29.35} &\textbf{29.19} &207.8G &5.9GB &22.68M  \\ \specialrule{.1em}{0em}{0em}
\end{tabular}
\end{table*}

\section{Extension: Experiments in Image Restoration}
Deep convolutional networks are widely used in image restoration tasks.
We adopt two state-of-the-art methods, MemNet \cite{Tai-MemNet-2017} and RDN \cite{zhang2020residual}, as the baseline models for image denoising and image super-resolution. As discussed in Section \ref{sec:Discussion}, efficient computation allows our proposed pyramid non-local block be incorporated into these low-level baseline models. The PNB acts as a basic component to exploit non-local pixelwise self-similarity.

We ameliorate the MemNet by inserting one PNB ahead of every five memory blocks. We also implement a variant of MemNet by stacking more convolutional layers for fair comparison. The deeper MemNet and PNB-enhanced MemNet have comparable parameter numbers. The performance of non-local block (NLB) \cite{wang2018non} is also reported. One NLB is inserted ahead of every five memory blocks of MemNet.  

Similarly, we ameliorate the RDN through inserting one PNB before every five residual dense blocks, resulting to a PNB-enhanced RDN. To demonstrate the effectiveness of PNB, we compare with deeper RDN by stacking more convolutional layers and NLB-enhanced RDN.

\textbf{Image Denoising:} To train MemNet and its related variants, we follow \cite{Tai-MemNet-2017} to use 300 images from the Berkeley Segmentation Dataset (BSD) \cite{MartinFTM01} as the training set. To train RDN and its related variants, we follow the protocols of \cite{zhang2020residual}, where the DIV2K dataset \cite{timofte2017ntire} are utilized as training data. We add white Gaussian noise with standard deviation $\sigma=30$ and $\sigma=50$ to the original images to synthesize noisy images, respectively. Results on three widely used benchmarks, Set12 \cite{zhang2017beyond}, BSD68 \cite{roth2009fields}, Urban 100 \cite{huang2015single}, are reported in Table \ref{table:denoise}. As we can see, PNB-enhanced MemNet and PNB-enhanced RDN achieves the best performance among their original networks and related variants, respectively. An example of visual comparison is presented in Fig. \ref{fig:denoising}. PNB-enhanced RDN recovers clearer structures than original RDN from severely degraded noisy images.

\textbf{Image Super-resolution:} To train MemNet and its related variants, we follow the experimental setting in \cite{Tai-MemNet-2017}. The training set includes 291 images where 200 images are from the training set of BSD  \cite{MartinFTM01} and other 91 images are from \cite{yang2010image}. To train RDN and its related variants, we follow \cite{zhang2020residual} to use high-quality dataset DIV2K and Flickr2K \cite{timofte2017ntire} as the training data. The low-resolution input images are synthesized by bicubic downsampling with factor of $3$, and then upscaled to the original size. Results on four widely used benchmarks, Set5 \cite{bevilacqua2012low}, Set14 \cite{zeyde2010single}, BSD100 \cite{MartinFTM01} and Urban100 \cite{huang2015single}, are presented in Table \ref{table:sisr}. The PNB-enhanced MemNet and PNB-enhanced RDN give rise to best results among their original networks and related variants, respectively. Fig. \ref{fig:sisr_memnet} and Fig. \ref{fig:sisr_rdn} show the super-resolution results. Both PNB-enhanced MemNet and PNB-enhanced RDN recover shaper and clearer edges than the original MemNet and RDN, respectively.

\section{Conclusions \& Future Work}
In this paper, we present a novel and effective pyramid non-local enhanced network (PNEN) for edge-preserving image smoothing. The proposed pyramid non-local block (PNB) is computation-friendly, which allows it be a plug-and-play component in existing deep methods for low-level image processing tasks. Methods incorporating our proposed pyramid non-local block achieve significantly improved performance in image denoising and image super-resolution.

In our pyramid structure, feature maps of different scales are independently generated with different convolutional kernels. However, a large kernel size is required for a large scale. To further reduce the computation and memory cost, a potential direction is to recursively generate multi-scale feature maps using convolutions with fixed and small kernels.

\bibliographystyle{IEEEtran}
\normalem
\bibliography{egbib}

\end{document}